\def\year{2021}\relax
\documentclass[letterpaper]{article} 
\usepackage{aaai21}  
\usepackage{times}  
\usepackage{helvet} 
\usepackage{courier}  
\usepackage[hyphens]{url}  
\usepackage{graphicx} 
\usepackage{natbib}  
\usepackage{caption} 
\usepackage{booktabs}
\usepackage{algorithmic,algorithm}
\usepackage{amsmath}
\usepackage{amssymb}
\usepackage{bm}
\newtheorem{definition}{Definition}

\newtheorem{theorem}{Theorem}
\newtheorem{lemma}{Lemma}

\usepackage{subcaption}
\usepackage{bm}

\title{Simplifying Deep Reinforcement Learning via Self-Supervision}
\author{
    Daochen Zha, Kwei-Herng Lai, Kaixiong Zhou, Xia Hu
    \\
}
\affiliations{
    Department of Computer Science and Engineering, Texas A\&M University \\
    \{daochen.zha, khlai037, zkxiong, xiahu\}@tamu.edu
}

\begin{document}

\maketitle

\begin{abstract}
Supervised regression to demonstrations has been demonstrated to be a stable way to train deep policy networks. We are motivated to study how we can take full advantage of supervised loss functions for stably training deep reinforcement learning agents. This is a challenging task because it is unclear how the training data could be collected to enable policy improvement. In this work, we propose Self-Supervised Reinforcement Learning (SSRL), a simple algorithm that optimizes policies with purely supervised losses. We demonstrate that, without policy gradient or value estimation, an iterative procedure of ``labeling" data and supervised regression is sufficient to drive stable policy improvement. By selecting and imitating trajectories with high episodic rewards, SSRL is surprisingly competitive to contemporary algorithms with more stable performance and less running time, showing the potential of solving reinforcement learning with supervised learning techniques. The code is available on GitHub\footnote{\url{https://github.com/daochenzha/SSRL}}.

\end{abstract}

\section{Introduction}
Harnessing the power of deep neural networks for function approximation is a long-standing challenge for reinforcement learning~\cite{sutton2018reinforcement}. Existing gradient-based methods for training neural networks assume the samples to be independently and identically distributed~(i.i.d.). Unfortunately, in the context of model-free reinforcement learning, we usually encounter highly correlated data that violate the i.i.d. assumption, which is problematic for deep learning methods.

A lot of research efforts have been dedicated to tackling the above problem, such as value-based methods~\cite{mnih2015human,hasselt2010double,hessel2018rainbow}, and policy-based approaches~\cite{mnih2016asynchronous,schulman2017proximal}. While deep reinforcement learning algorithms have evolved to be increasingly powerful, they are notoriously unstable and hard to train~\cite{henderson2018deep}, which may be due to various factors, such as the high variance of policy gradient and the overestimation of the values~\cite{van2016deep}. This makes the algorithms hard to use and even leads to reproducibility issues for reinforcement learning~\cite{islam2017reproducibility} and also its applications, such as neural architecture search~\cite{li2020random}.

In this work, we approach this issue from the perspective of supervised learning, which has been demonstrated to be a more stable way to train deep policy networks. Our motivation is from the success of imitation learning~\cite{ho2016generative}, which directly learns a mapping from states to actions with supervised regression to demonstrations. From the perspective of deep reinforcement learning, recent work observes that, by simply training the policy based on the demonstrations of a reinforcement learning agent, we can obtain a much more stable model~\cite{rusu2015policy} and achieve faster convergence~\cite{oh2018self,lin2019ranking,gangwani2018learning,li2020autood,zha2021rank}. These interesting findings motivate us to study how we can take full advantage of supervised learning towards stable policy improvement.

The main challenge of employing supervised learning for policy improvement is the difficulty of identifying demonstrations. In imitation learning, we have access to expert demonstrations, and thus we can guarantee policy improvement with supervised regression. However, it is unlikely to hit expert demonstrations with an untrained policy in the context of reinforcement learning. Given only sparse rewards from the environment, it is nontrivial to determine which trajectories should be regarded as demonstrations.

To address the above challenge, we propose Self-Supervised Reinforcement Learning~(SSRL). The key idea of SSRL is to keep asking the question ``\emph{what data are more likely to be sampled by a policy that is better than the current policy?}". The intuition behind it is that as long as we can collect sufficient demonstrations from a better policy, we can improve the current policy with supervised regression to these data. The agent achieves policy improvement by iteratively executing a two-step self-supervised procedure: (1) collect data from the environment and ``label" a part of highly rewarded data as demonstrations; (2) conduct supervised regression to these ``labeled" data to improve itself. With this procedure, the agent continuously refines the demonstrations, and eventually finds the expert demonstrations and converges to the optimal policy.

This work conducts a pilot study with empirical and theoretical evidence to show that the above simple idea will drive stable policy improvements in many application scenarios. We design a wide range of experiments to test SSRL on various tasks, including deterministic/non-deterministic MDPs, discrete/continuous action spaces, high-dimensional image inputs, and hard exploration domains. While our goal is not to match the state-of-the-arts, we find that a simple ranking-based algorithm can surprisingly achieve competitive or even better performance in many environments than very mature policy- or value-based algorithms, such as DQN, PPO, and DDPG. We show that SSRL achieves almost monotonous improvement during training and is stable after converged. Moreover, given the simplicity of the approach, SSRL is easy to implement and requires very few computational resources. To summarize, we make the following contributions:

\begin{itemize}
    \item Propose a simple self-supervised procedure that trains reinforcement learning agents with supervised regression.
    \item Theoretically prove that SSRL will drive policy improvement in deterministic MDPs.
    \item Conduct experiments on environments with both discrete and continuous action spaces to show that SSRL is competitive to state-of-the-art model-free algorithms, such as PPO and DDPG, in terms of sample efficiency with more stable performance and much less running time.
    \item Demonstrate that SSRL can well deal with high-dimensional inputs and can be enhanced with exploration strategies, showing the potential of solving reinforcement learning with supervised learning techniques.
\end{itemize}

\section{Preliminaries}
\subsection{Reinforcement Learning}
We consider finite-horizon Markov Decision Process (MDP) $\mathcal{M} = \langle \mathcal{S}, \mathcal{A}, \mathcal{P}, r, \gamma, p_0 \rangle$, where $\mathcal{S}$ is the set of states, $\mathcal{A}$ is the set of actions, $\mathcal{P}: \mathcal{S} \times \mathcal{A} \to \mathcal{S}$ is the state transition function, $r: \mathcal{S} \to \mathbb{R}$ is the reward function, $\gamma \in (0, 1)$ is the discount factor, and $p_0$ is the distribution of the initial state. At each timestep $t$, an agent takes action $a_t \in \mathcal{A}$ in state $s_t \in \mathcal{S}$ and observes the next state $s_{t+1}$ with a scalar reward $r_t$, resulting in a trajectory $\tau = \{ \langle s_t, a_t, r_t \rangle \}_{i=0}^{T}$, where $T$ is the timestep at which the game terminates. We use $r(\tau) = \sum_{t=0}^T r_t$ to denote the trajectory cumulative reward. A stochastic policy $\pi: \mathcal{S} \times \mathcal{A} \to \mathbb{R}$ gives a probability for each state-action pair $\pi(a|s)$, where $s \in \mathcal{S}$, $a \in \mathcal{A}$, and $\sum_{a \in \mathcal{A}} \pi(a|s) = 1$. The objective of reinforcement learning is to learn a policy $\pi$ that maximizes the expected discounted cumulative reward $\mathbb{E}[\sum_{t=0}^{T}\gamma^t r_t]$. With neural function approximators, we use $\pi_\theta$ to denote a parameterized policy with parameters $\theta$.

\subsection{Imitation Learning}
Imitation learning provides an alternative way to solve the above MDP. Suppose we are not allowed to use the reward signals, but instead are given some expert demonstrations $\mathcal{D} = \{\langle s_i, a_i \rangle\}^N_{i=1}$, where $N$ is the size of the data. Imitation learning assumes that the data are generated by an expert policy $\pi_{E}$, and learns a policy $\pi: \mathcal{S} \times \mathcal{A} \to \mathbb{R}$ that reproduces similar actions as $\pi_{E}$ given the same states, which indirectly optimizes the expected cumulative reward (if $\pi_{E}$ is optimal). In this work, we employ behavior cloning, the simplest form of imitation learning, where supervised learning methods are adopted to learn the mapping $\mathcal{S} \times \mathcal{A} \to \mathbb{R}$.



\begin{figure}[t]
    \centering
    \includegraphics[width=8cm]{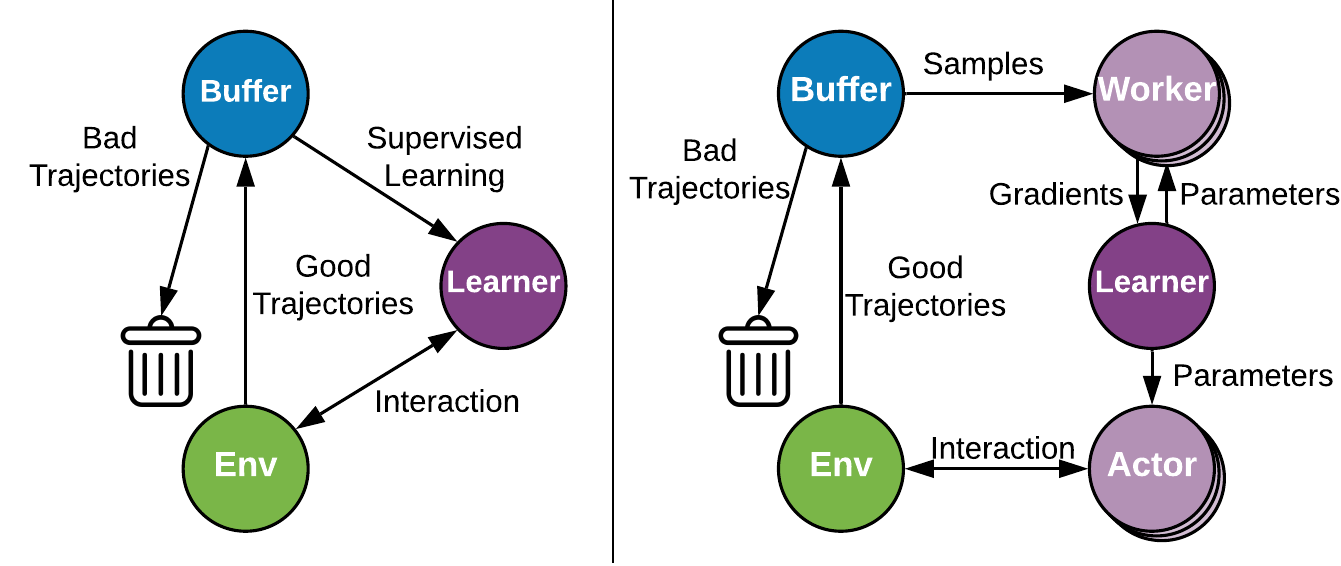}
    \caption{Self-Supervised Reinforcement Learning~(SSRL) with a single learner~(\textbf{left}) and multiple actors/workers~(\textbf{right}). The buffer continuously refines the data by collecting highly-rewarded trajectories from the environment and discarding the bad ones. The agent improves the policy with supervised learning by treating the data in the buffer as demonstrations.}
    \label{fig:overview}
\end{figure}

\section{Self-Supervised Reinforcement Learning}
\label{sec:method}
In this section, we first introduce the high-level idea of Self-Supervised Reinforcement Learning~(SSRL). Then we instantiate SSRL with a ranking buffer to select and imitate highly-rewarded trajectories. Finally, we provide some theoretical insights for policy improvements of SSRL.

\subsection{Self-Supervised Training of Policy Networks}
An overview is illustrated in Figure~\ref{fig:overview}. The core idea of SSRL is to give preferences to the behaviors that are more likely to be better than the average behaviors of the current policy $\pi$ through interacting with the environment. Then, a slightly better policy $\pi': \mathcal{S} \times \mathcal{A} \to \mathbb{R}$ can be trained via any supervised learning techniques by treating the selected behaviors as demonstrations. Specifically, we first randomly initialize a policy $\pi_\theta$ with parameters $\theta$ and a buffer $\mathcal{D}$ which stores past state-action pairs $\langle s, a \rangle$. Then we optimize policies with the following iterative procedure:

\begin{enumerate}
    \item Collect trajectories with the current policy. Update the buffer with good state-action pairs. 
    \item Update $\pi_\theta$ with any supervised learning methods using the state-action pairs in the buffer.
    \item If not converged, go to 1.
\end{enumerate}

Unlike policy- or value-based algorithms that design various loss functions to update the policy or value networks, we alternatively focus on collecting good trajectories to describe the desired behaviors of the agent. Good trajectories are defined as the ones that achieve high episodic rewards since episodic rewards are usually used as the criterion to measure the performance of the agent. With sufficient good demonstrations, the policy can be naturally obtained via imitation learning. Starting from random policy, we collect some good trajectories and improve the policy with supervised learning; in the next iteration, we sample data with slightly better policy and thus are more likely to find better trajectories. This design allows us to use simple supervised losses to train policy networks. As long as the distribution of the state-action pairs in the buffer does not change much, i.e., we only update a small portion of the buffer in each iteration, the training will be close to supervised learning with stable policy updates.

The above optimizing procedure relies on the ability of the agent to discover good trajectories in the first place. Thus, it is crucial that the agent can explore the environment and hit good trajectories. We find that a stochastic policy works well since it can occasionally generate some better trajectories due to the policy's randomness. Any exploration techniques could be naturally incorporated into step 1 to enhance efficiency. To make our contribution focused, we keep the algorithm simple and do not use any exploration technique in the experiments.

Note that we do not expect the state-action pairs in the buffer $\mathcal{D}$ to be optimal at the beginning. The demonstrations could be optimal, fair, or even terrible, as long as they perform slightly better than the average behaviors of the current policy. Unlike the previous work where the demonstrations are assumed to be perfect~\cite{ziebart2008maximum}, or near optimal~\cite{lin2019ranking}, our framework continuously refines the demonstrations in the buffer during the learning process and eventually finds the expert-level demonstrations and discovers a good policy.

Our framework could also be applied in the scenarios where some demonstrations are available. By decoupling reinforcement learning into data collection and supervised regression, our objective instead becomes collecting enough good demonstrations rather than directly optimizing policies. As such, we can populate the buffer with the known demonstrations in the initial stage to guide the exploration. We will explore the possibility of leveraging known demonstrations under our framework in our future work.

\subsection{Trajectory Selection with Ranking Buffer}

We propose to use a simple ranking buffer to instantiate the above idea. Specifically, we assign the episodic reward to all the state-action pairs in the episode and store them into the buffer for each generated episode. We choose to operate on the state-action pairs for the ease of implementation since behavior cloning will sample and update with state-action pairs in each iteration. The ranking buffer selects highly-rewarded state-action pairs by discarding the ones with low episodic rewards. For a sampled batch $\{(s_i, a_i)\}_{i=1}^N$, the policy is either updated via behavior cloning with log loss for discrete action space:
\begin{equation}
    \label{eqn:2}
    \mathcal{L}_{\text{discrete}} = - \frac{1}{N}\sum_{i=1}^N \sum_{a \in \mathcal{A}} \text{I}(a_i, a) \log \pi_\theta(a|s_i),
\end{equation}
where $\text{I}(a_i, a)=1$ iff $a_i=a$ and 0 otherwise, or mean-square-error loss for continuous action space:
\begin{equation}
    \label{eqn:1}
    \mathcal{L}_{\text{continuous}} = \frac{1}{N}\sum_{i=1}^N (\pi_\theta(a_i|s_i) - a_i)^2,
\end{equation}
where $a_i$ is a continuous vector. We summarize the procedure in Algorithm~\ref{alg:1}. The $\pi_\theta$ is a stochastic policy that samples actions from multinomial or Gaussian distributions for discrete or continuous cases, respectively. The stochastic nature of policy ensures that $\pi(a|s) >0, \forall s \in \mathcal{S}, \forall a \in \mathcal{A}$ so that any trajectories is reachable. As a result, the agent will occasionally hit some good trajectories due to the randomness. This property is crucial for the algorithm because it relies on the ability of the agent to discover good trajectories in the first place.

The ranking buffer achieves a balance between exploration and stability. Specifically, the ranking mechanism will make sure that only highly rewarded state-action pairs are stored in the buffer. The selection will result in very few state-action pairs being updated in each iteration. Thus, the state-action pairs in the buffer will not change much during the training process. As such, the supervised update will be stable since the underlying distribution does not change much. The ranking buffer ensures that training will not be affected by those irrelevant trajectories with low reward and makes the agent focus on the good trajectories.

While the above instance of SSRL seems trivial, we find in practice that it surprisingly delivers non-trivial performance in many tasks. We believe the above simple algorithm can be adapted to achieve even stronger performance in real-world applications, e.g., with some exploration strategies or better buffer management strategies. Nonetheless, instead of desperately optimizing the performance with more engineering efforts, we focus on the above simple algorithm to better understand the potential of SSRL.

\begin{algorithm}[t]
\caption{An instance of SSRL with a ranking buffer}
\label{alg:1}
\setlength{\intextsep}{0pt} 
\begin{algorithmic}[1]
\STATE \textbf{Input:} training steps $S$, buffer size $D$
\STATE Initialize $\pi_{\theta}$, replay buffer $\mathcal{D}$.
\FOR{iteration = $1$, $2$, ... until convergence}
    \STATE Execute $\pi_\theta$ to generate an episode ${\tau}$
    \STATE Assign the episodic reward $r(\tau)$ to all the state-action pairs in $\tau$ and store them into buffer $\mathcal{D}$
    \STATE Rank the state-action pairs in $\mathcal{D}$ based on $r(\tau)$
    \IF {$\mathcal{D}.length > D$}
        \STATE Discard the state-action pairs with low reward
    \ENDIF
    \FOR{step = $1$, $2$, .., $S$}
        \STATE Sample a batch of state-action pairs from $\mathcal{D}$ and update $\pi_\theta$ with Eq.~\ref{eqn:1} or Eq.~\ref{eqn:2}
    \ENDFOR
\ENDFOR
\end{algorithmic}
\end{algorithm}

\subsection{Theoretical Justification}
\label{sec:theo}
We justify that our framework will theoretically drive policy improvement in deterministic MDPs where the environment is not susceptible to randomness (i.e., the same action will always has the same outcome in a given state).

To prove this, we first define the uniformly-distributed condition for a set of trajectories $\bm{\tau}$ as follows. For convenience, we use notation $C$ to count the number of transitions in $\bm{\tau}$. Specifically, we define $C(\bm{\tau}, s, a ,s', t)$ as the number transitions in $\bm{\tau}$ that take action $a$ in state $s$ and transit to state $s'$ at timestep $t$. In what follows, we abuse the notations to represent summarizations of the counts. For example, we use $C(\bm{\tau}, s, a ,\cdot, t)$ to represent the number of transitions that take action $a$ in state $s$ at timestep $t$, i.e., $C(\bm{\tau}, s, a ,\cdot, t) = \sum_{s'} C(\bm{\tau}, s, a ,s', t)$, and we use $C(\bm{\tau}, s,\cdot ,\cdot, t)$ to represent the number of transitions in state $s$ at timestep $t$, i.e., $C(\bm{\tau}, s,\cdot ,\cdot, t) = \sum_{s'} \sum_a C(\bm{\tau}, s, a ,s', t)$.


\begin{definition}[Uniformly-Distributed]
\label{def:1}
Trajectories $\bm{\tau}$ are uniformly-distributed if the initial states frequencies are consistent with the initial states distribution of the environment, i.e, $\forall s, p_0 (s) = \frac{C(\bm{\tau}, s, \cdot, \cdot, 0)}{|\bm{\tau}|}$, the state transition frequencies for all timesteps are consistent with the state transition probabilities of the environment, i.e.,$\forall s, a, s', t, \mathcal{P}(s' | s, a) = \frac{C(\bm{\tau}, s, a, s', t)}{C(\bm{\tau}, s, a, \cdot, t)}$.
\end{definition}

In a deterministic MDP, the uniformly-distributed condition will hold no matter how we collect the data because (1) the initial state is deterministic; (2) the state transition is deterministic so that the state transition frequencies are always consistent with the state transition probabilities. Now we prove that if the uniformly-distributed condition holds, the supervised learning step can drive policy improvement.

\begin{theorem}[Policy-Improvement]
\label{theorem:1}
Suppose the uniformly-distributed condition holds for the trajectories in $\mathcal{D}$. We can construct a hypothetical policy $\widetilde{\pi}$ whose expected cumulative rewards is at least as good as that of the current policy $\pi$, and the supervised learning step is equivalent to imitating this hypothetical policy.
\end{theorem}

Theorem~\ref{theorem:1} can be derived by constructing a hypothetical policy with the data in the buffer. We prove that the expected cumulative rewards of the hypothetical policy are at least as good as the current policy and are strictly better if the average reward in the buffer is larger than the average reward of the current policy. We show that the supervised learning step is equivalent to imitating the hypothetical policy, naturally leading to policy improvement. We provide the complete proof in \textbf{Appendix A}. While analyzing the policy improvement property in stochastic environments is difficult, our empirical results suggest that SSRL can also achieve good results. In the future, we will try extending the theorem to stochastic environments.

\section{Experiments}
We design the experiments to study the questions as follows:
\begin{itemize}
    \item \textbf{Q1:} How does the data in the buffer of SSRL evolve throughout the training process in deterministic environments?
    \item \textbf{Q2:} How will SSRL perform in non-deterministic discrete and continuous control tasks?
    \item \textbf{Q3:} Is the training of SSRL stable across various seeds?
    \item \textbf{Q4:} Can SSRL handle raw image pixels in inputs?
    \item \textbf{Q5:} Can existing exploration strategies be applied to enhance the exploration ability of SSRL?
\end{itemize}
We provide the hyperparemeters and more details of all the experiments in \textbf{Appendix B}.

\subsection{A Case Study on Taxi Environment} \label{exp:1}

\begin{figure}[t]
  \centering

  \begin{subfigure}[b]{0.15\textwidth}
    \centering
    \includegraphics[width=0.98\textwidth]{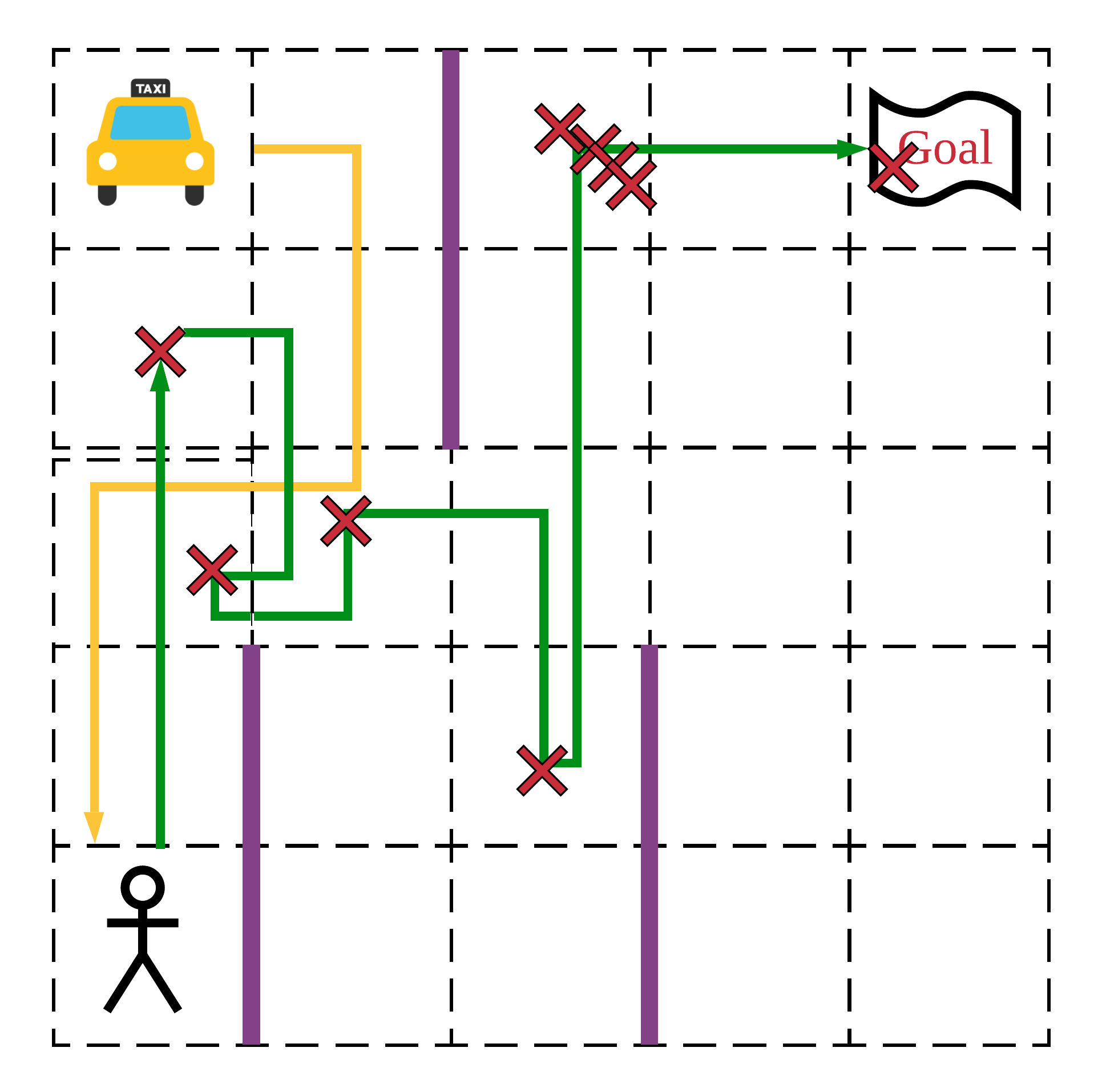}
  \end{subfigure}%
  \begin{subfigure}[b]{0.15\textwidth}
    \centering
    \includegraphics[width=0.98\textwidth]{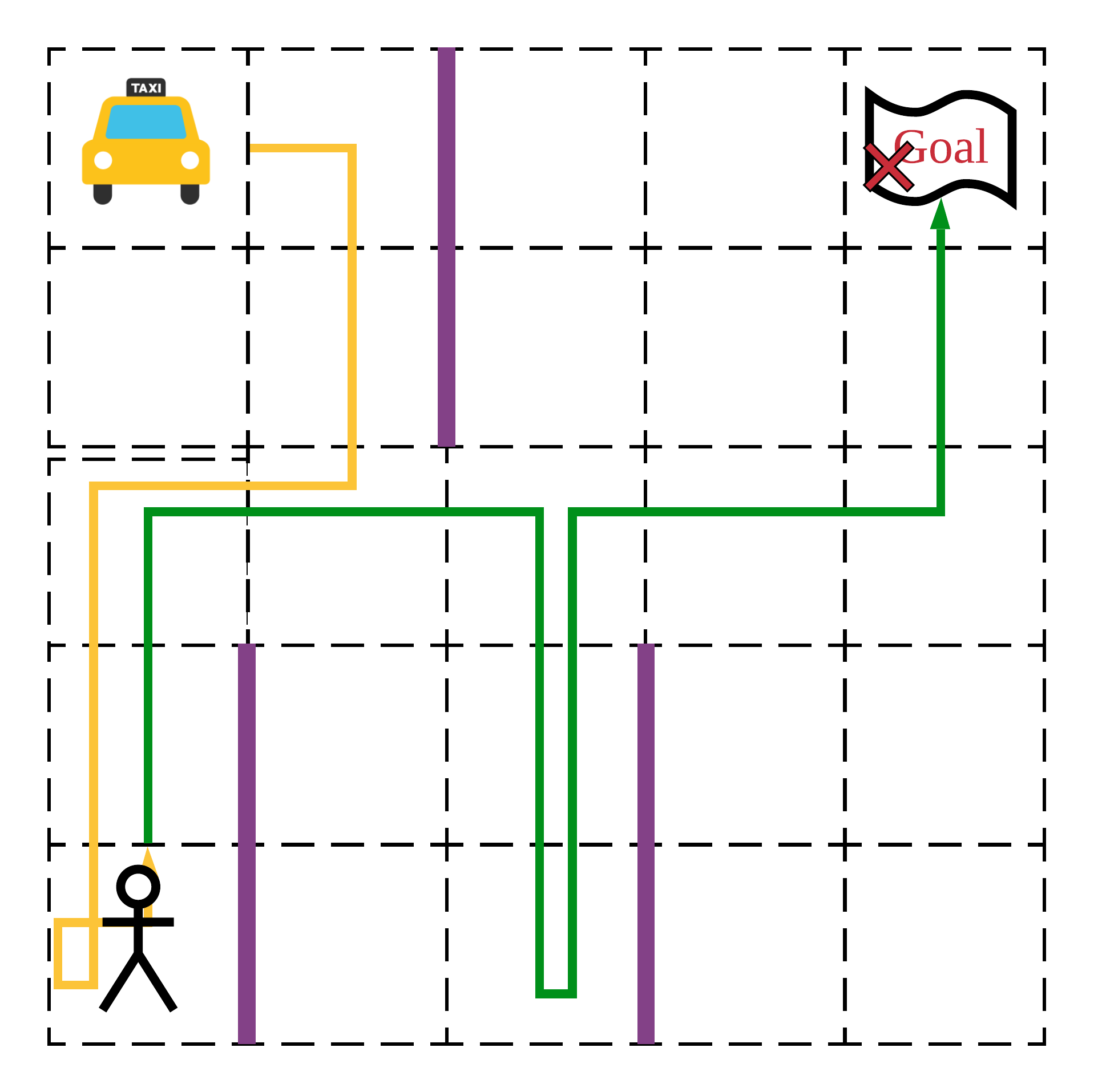}
  \end{subfigure}%
  \begin{subfigure}[b]{0.15\textwidth}
    \centering
    \includegraphics[width=0.98\textwidth]{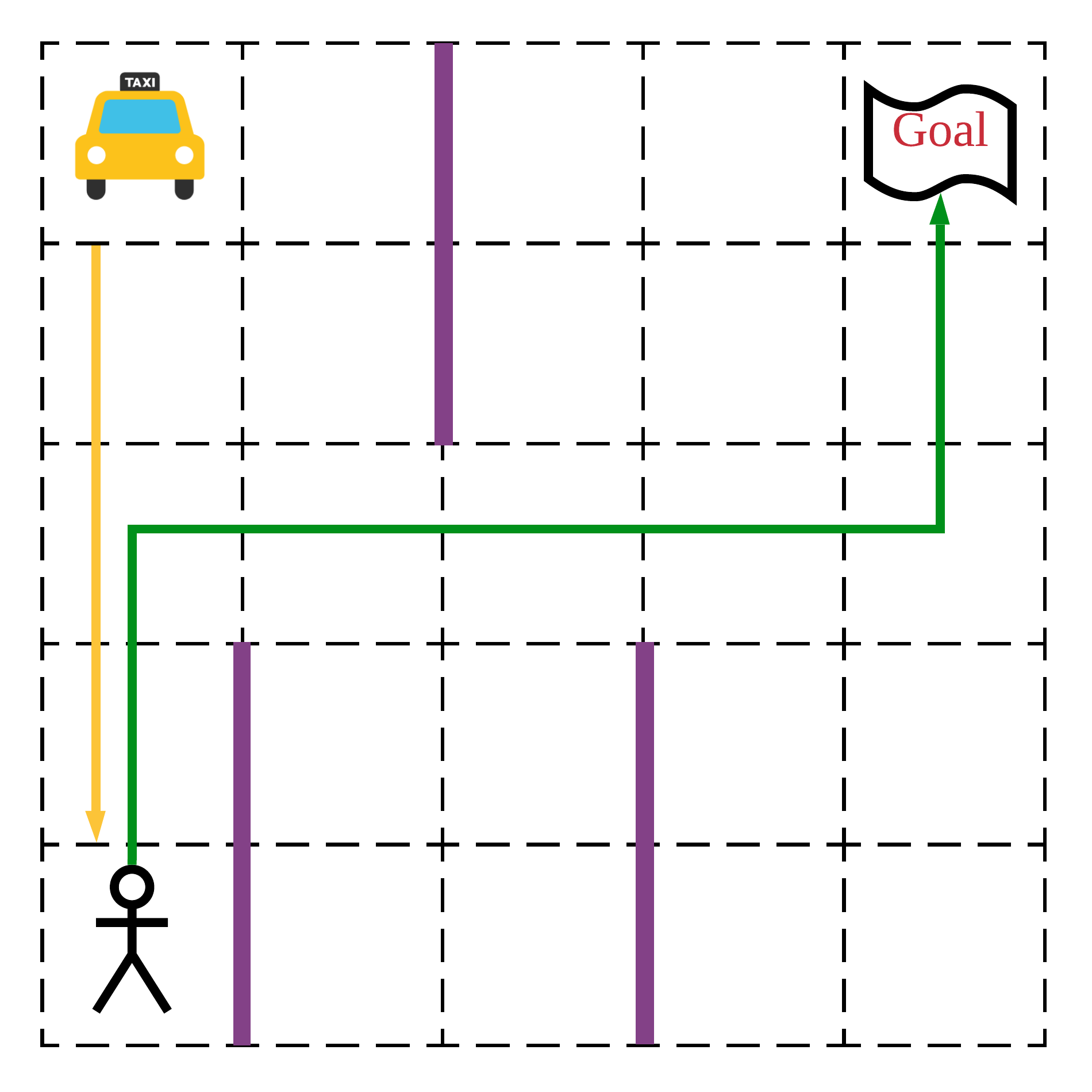}
  \end{subfigure}%
 
  \begin{subfigure}[b]{0.15\textwidth}
    \centering
    \includegraphics[width=0.98\textwidth]{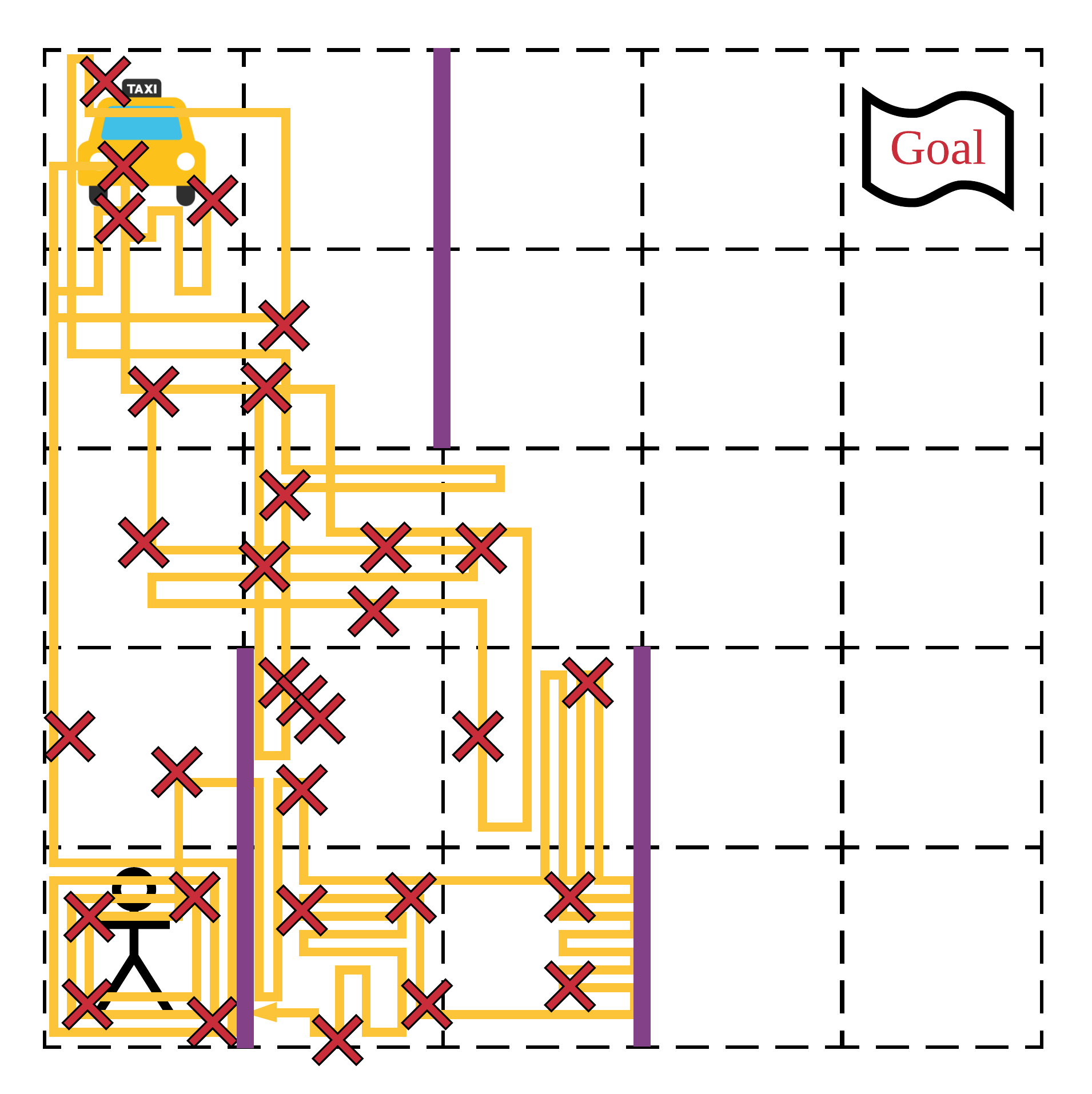}
    \caption{Early Stage}
  \end{subfigure}%
  \begin{subfigure}[b]{0.15\textwidth}
    \centering
    \includegraphics[width=0.98\textwidth]{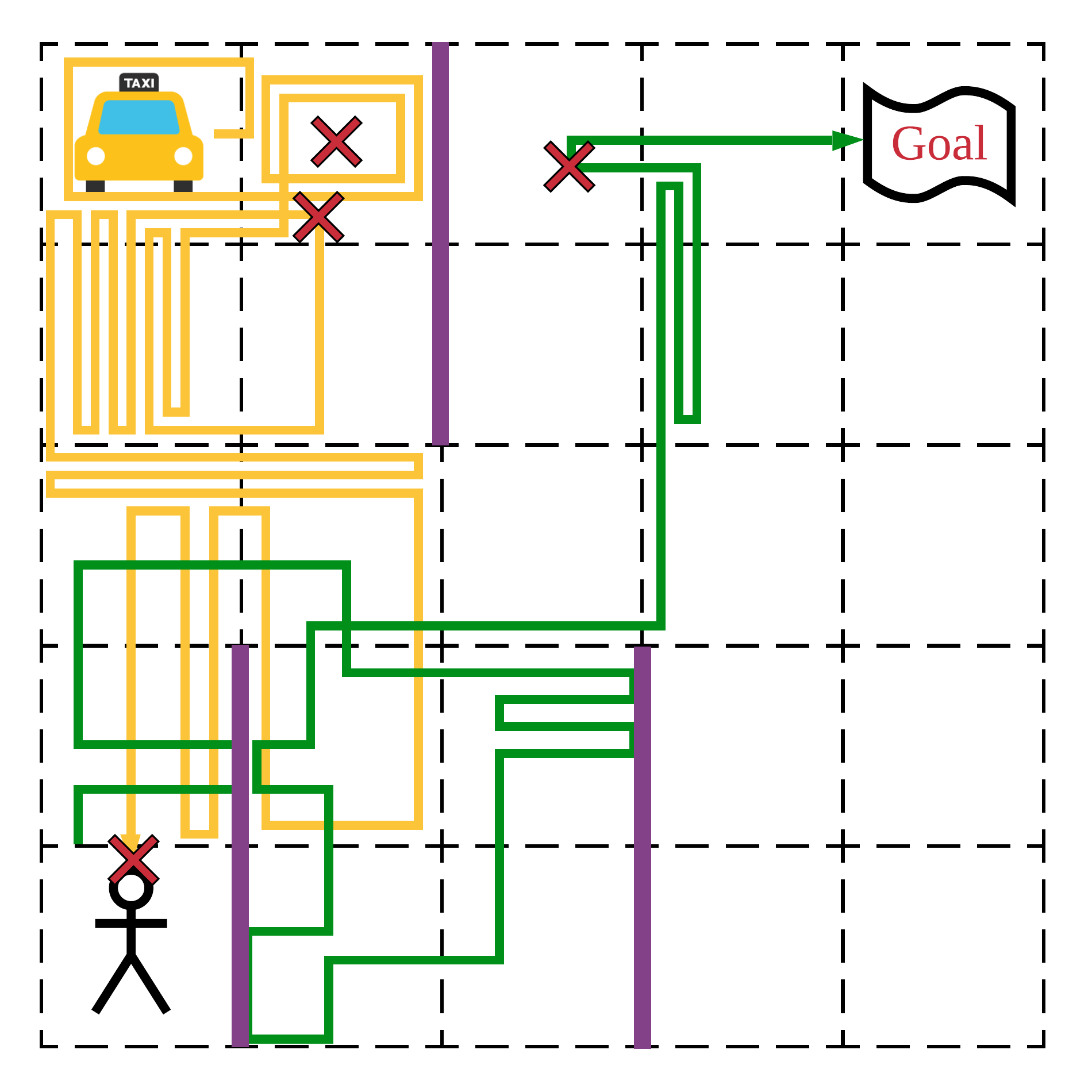}
    \caption{Middle Stage}
  \end{subfigure}%
  \begin{subfigure}[b]{0.15\textwidth}
    \centering
    \includegraphics[width=0.98\textwidth]{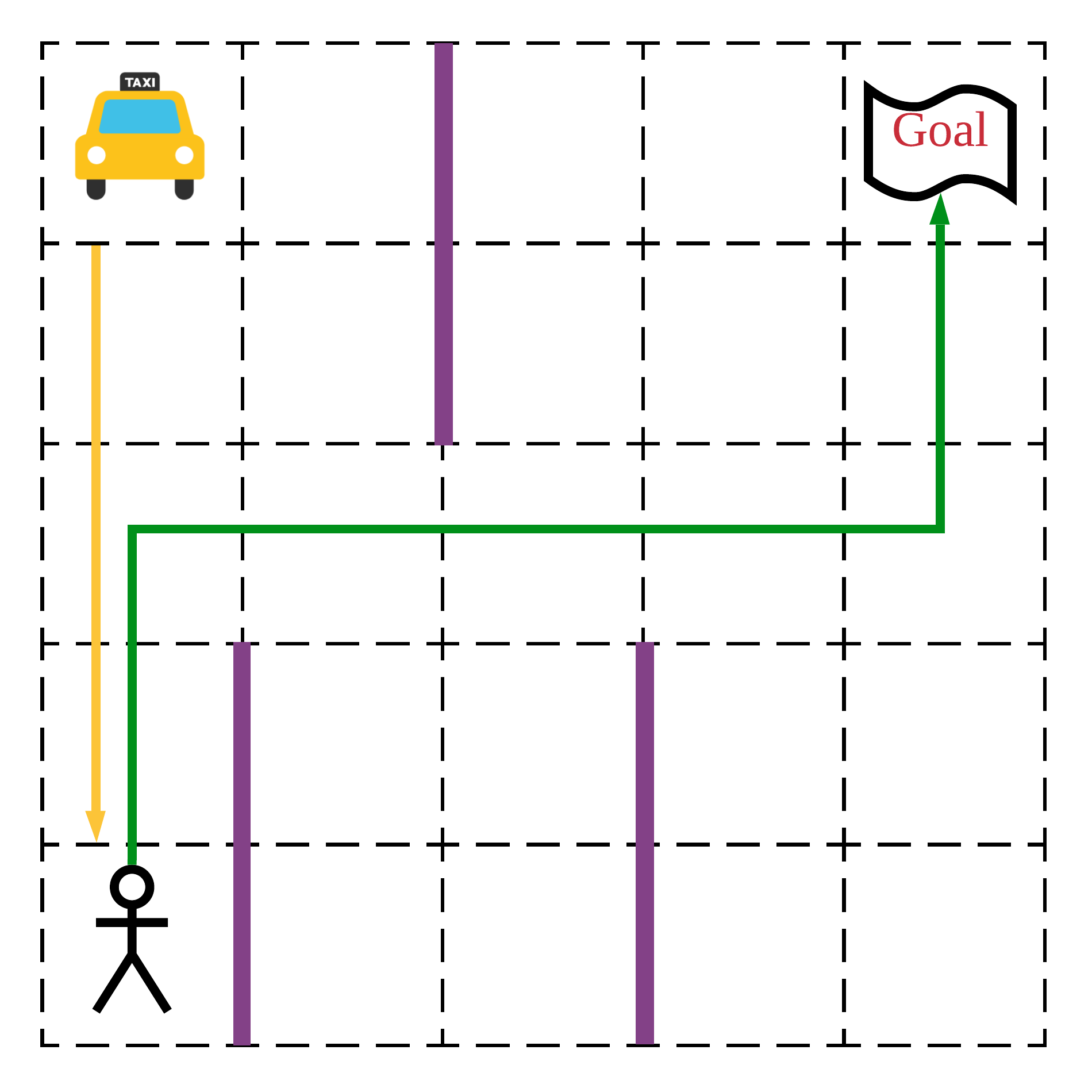}
    \caption{Final Stage}
  \end{subfigure}%
  \caption{A case study on a Taxi environment, where the taxi (top-left corner) needs to pick up the passenger (bottom-left corner) and drive her to the goal (top-right corner). We visualize the best episode (top row) and the worst episode (bottom row) in the buffer throughout the training process. Yellow line represents the cases when the passenger is not picked up; green line suggests the passenger has been picked up and is in the taxi; $\times$ denotes an invalid Pick Up or Drop Off action, which is largely penalized.}
  \label{fig:casestudy}
\end{figure}

To answer \textbf{Q1}, we consider a simple Taxi environment (Figure~\ref{fig:casestudy}). This is a deterministic environment where the taxi needs to pick up the passenger and drive her to the goal. The possible actions are Up, Down, Left, Right, Pick Up, and Drop Off. For every timestep spent, the agent will be given a $-1$ penalty. For every wrong Pick Up or Drop Off, the agent will receive a $-10$ penalty. The agent will receive a $+20$ reward if it navigates the goal. To understand how SSRL behaves, we visualize the trajectories with the highest total reward (top row) and the lowest total reward (bottom row) in the buffer throughout the training process (we allow the buffer to keep the whole episodes for visualization).

In the early stage, both the best and the worst episodes do not perform well. While the best episode has navigated the goal, it chooses a sub-optimal path and performs a lot of invalid Pick Up or Drop Off. Nevertheless, the ranking mechanism will give preference to the best episode and encourage the agent to reproduce good behaviors with supervised learning. In the middle stage, the agent gradually discovers better episodes and stores them in the buffer. The ranking mechanism will again favor better episodes so that the agent will learn to generate trajectories with high rewards. Finally, both the best and the worst episodes become optimal, and the policy naturally becomes optimal. This procedure differs from the policy- and value-based algorithms in that we use the data in the buffer to describe the agent's desired behavior. In practice, the distribution shift in the buffer is usually small since the ranking buffer only stores those highly-rewards episodes, which enables stable policy improvement with supervised learning.

\begin{figure*}[t]
  \centering
  \begin{subfigure}[b]{0.6\textwidth}
    \includegraphics[width=1.0\textwidth]{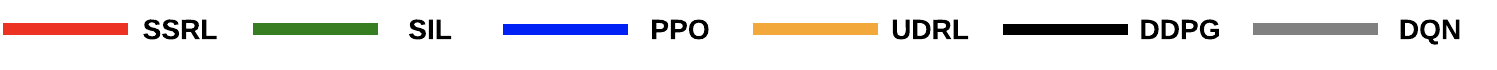}
  \end{subfigure}

  \begin{subfigure}[b]{0.2\textwidth}
    \centering
    \includegraphics[width=0.9\textwidth]{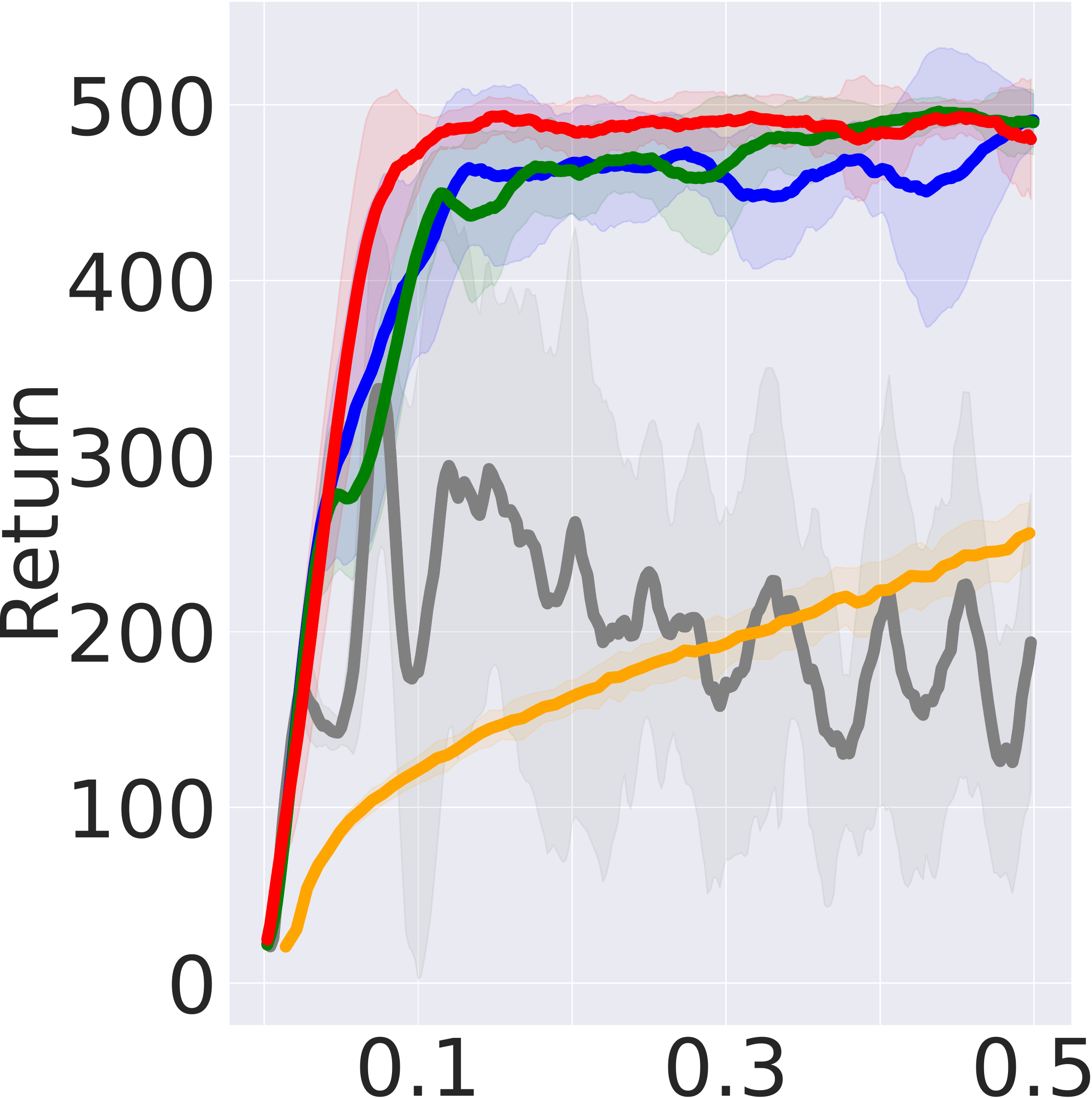}
    \vspace{4pt}
  \end{subfigure}%
  \begin{subfigure}[b]{0.2\textwidth}
    \centering
    \includegraphics[width=0.9\textwidth]{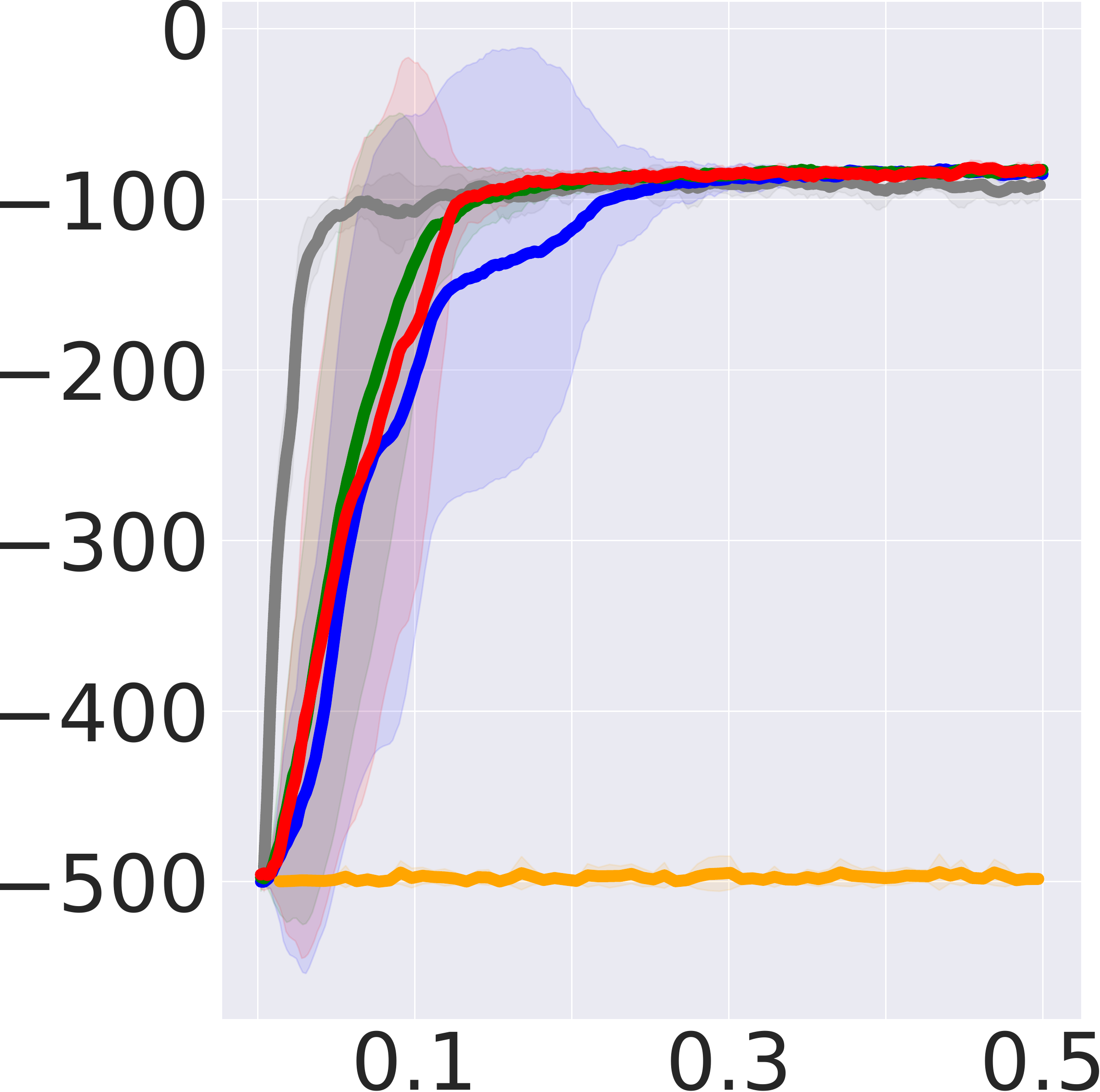}
    \vspace{4pt}
  \end{subfigure}%
  \begin{subfigure}[b]{0.2\textwidth}
    \centering
    \includegraphics[width=0.9\textwidth]{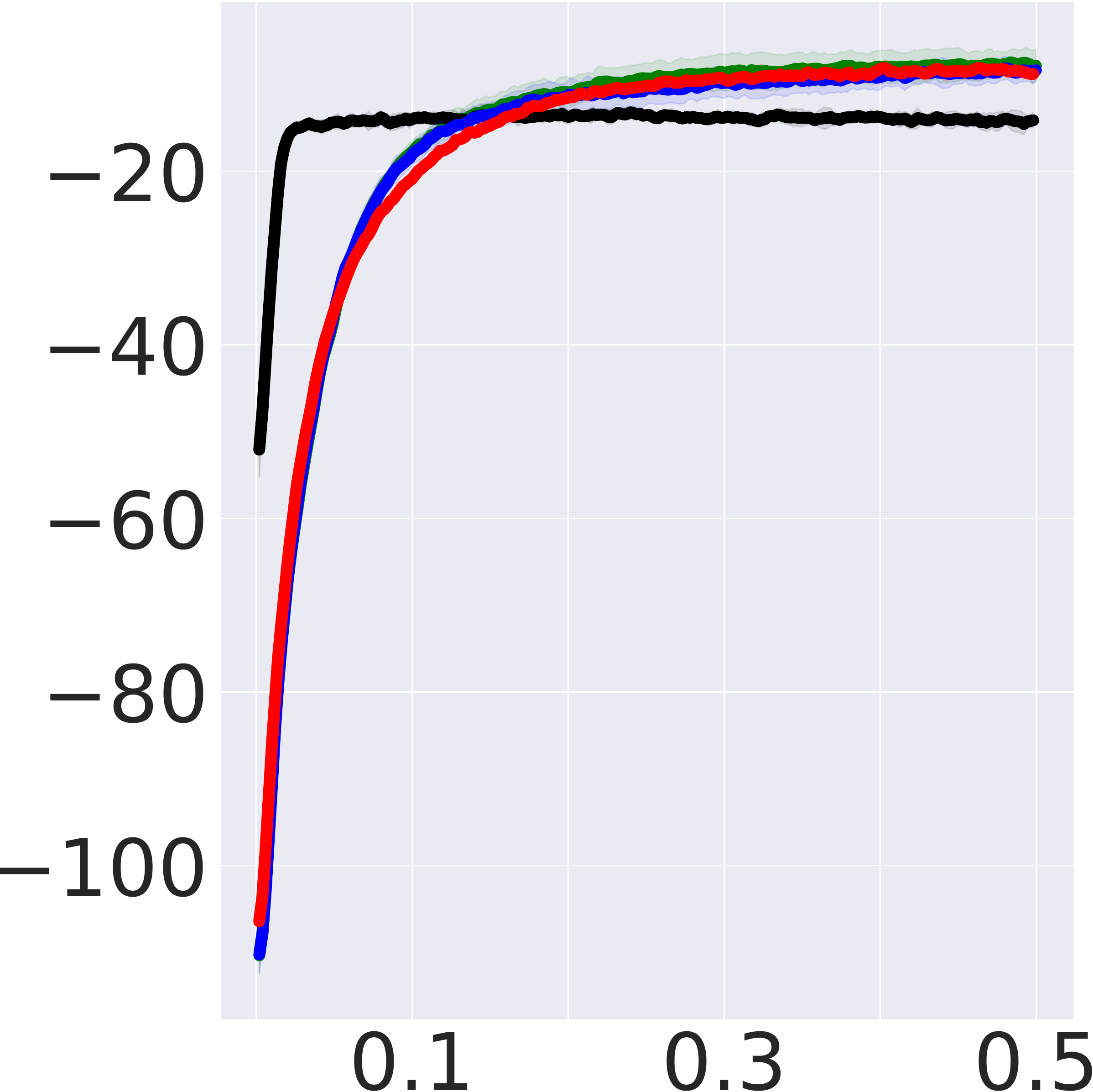}
    \vspace{4pt}
  \end{subfigure}%
  \begin{subfigure}[b]{0.2\textwidth}
    \centering
    \includegraphics[width=0.9\textwidth]{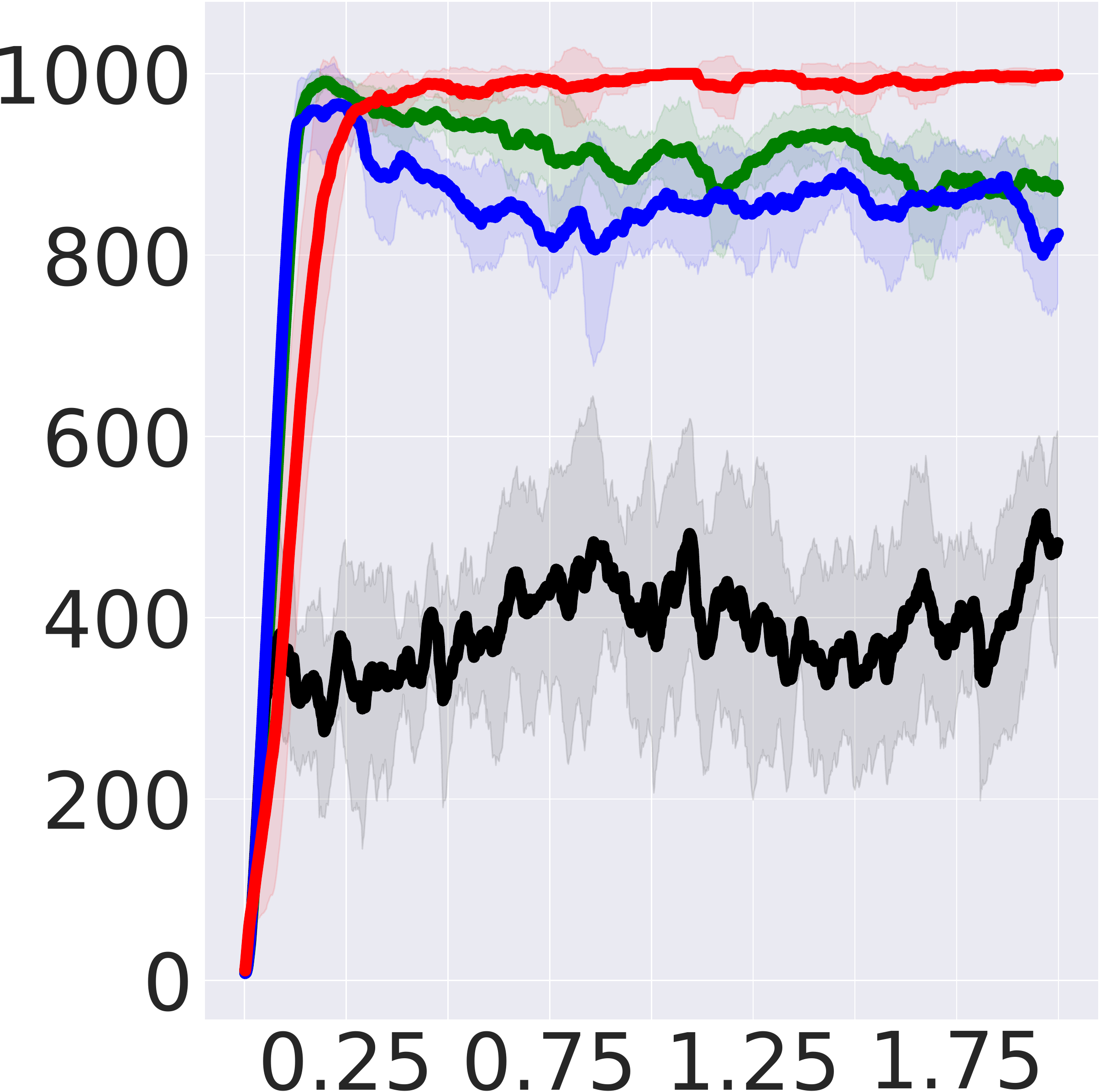}
    \vspace{4pt}
  \end{subfigure}%
  \begin{subfigure}[b]{0.2\textwidth}
    \centering
    \includegraphics[width=0.9\textwidth]{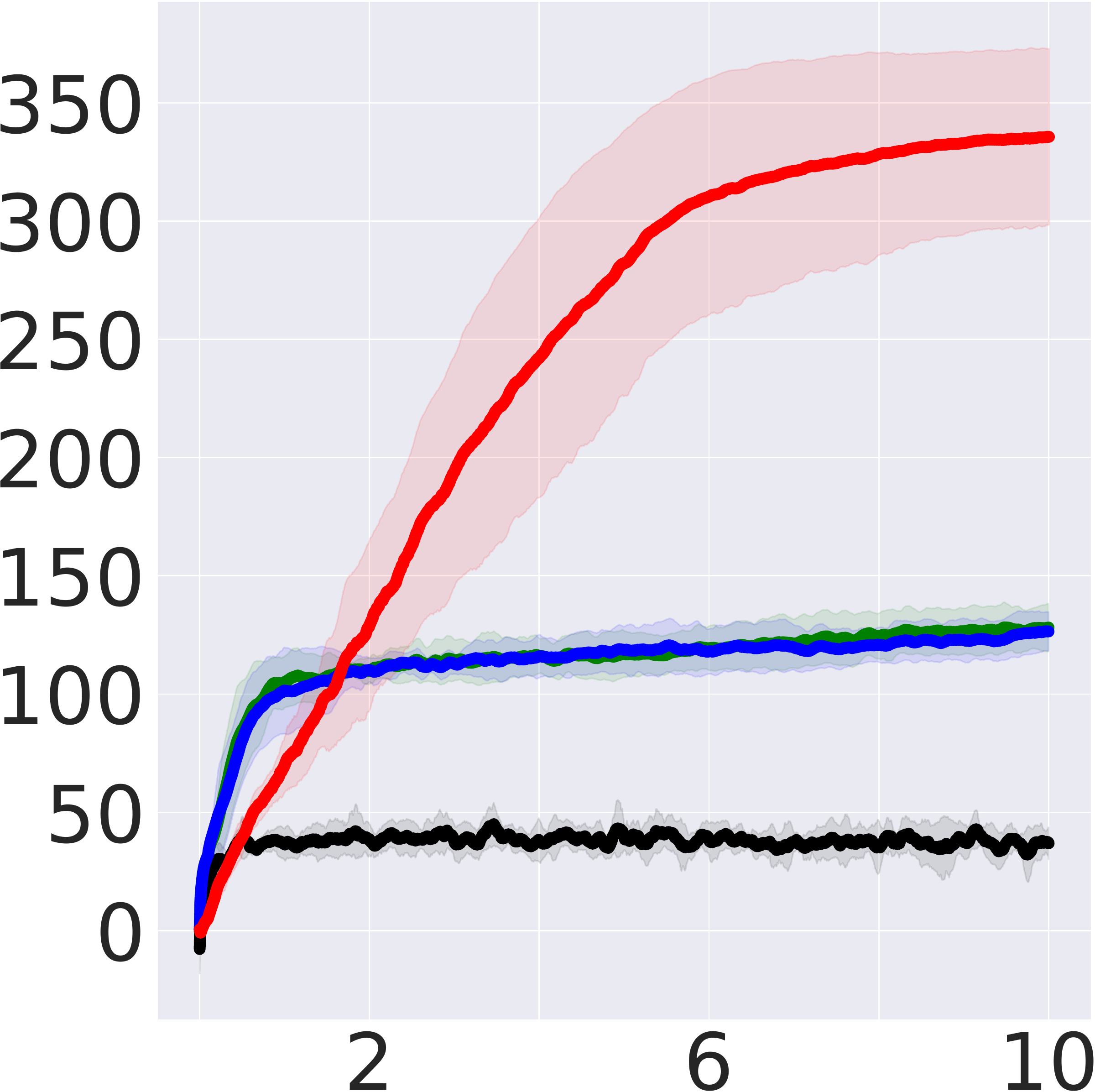}
    \vspace{4pt}
  \end{subfigure}%
  
    \begin{subfigure}[b]{0.2\textwidth}
    \centering
    \includegraphics[width=0.9\textwidth]{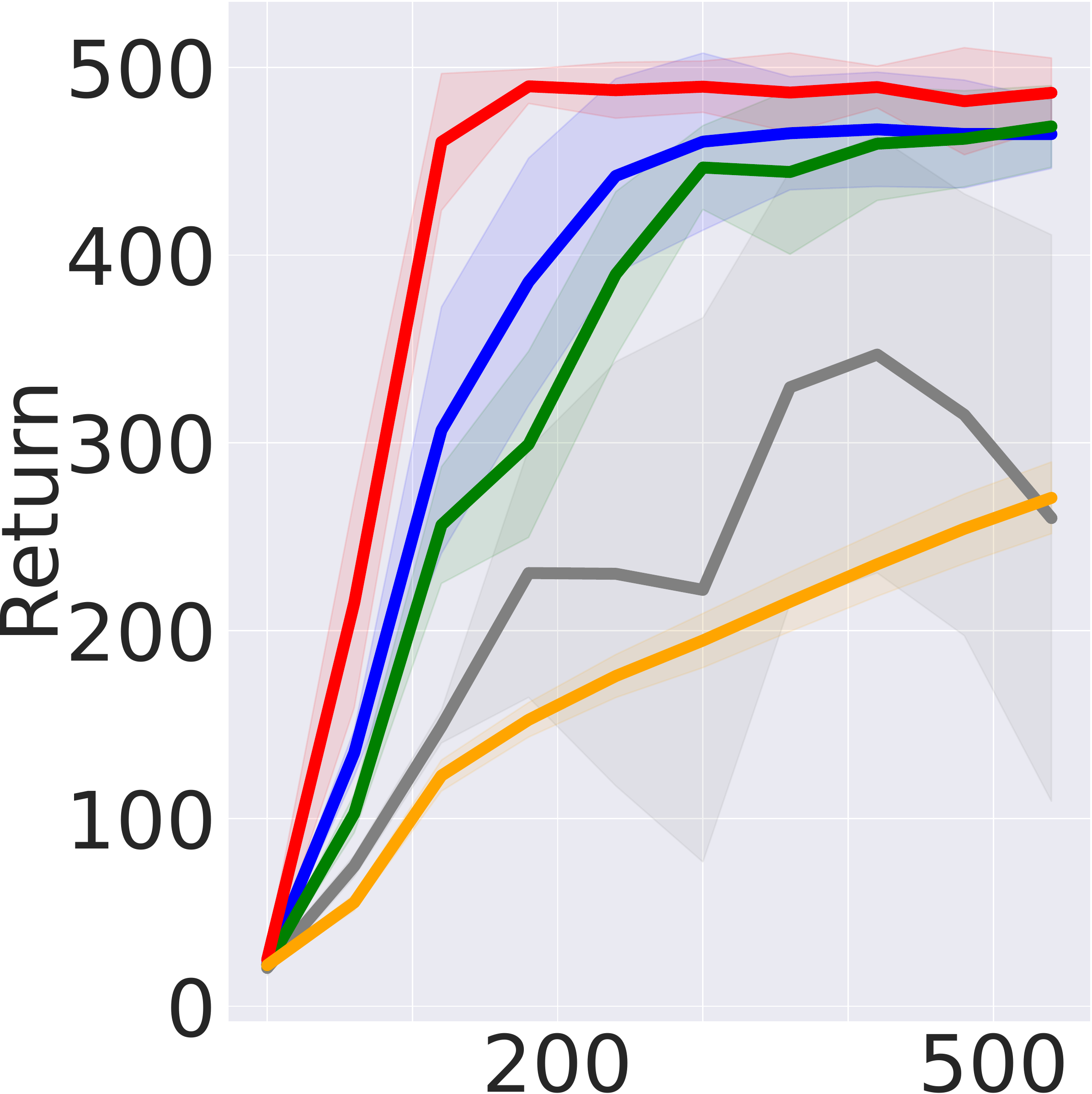}
    \caption{CartPole-v1}
    \label{fig:3-a}
  \end{subfigure}%
  \begin{subfigure}[b]{0.2\textwidth}
    \centering
    \includegraphics[width=0.9\textwidth]{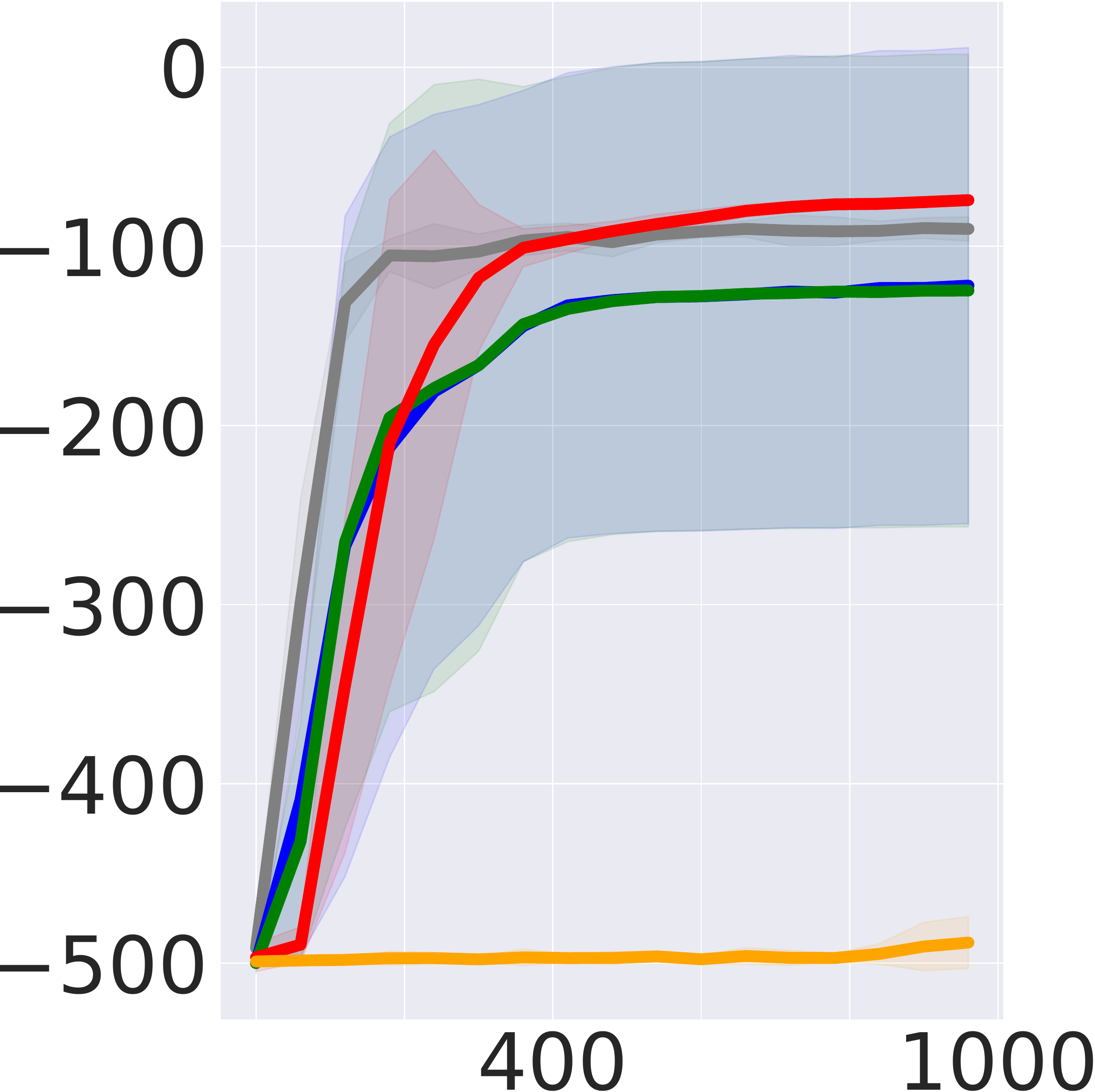}
    \caption{Acrobot-v1}
    \label{fig:3-b}
  \end{subfigure}%
  \begin{subfigure}[b]{0.2\textwidth}
    \centering
    \includegraphics[width=0.9\textwidth]{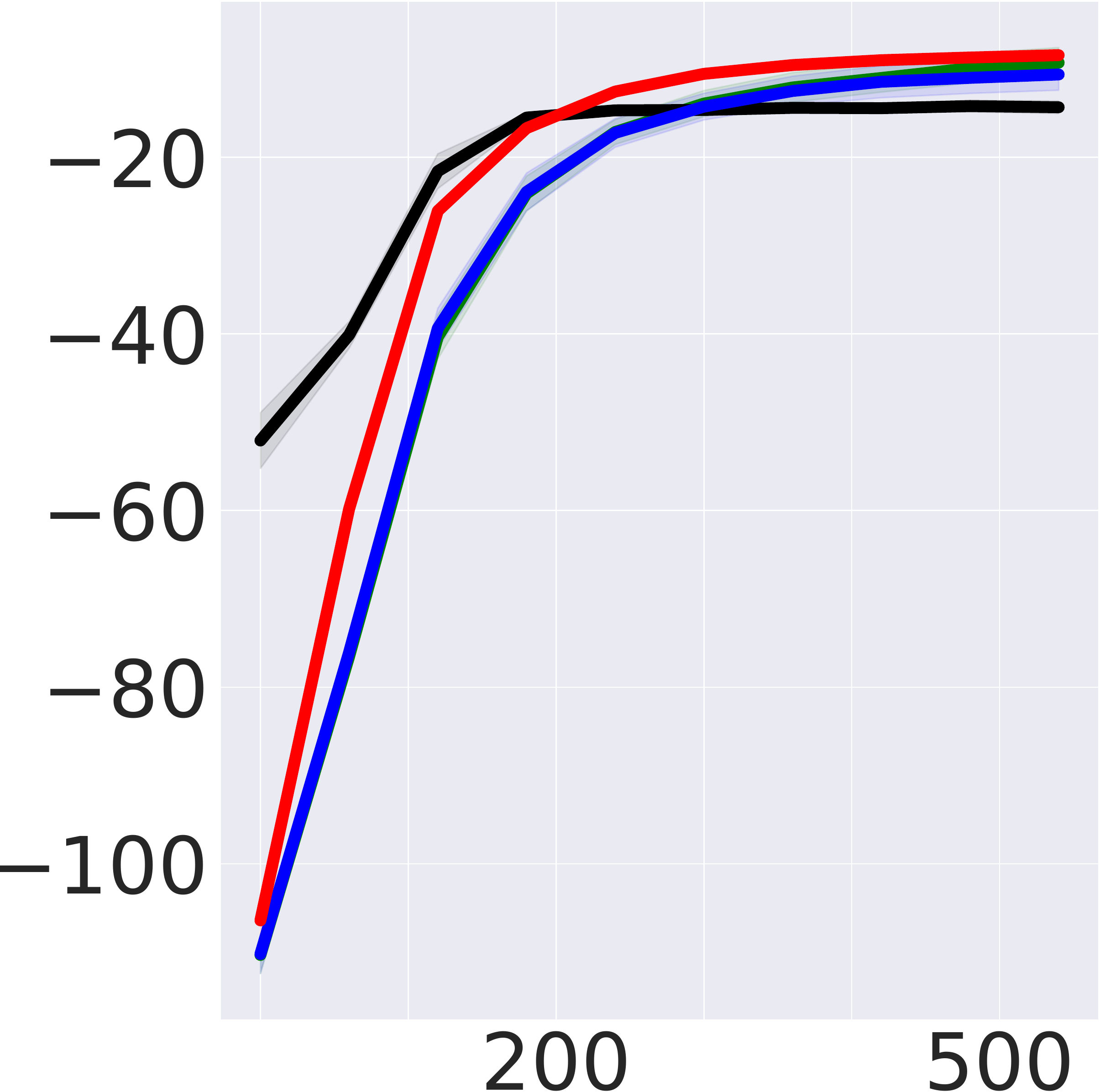}
    \caption{Reacher-v2}
    \label{fig:3-c}
  \end{subfigure}%
  \begin{subfigure}[b]{0.2\textwidth}
    \centering
    \includegraphics[width=0.9\textwidth]{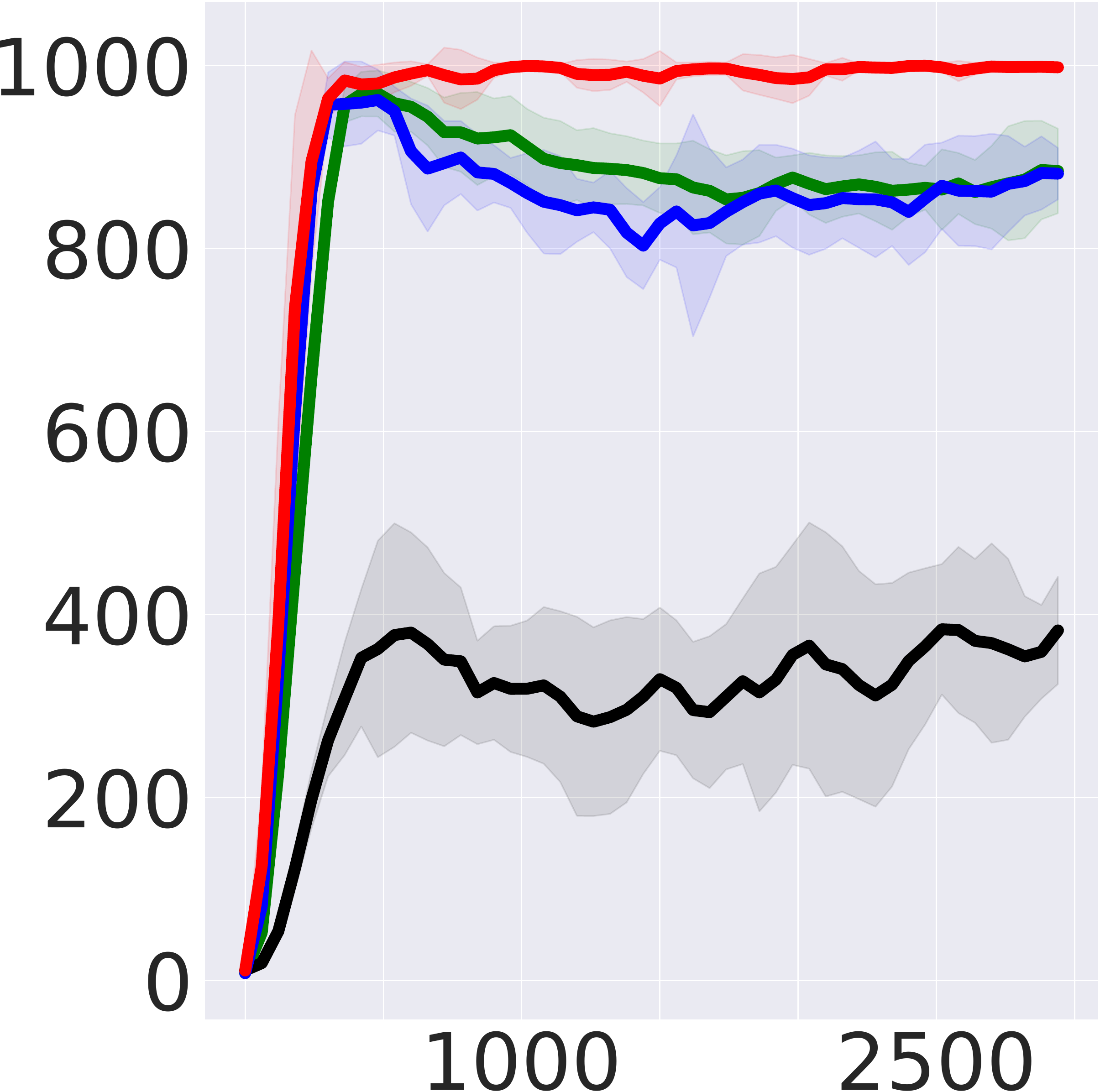}
    \caption{InvertedPendulum-v2}
    \label{fig:3-d}
  \end{subfigure}%
  \begin{subfigure}[b]{0.2\textwidth}
    \centering
    \includegraphics[width=0.9\textwidth]{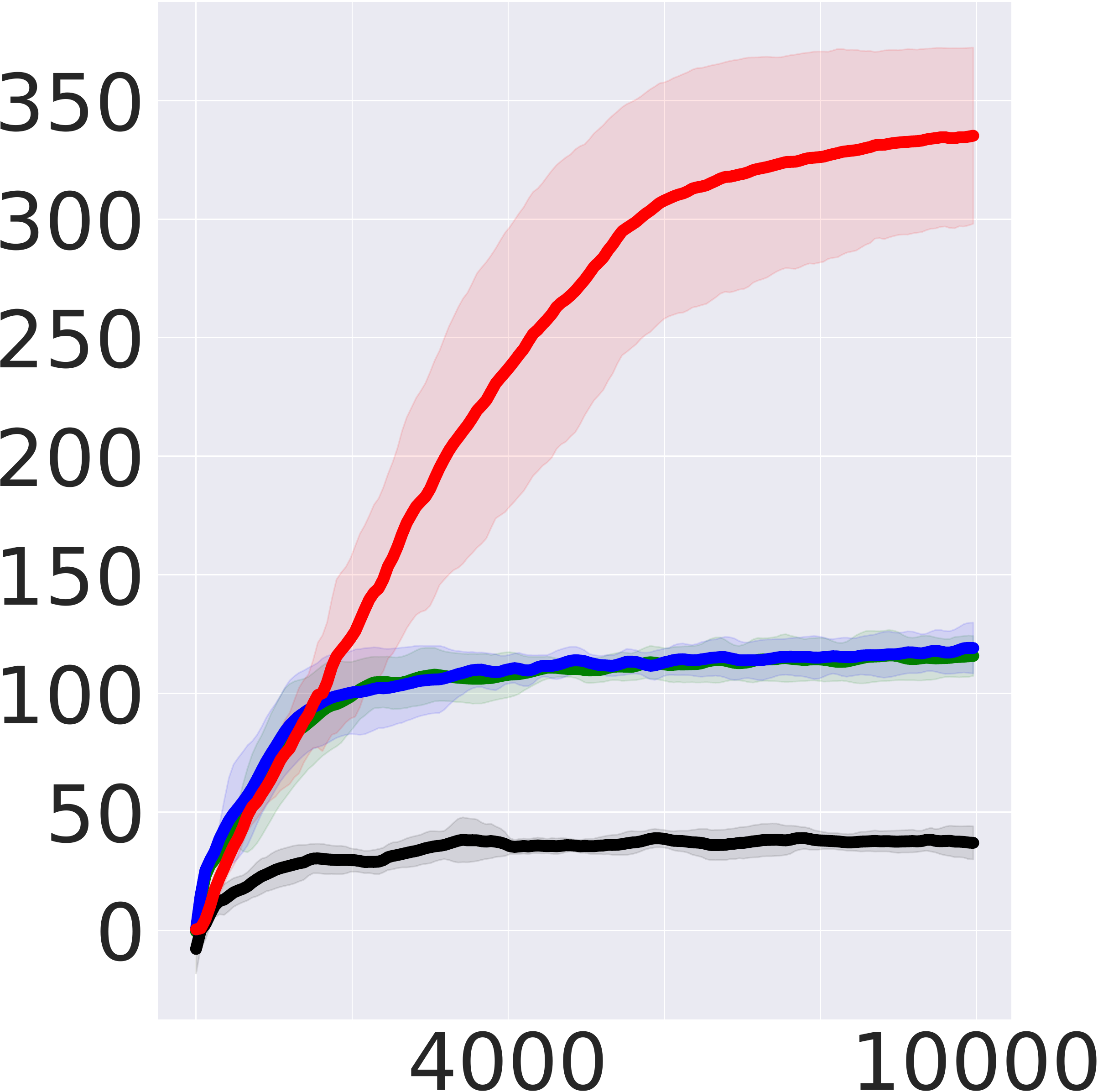}
    \caption{Swimmer-v2}
    \label{fig:3-e}
  \end{subfigure}%
  \caption{Performance on discrete and continuous control tasks. All the experiments are run $10$ times with seeds $0$ to $9$. The shaded area represents mean $\pm$ standard deviation. We show the returns with respect to timesteps~(top row) and running time~(bottom row). For a fair comparison of running time, we run a single process for each of the experiments. The x-axis in the top row denotes millions of timesteps. The x-axis in the bottom row denotes the running time in seconds.}
  \label{fig:3}
\end{figure*}


\subsection{Simulated Control Tasks}
\label{exp:2}
To study \textbf{Q2}, we evaluate SSRL on a few benchmark discrete and continuous control tasks in OpenAI Gym\footnote{\url{https://gym.openai.com/}}. We compare SSRL against the state-of-the-art methods that use policy gradients and/or value functions, i.e., Self-Imitation Learning~(SIL)~\cite{oh2018self}, PPO~\cite{schulman2017proximal}, DDPG~\cite{lillicrap2016continuous} and DQN~\cite{mnih2015human}. Note that SIL also imitates agents' own good experiences but still relies on policy gradients and value functions. We further include Upside Down Reinforcement Learning~(UDRL)~\cite{srivastava2019training} which also adopts a ranking buffer and a supervised learning loss, but requires an additional command input. We use the SIL code\footnote{\url{https://github.com/junhyukoh/self-imitation-learning}} and UDRL code\footnote{\url{https://github.com/BY571/Upside-Down-Reinforcement-Learning}} by their authors. For other baselines, we use the implementations in OpenAI baselines\footnote{\url{https://github.com/openai/baselines}}. We conduct hyperparameters search for both SSRL and the baselines.

We first evaluate the sample efficiency of SSRL. We report the average return over recent $100$ episodes with respect to the number of interaction timesteps in the top row of Figure~\ref{fig:3}. The results show that SSRL can successfully solve various control tasks and is competitive with state-of-the-art value-based methods. An interesting observation is that, on Swimmer-v2, our simple algorithm significantly beats SIL, PPO and DDPG. Specifically, SSRL achieves an average reward of around $350$ while PPO and SIL are stuck at around $130$ and DDPG is stuck at around $50$. In other tasks, SSRL also delivers competitive or better performance than the baselines. Therefore, SSRL is a promising alternative to solve reinforcement learning tasks.

When looking into the running time in the bottom row of Figure~\ref{fig:3}, we observe that our SSRL converges much faster than the baselines on almost all tasks. It is expected since the implementation of SSRL is very simple with just supervised learning update, and SSRL only uses a very small buffer, without value estimation, advantage computation or the use of target networks.

We also observe that the learning curve of SSRL is smoother and more stable than the baselines. For example, in CartPole-v1, Arcrobot-v1 and InvertedPendulum-v2 the learning curves of SSRL have much smaller standard errors than PPO and SIL. Moreover, SSRL achieves almost monotonous improvement and is stable after converged, whereas value-based or policy-based methods suffer from significant performance drop on some of the environments. We conjecture that the stability of SSRL may be inherited from supervised learning which is a more stable way to train deep neural networks.

\subsection{A Hundred Seeds Evaluation}
\label{exp:3}
\begin{figure*}[t]
  \centering
    \begin{subfigure}[b]{0.2\textwidth}
    \centering
    \includegraphics[width=0.9\textwidth]{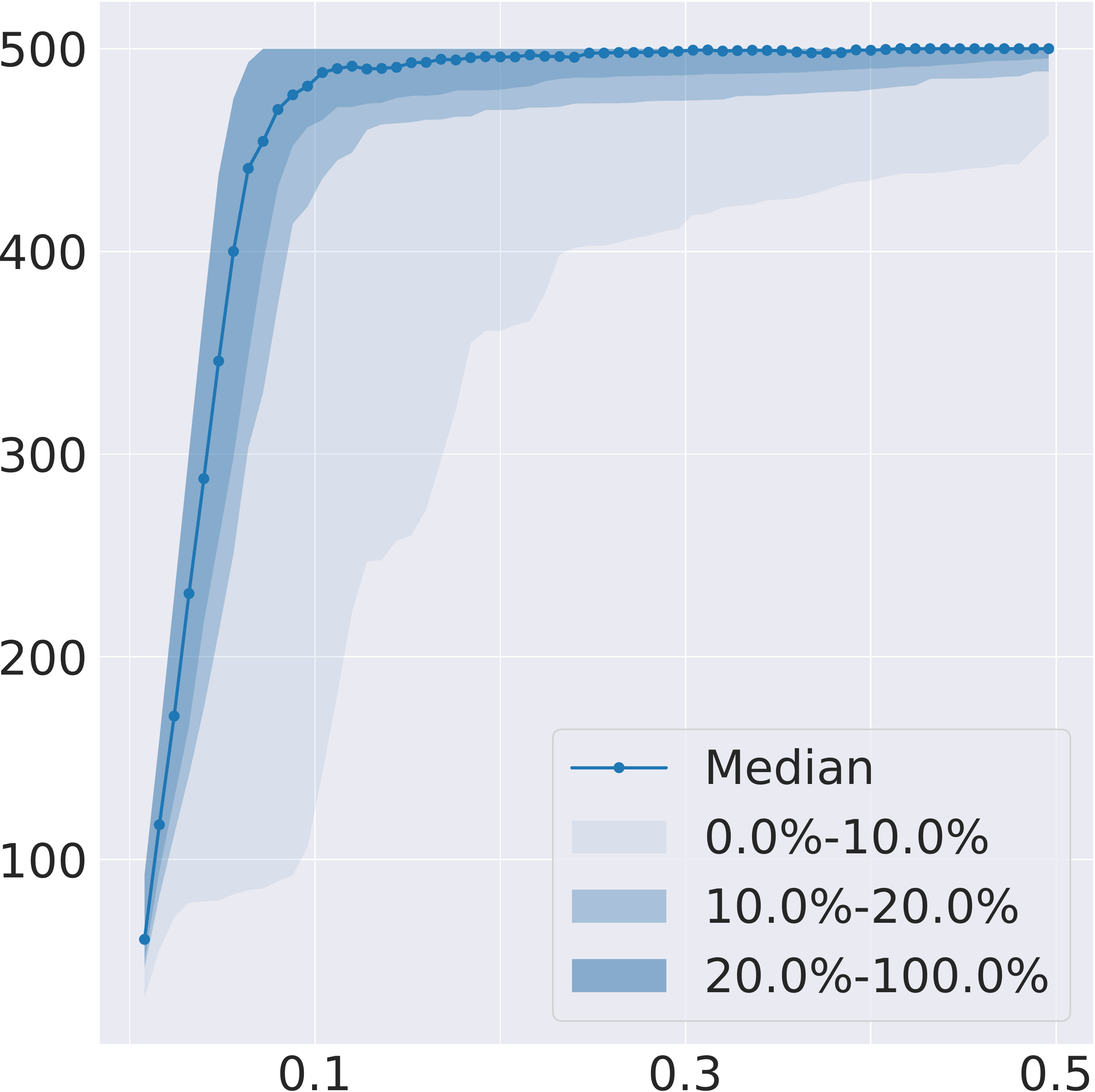}
    \caption{CartPole-v1}
    \label{fig:4-a}
  \end{subfigure}%
  \begin{subfigure}[b]{0.2\textwidth}
    \centering
    \includegraphics[width=0.9\textwidth]{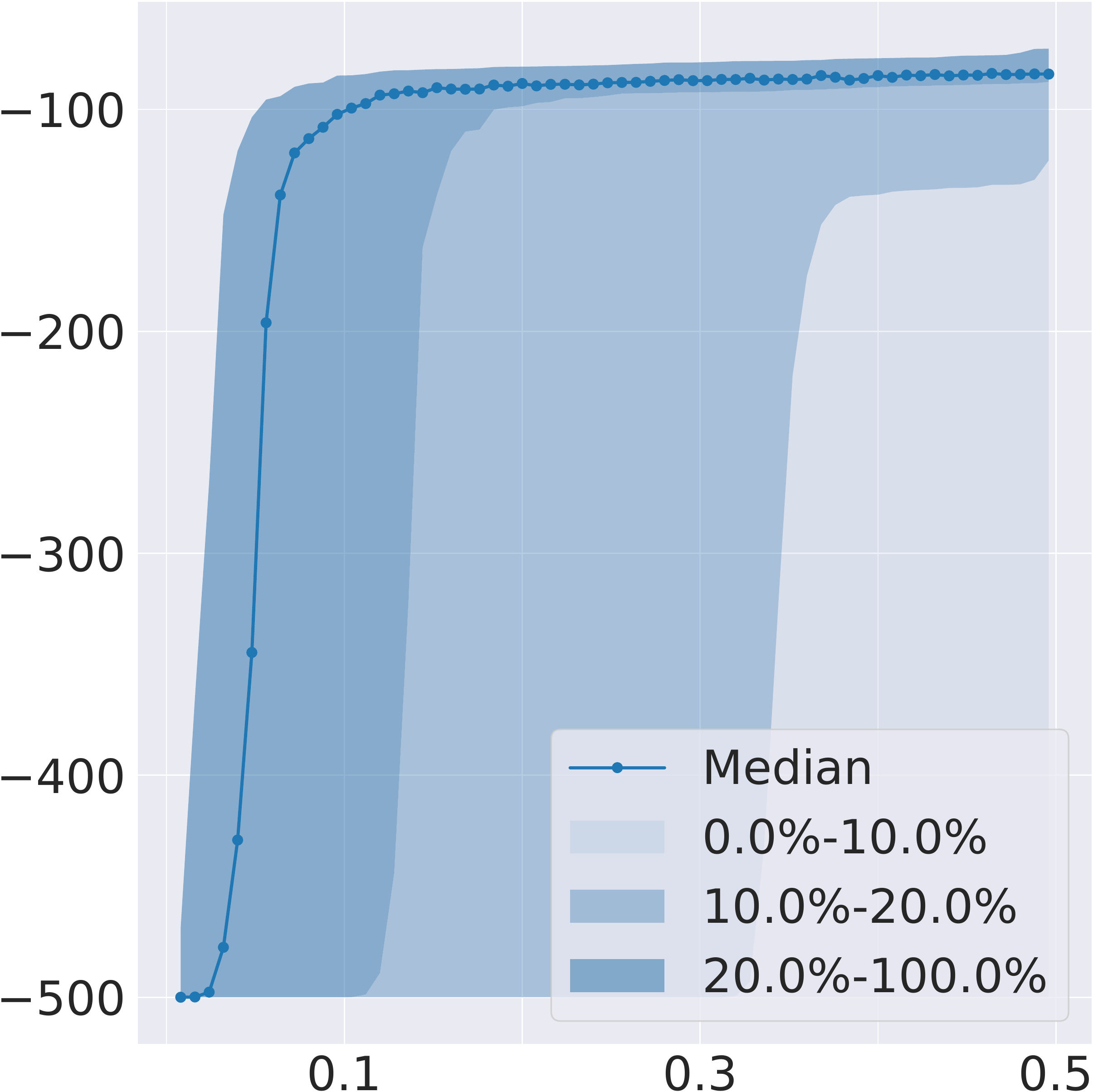}
    \caption{Acrobot-v1}
    \label{fig:4-b}
  \end{subfigure}%
  \begin{subfigure}[b]{0.2\textwidth}
    \centering
    \includegraphics[width=0.9\textwidth]{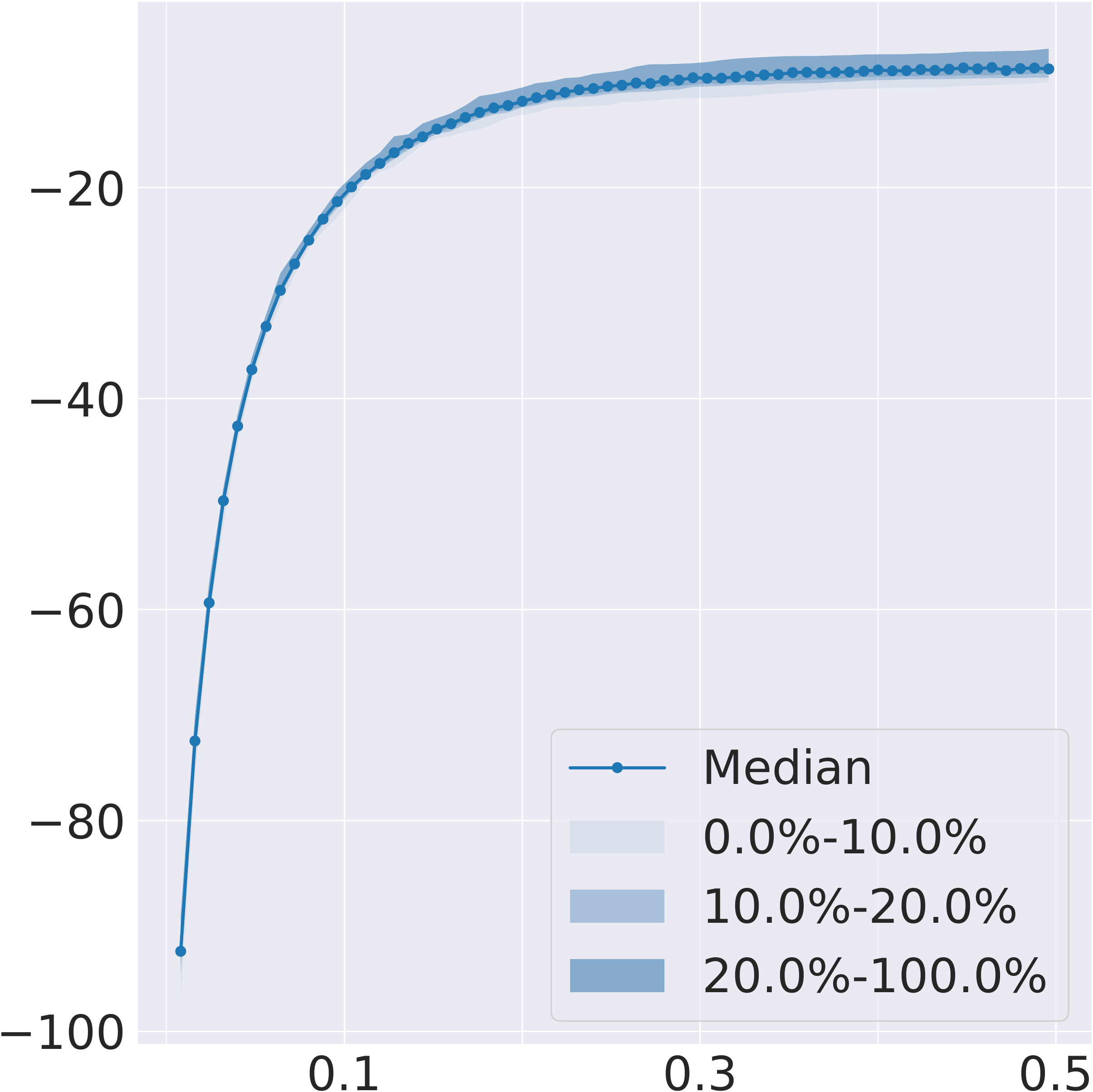}
    \caption{Reacher-v2}
    \label{fig:4-c}
  \end{subfigure}%
  \begin{subfigure}[b]{0.2\textwidth}
    \centering
    \includegraphics[width=0.9\textwidth]{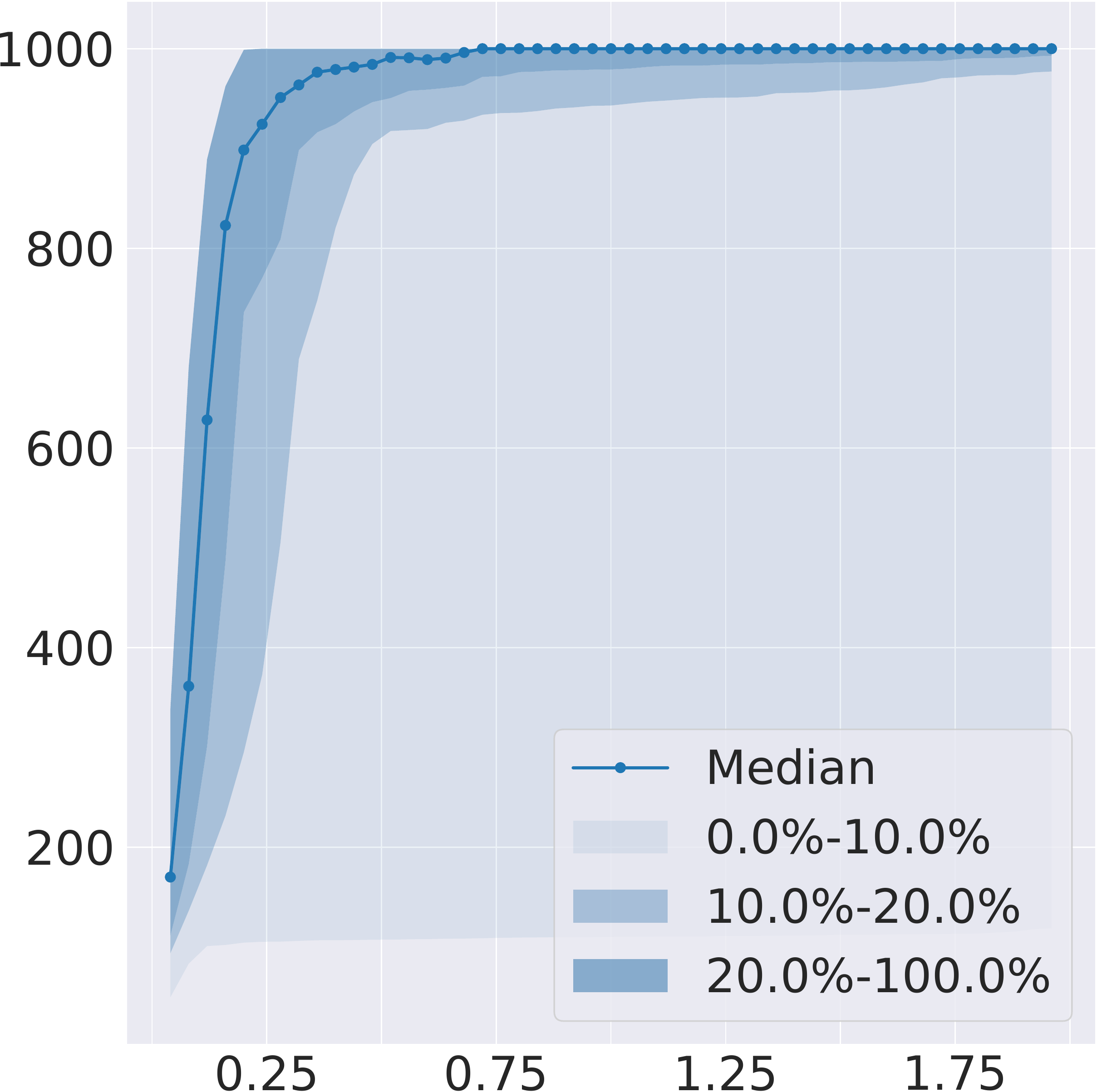}
    \caption{InvertedPendulum-v2}
    \label{fig:4-d}
  \end{subfigure}%
  \begin{subfigure}[b]{0.2\textwidth}
    \centering
    \includegraphics[width=0.9\textwidth]{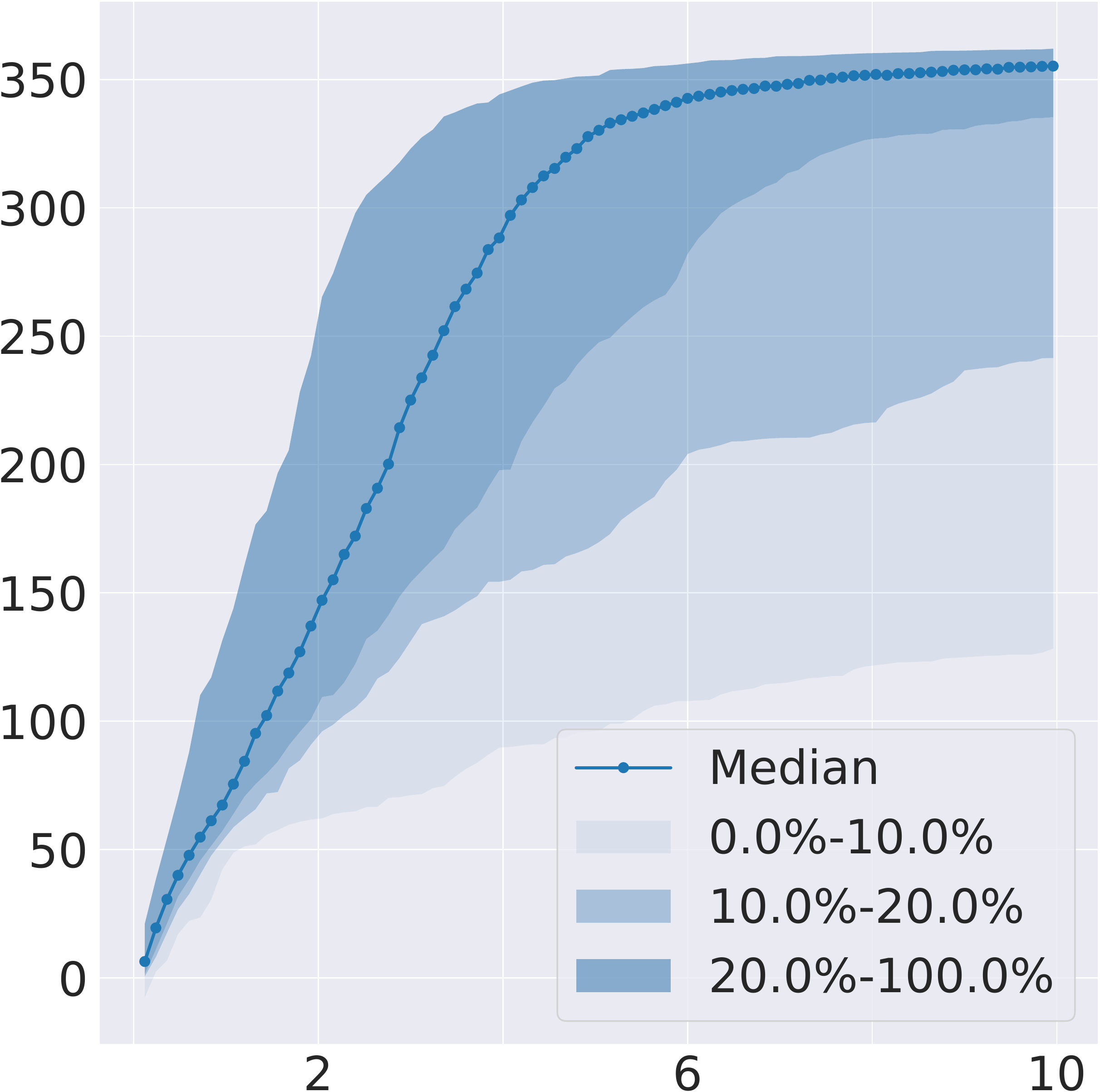}
    \caption{Swimmer-v2}
    \label{fig:4-e}
  \end{subfigure}%
  \caption{A hundred seeds evaluation. All the experiments are run a hundred times with random seeds $0$ to $99$. The dotted line represents the mean performance, and the shaded areas represent the percentage. The x-axis denotes millions of timesteps.}
  \label{fig:4}
\end{figure*}

Deep reinforcement learning algorithms are notoriously unstable and sensitive to different random seeds~\cite{henderson2018deep}. To address \textbf{Q3}, we evaluate SSRL with $100$ different random seeds to rigorously evaluate its stability. The performance distributions and median performance are reported in Figure~\ref{fig:4}. As expected, different random seeds lead to diverse performance. Nonetheless, SSRL succeeds in training policies with a large fraction of seeds. We observe consistent performance with at least 80\% of the seeds.

\subsection{Playing Games from Raw Image Pixels}
\label{exp:4}
Learning to control directly from high-dimensional pixel inputs has been a very challenging task before DQN~\cite{mnih2015human}. To better understand the capability of SSRL, we study \textbf{Q4} by developing a variant of SSRL and test it on the Atari Pong game. Specifically, to accelerate the training, we distribute the learning process with multiple actors who are responsible for collecting data from the environment, and multiple workers who are responsible for sampling data and updating the model~(see right-hand side of Figure~\ref{fig:overview}). Besides, we use a large buffer with a size of $10^6$ because we need more data to train the model from high-dimensional pixels. Since ranking the experiences in a large buffer is time-consuming, we instead directly feed winning trajectories into a ring buffer without further ranking. We compare SSRL with several state-of-the-art methods, including Rainbow~\cite{hessel2018rainbow}, IQN~\cite{dabney2018implicit}, C51~\cite{bellemare2017distributional}, DQN~\cite{mnih2015human}, A2C+SIL~\cite{oh2018self} and A2C~\cite{mnih2016asynchronous}. For A2C, we use the implementation in OpenAI baselines. For other algorithms, we use the implementations in Dopamine\footnote{\url{https://github.com/google/dopamine}}~\cite{castro2018dopamine}.

\begin{figure}[t]
  \centering
  \begin{subfigure}[b]{0.17\textwidth}
    \centering
    \includegraphics[width=0.9\textwidth]{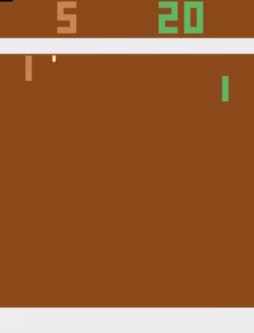}
  \end{subfigure}%
  \begin{subfigure}[b]{0.30\textwidth}
    \centering
    \includegraphics[width=0.9\textwidth]{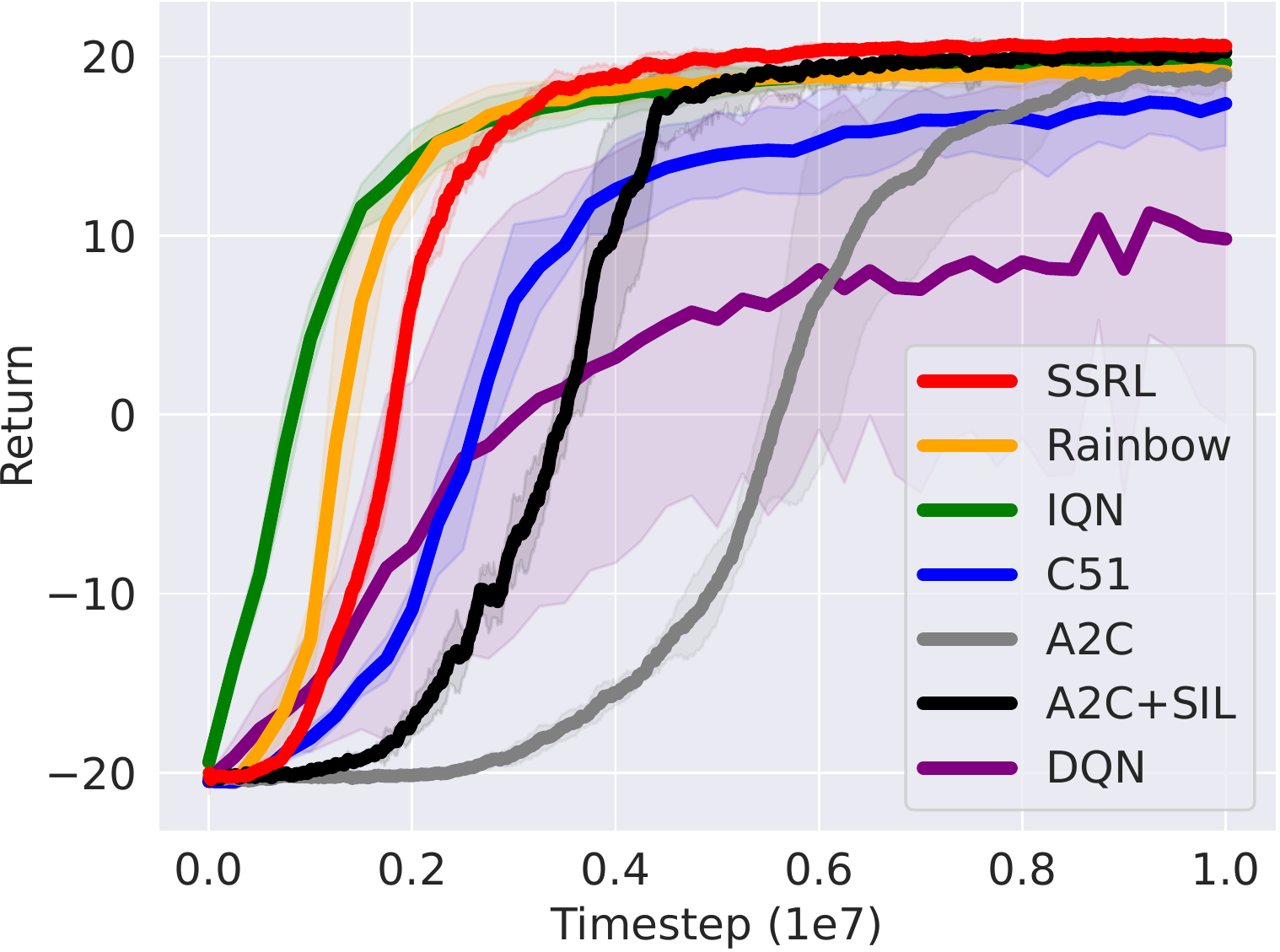}
  \end{subfigure}%
  \caption{Performance comparison of SSRL against the state-of-the-arts on Atari Pong. The results are averaged over $5$ different seeds. The shaded area represents mean $\pm$ standard deviation.}
  \label{fig:5}
  \vspace{-9pt}
\end{figure}

The results in Figure~\ref{fig:5} show that the proposed SSRL successfully solves the Pong game from raw image pixels. It provides competitive sample efficiency as commonly used policy-based and value-based methods. We note that Rainbow has incorporated several advances into DQN, such as prioritized experience replay~\cite{schaul2015prioritized} and Double DQN~\cite{van2016deep}, and IQN advances DQN with the consideration of risks, while SSRL remains simple. How we can further enhance SSRL and comparing it with the state-of-the-art reinforcement learning methods on the full Atari-57 benchmark will be our future work.

\subsection{Combining with Exploration Strategies on Hard Exploration Domains}
\label{exp:5}
Efficient exploration is one of the fundamental challenges in reinforcement learning~\cite{thrun1992efficient}. Contemporary exploration strategies are usually coupled with value-based reinforcement learning algorithms~\cite{pathak2017curiosity}. To investigate \textbf{Q5}, we study whether existing exploration strategies can be applied in SSRL. We focus on the count-based method~\cite{bellemare2016unifying}, which is a commonly used baseline in the literature. Specifically, count-based exploration gives an additional bonus of $\beta \sqrt{N(s)}$ to encourage the behavior of visiting new states, where $N(s)$ is the visiting frequency of state $s$ and $\beta$ is the hyperparameter.

\begin{figure}[t]
  \centering
  \begin{subfigure}[b]{0.13\textwidth}
    \centering
    \includegraphics[width=0.9\textwidth]{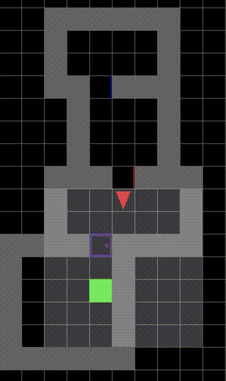}
  \end{subfigure}%
  \begin{subfigure}[b]{0.30\textwidth}
    \centering
    \includegraphics[width=0.9\textwidth]{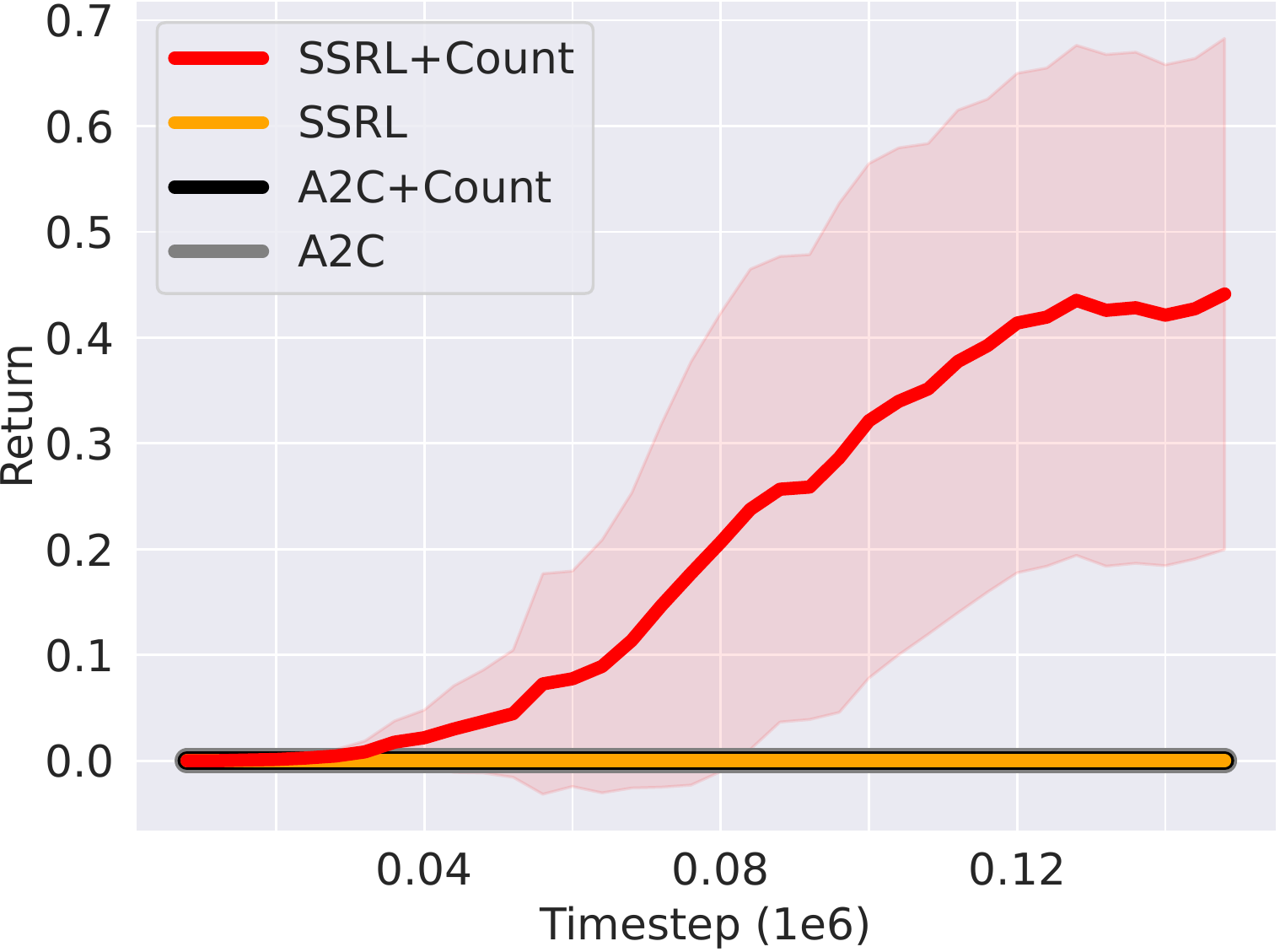}
  \end{subfigure}%
  \caption{The performance of SSRL and A2C with count-based exploration on a hard exploration domain using seeds $0$ to $2$. The shaded area represents mean $\pm$ standard deviation. Only SSRL with count-based exploration successfully navigates the goal.}
  \label{fig:6}
\end{figure}

We conduct experiments on MiniGrid-MultiRoom-N4-S5-v0\footnote{\url{https://github.com/maximecb/gym-minigrid}}~\cite{gym_minigrid}, illustrated in Figure~\ref{fig:6}. In each episode, four rooms are randomly generated with maximum size of $5$. At each step, the agent~(red triangle) observes the image pixels in front of it and can select one of the $8$ actions, such as turn left, turn right and open the door. A sparse reward will be given if the agent reaches the goal~(green), with a small penalty of the number of steps taken. This multi-room environment is extremely difficult to solve using reinforcement learning algorithms alone due to the sparse rewards.

Figure~\ref{fig:6} shows the performance of SSRL with or without count-based exploration. We also include A2C~\cite{mnih2016asynchronous} and A2C with count-based exploration for comparison. We observe that only SSRL with count-based exploration successfully navigates the goal. A possible explanation is that the ranking buffer of SSRL can better capture the positive reward signals. The results suggest that SSRL can be naturally combined with exploration strategies.

\section{Related Work}

\textbf{Self-Imitation Learning.} Self-Imitation Learning~\cite{oh2018self} proposes to imitate the agents' own good experiences to enhance the exploration for actor-critic methods~\cite{oh2018self}. Similarly, Ranking Policy Gradient~\cite{lin2019ranking} uses supervised learning as a separate module to achieve faster convergence for policy gradient. Dual Policy Distillation~\cite{lai2020dual} instantiates a peer agent and performs imitation learning with each other to improve sample efficiency. Upside Down Reinforcement Learning~\cite{srivastava2019training} also adopts supervised learning losses, whose objective is to learn a mapping of states and commands to actions. However, the learned policy is sensitive to the command~\cite{srivastava2019training}, which may be difficult to specify in practice. RAPID similarly uses self-imitation learning to encourage exploration by optimizing the coverage rate~\cite{zha2021rank}. While these studies shed light on the power of supervised learning in reinforcement learning, their understanding of supervised learning is limited to an enhancement for policy gradients~\cite{oh2018self,zha2021rank}, a module to achieve faster convergence~\cite{lin2019ranking,lai2020dual}, or a way to incorporate commands~\cite{srivastava2019training}. Our work pushes forward the understanding of self-imitation from a different perspective by describing the desired behaviors with data collection.

\textbf{Experience Replay Buffer} Experience replay mechanism is an important component to stabilize deep reinforcement learning by breaking the temporal correlations~\cite{lin1992self,lin1993reinforcement,mnih2015human}. While many studies have been focused on advancing experience replay~\cite{andrychowicz2017hindsight,pan2018organizing,zhang2017deeper,novati2018remember,zha2019experience}, the main goal of the prior work is to accelerate or stabilize the training of value functions. Although we use a replay buffer to store past experiences, we treat the past experiences as demonstrations and directly learn a policy function with supervised regression without the learning of value functions.

\textbf{Monte-Carlo Methods and Direct Policy Search.} Monte-Carlo (MC) methods are traditional reinforcement learning algorithms for episodic tasks~\cite{sutton2018reinforcement}. Recent work shows that Monte-Carlo methods can deliver strong results in large-scale card games with sufficient samples \cite{zha2019rlcard,zha2021douzero}. Unlike~\cite{zha2021douzero}, we introduce a ranking buffer to improve sample efficiency. Direct policy search also relies on Monte-Carlo simulation and episode-level reward, but treats the problem as a black-box optimization and directly search the policy in parameter space, such as Cross-Entropy Method~(CEM)~\cite{mannor2003cross}, random search~\cite{mania2018simple}, and evolution strategies~\cite{salimans2017evolution}. While these methods are shown to have a competitive performance on linear model~\cite{mania2018simple}, they are sample intensive when using deep models~\cite{salimans2017evolution}. While SSRL also guides the policy update with data collection, it uses gradient-based supervised loss to train the network, which can easily generalize to deep and complex network architectures without the cost of sample efficiency.



\textbf{Expectation-Maximization (EM) Methods} EM-based methods formulate reinforcement learning as a probabilistic inference problem and tackle it with Expectation-Maximization framework~\cite{dayan1997using,peters2007reinforcement,kober2009policy,levine2013variational,abdolmaleki2018maximum}. These methods decompose the probability of optimally into a variational distribution and the KL divergence between the policy distribution and the variational distribution. Then the variational distribution and the policy are iteratively optimized in an on-policy manner. While SSRL shares a similar idea of optimizing the policy network with regression, SSRL iteratively performs rollouts and conducts regression in an off-policy way. Prior work usually requires unbiased Q-value estimation in Expectation step~\cite{kober2009policy,abdolmaleki2018maximum}, whereas SSRL directly conducts supervised learning using good demonstrations without the use of Q-function.

\section{Conclusions and Future Work}

This work presents a pilot study for a very simple idea that iteratively collects highly-rewarded episodes and imitates them with supervised losses. We instantiate this idea with a ranking algorithm, dubbed SSRL, to select the top rewarded episodes greedily. Theoretically, we show that, under some conditions such as deterministic MDPs, SSRL can guarantee policy improvement. Empirically, we test SSRL across different tasks, including deterministic/non-deterministic MDPs, discrete/continuous action spaces, high-dimensional image inputs, and hard exploration domains. SSRL delivers competitive results in terms of sample efficiency with more stable performance and much less running time, showing the potential of solving RL with purely supervised losses.


\textbf{Limitations and Future Work.} This work focuses on a very simple instance of SSRL. There are two limitations to applying this simple algorithm to large-scale, complex, and stochastic environments. First, SSRL relies on the agent's ability to discover good episodes in the first place, which could be problematic in sparse environments. Second, while our theoretical analysis gives positive results in deterministic MDPs, our assumptions may not hold in more complex and stochastic environments. We hope the insight that our simple algorithm can work well in many environments can motivate future research to investigate SSRL in more complex environments, theoretically and empirically.

\bibliography{ref}

\appendix

\section{Appendix}
\subsection{Proof to Theorem 1}
We give proof to Theorem~\ref{theorem:1} in the main paper. To be self-contained, we first list the notations as follows.

For convenience, we use notation $C$ to count the number of transitions in $\bm{\tau}$. Specifically, we define $C(\bm{\tau}, s, a ,s', t)$ as the number transitions in $\bm{\tau}$ that take action $a$ in state $s$ and transit to state $s'$ at timestep $t$. In what follows, we abuse the notations to represent summarizations of the counts. For example, we use $C(\bm{\tau}, s, a ,\cdot, t)$ to represent the number of transitions that take action $a$ in state $s$ at timestep $t$, i.e., $C(\bm{\tau}, s, a ,\cdot, t) = \sum_{s'} C(\bm{\tau}, s, a ,s', t)$, and we use $C(\bm{\tau}, s,\cdot ,\cdot, t)$ to represent the number of transitions in state $s$ at timestep $t$, i.e., $C(\bm{\tau}, s,\cdot ,\cdot, t) = \sum_{s'} \sum_a C(\bm{\tau}, s, a ,s', t)$. The uniformly-distributed condition is defined as follows.

\begin{definition}[Uniformly-Distributed]
\label{def:1}
Trajectories $\bm{\tau}$ are uniformly-distributed if the initial states frequencies are consistent with the initial states distribution of the environment, i.e, $\forall s, p_0 (s) = \frac{C(\bm{\tau}, s, \cdot, \cdot, 0)}{|\bm{\tau}|}$, the state transition frequencies for all timesteps are consistent with the state transition probabilities of the environment, i.e.,$\forall s, a, s', t, \mathcal{P}(s' | s, a) = \frac{C(\bm{\tau}, s, a, s', t)}{C(\bm{\tau}, s, a, \cdot, t)}$.
\end{definition}

In deterministic MDPs, the initial state distribution and the transition will be fixed. Thus, uniformly-distributed condition will perfectly hold. In the rest of the proof, we assume trajectory and episode are equivalent, i.e., a trajectory represents the state-action pairs in a whole episode. We will use the following definitions to describe the reward of trajectories and the improvement of trajectories.

\begin{definition}[Trajectory Discounted Cumulative Reward]
\label{def:2}
For a trajectory $\tau = \{ \langle s_t, a_t, r_t \rangle \}_{t=0}^{T}$, the trajectory discounted cumulative reward is defined as $R(\tau) = \sum_{t=0}^T \gamma^t r_t$, where $T$ is the terminal timestep.
\end{definition}
\begin{definition}[Expected Discounted Cumulative Reward]
\label{def:3}
For a policy $\pi$, the expected discounted cumulative reward is defined as $R(\pi) = \mathbb{E}_{\pi}[\sum_{t=0}^T \gamma^t r_t]$, where T is the terminal timestep.
\end{definition}
\begin{definition}[Trajectories Improvement]
\label{def:4}
Trajectories improvement $\delta(\bm{\tau})$ of $\bm{\tau} = \{\bm{\tau}_i\}^N_{i=1}$ is defined as the difference between mean trajectory discounted cumulative reward and the expected discounted cumulative reward of the current policy, i.e., $\delta(\bm{\tau}) = \sum_i \frac{R(\bm{\tau}_i)}{|\bm{\tau}|} - R(\pi)$
\end{definition}

Our proof is built upon hypothetical policy, which is defined as follows.

\begin{definition} [Hypothetical Policy]
\label{def:5}
Given $\bm{\tau}$, the hypothetical policy is defined as 
\begin{equation}
\label{eqn:1}
    \widetilde{\pi}(a|s) = \frac{C(\bm{\tau}, s, a, \cdot, \cdot)}{C(\bm{\tau}, s, \cdot, \cdot, \cdot)}
\end{equation}
\end{definition}

We will use the notations of discounted visitation frequencies as defined in~\cite{schulman2015trust}.
\begin{equation}
\label{eqn:2}
    \rho_{\pi}(s) = \sum_{t=0}^{T} \gamma^t p_{\pi}(s_t=s),
\end{equation}
where $p_{\pi}(s_t=s)$ is the probability of reaching state $s$ at timestep $t$. Note that $\sum_{s \in \mathcal{S}} p_\pi(s_t = s) = 1$. We can conveniently use discounted visiting frequencies to represent the performance of a policy.

Before proving Theorem~\ref{theorem:1}, we prove the following lemma.

\begin{lemma} [Expected Performance]
\label{lam:1}
The expected discounted cumulative reward of policy $\pi$ is weighted sum of $r(s)$, determined by $\rho_\pi(s)$, i.e., $R(\pi) = \sum_{s \in \mathcal{S}} \rho_\pi(s)r(s)$.
\end{lemma}
\textbf{Proof:} Based on Definition~\ref{def:3}, $R(\pi)$ can be rewritten as
\begin{align}
R(\pi) & = \mathbb{E}_\pi [\sum_{t=0}^{T} \gamma^t r_t] \label{eqn:3} \\
& = \sum_{t=0}^{T} \gamma^t \mathbb{E}_\pi [r_t] \label{eqn:4} \\
& = \sum_{t=0}^{T} \gamma^t \sum_{s \in \mathcal{S}} p_\pi(s_t = s) r(s) \label{eqn:5} \\
& = \sum_{s \in \mathcal{S}} \sum_{t=0}^{T} \gamma^t  p_\pi(s_t = s) r(s) \label{eqn:6} \\
& = \sum_{s \in \mathcal{S}} \rho_\pi(s) r(s) \label{eqn:7}
\end{align}
Note that Equation~(\ref{eqn:7}) is based on definition of discounted visiting frequencies in Equation~(\ref{eqn:2}). Lemma~\ref{lam:1} follows.

Now we prove Theorem~\ref{theorem:1} as folows.

\begin{theorem}[Policy-Improvement]
\label{theorem:1}
If uniformly-distributed condition holds for the trajectories in $\mathcal{D}$, we can construct a hypothetical policy $\widetilde{\pi}$ whose expected cumulative rewards is at least as good as that of the current policy $\pi$, and the supervised learning step is equivalent to imitating this better hypothetical policy.
\end{theorem}

\textbf{Proof:} We make use of discounted visiting frequencies to represent expected discounted cumulative reward~(Lemma~\ref{lam:1}). Then we show that the expected discounted cumulative reward is equivalent to the mean trajectories cumulative reward.


\begin{table*}[t]
\centering
\caption{Hyperparameters of SSRL in simulated control tasks.}
\label{tbl:1}
\begin{tabular}{l|l}
\toprule
Hyperparameter & Values \\ \midrule
Buffer size   & Searched from $\{1000, 5000\}$ \\
Batch size   & $256$ \\
Learning rate   & Searched from $\{10^{-4}, 2.5 \times 10^{-4}, 5 \times 10^{-4}, 2.5 \times 10^{-4}, 10^{-3}\}$ \\
Rollout steps   & Searched from $\{1, 5\}$ \\
Training steps   & $5$ \\
\bottomrule
\end{tabular}
\end{table*}

\begin{table*}[t]
\centering
\caption{Hyperparameters of SSRL on each of the environments.}
\label{tbl:2}
\begin{tabular}{l|c|c|c|c|c|c}
\toprule
Environment & Random seeds & Buffer size & Batch size & Learning rate & Rollout steps & Training steps \\ \midrule
CartPole-v1   & $0$ to $9$ & 1000 & 256 & $10^{-3}$ & $1$ & $5$ \\
Acrobot-v1   & $0$ to $9$ & 1000 & 256 & $7.5 \times 10^{-4}$ & $5$ & $5$ \\
Reacher-v2   & $0$ to $9$ & 5000 & 256 & $10^{-3}$ & $5$ & $5$ \\
InvertedPendulum-v2   & $0$ to $9$ & 1000 & 256 & $2.5 \times 10^{-4}$ & $1$ & $5$ \\
Swimmer-v2   & $0$ to $9$ & 5000 & 256 & $5 \times 10^{-4}$ & $5$ & $5$ \\

\bottomrule
\end{tabular}
\end{table*}

\begin{table*}[t]
\centering
\caption{Hyperparameters of PPO and Self-Imitation Learning (SIL) in simulated control tasks. We follow the hyperparameters search strategy in SIL paper.}
\label{tbl:3}
\begin{tabular}{l|l}
\toprule
Hyperparameter & Values \\ \midrule
Learning rate   & Searched from $\{3 \times 10^{-4}, 10^{-4}, 5 \times 10^{-5}, 3 \times 10^{-5}\}$ \\
Horizon   & $2048$ \\
Number of epochs & $10$ \\
Minibatch size & $64$ \\
Discount factor $\gamma$ & $0.99$ \\
GAE parameter $\lambda$ & $0.99$ \\
Entropy regularization $\alpha$ & $0$ \\
\midrule
SIL update per batch & $10$ \\
SIL batch size & $512$ \\
SIL loss weight & $0.1$ \\
SIL value loss weight $\beta$ & Searched from $\{0.01, 0.05\}$ \\
Replay buffer size & $50000$ \\
Exponent for prioritization & Searched from $\{0.6, 1.0\}$ \\
Bias correction for prioritized replay & $0.1$ \\
\bottomrule
\end{tabular}
\end{table*}

Given $p_{\widetilde{\pi}}(s_t = s)$, $\forall s$, the visiting frequency of state $s$ at timestep $t+1$ can be derived by
\begin{align}
p_{\widetilde{\pi}}(s_{t+1} = s) & = \sum_{s' \in \mathcal{S}} p_{\widetilde{\pi}}(s_t = {s'}) \sum_{a \in \mathcal{A}} \widetilde{\pi}(a|s') \mathcal{P}(s|s', a) \label{eqn:8}\\
& = \sum_{s' \in \mathcal{S}} \sum_{a \in \mathcal{A}} p_{\widetilde{\pi}}(s_t = {s'}) \widetilde{\pi}(a|s') \mathcal{P}(s|s', a). \label{eqn:9}
\end{align}
Since the trajectories $\bm{\tau}$ in the buffer are uniformly distributed, we have
\begin{equation}
\label{eqn:10}
\mathcal{P}(s | s', a) = \frac{C(\bm{\tau}, s', a, s, t)}{C(\bm{\tau}, s', a, \cdot, t)}.
\end{equation}
Plugging Equation~(\ref{eqn:1}) and Equation~(\ref{eqn:10}) into Equation~(\ref{eqn:9}), we get
\begin{align}
& p_{\widetilde{\pi}}(s_{t+1} = s) \label{eqn:11} \\
 = & \sum_{s' \in \mathcal{S}} \sum_{a \in \mathcal{A}} p_{\widetilde{\pi}}(s_t = {s'}) \frac{C(\bm{\tau}, s', a, \cdot, \cdot)}{C(\bm{\tau}, s', \cdot, \cdot, \cdot)} \times \frac{C(\bm{\tau}, s', a, s, t)}{C(\bm{\tau}, s', a, \cdot, t)} \label{eqn:12}\\
 = & \sum_{s' \in \mathcal{S}} \sum_{a \in \mathcal{A}} p_{\widetilde{\pi}}(s_t = {s'}) \frac{C(\bm{\tau}, s', a, \cdot, t)}{C(\bm{\tau}, s', \cdot, \cdot, t)} \times \frac{C(\bm{\tau}, s', a, s, t)}{C(\bm{\tau}, s', a, \cdot, t)} \label{eqn:13}\\
 = & \sum_{s' \in \mathcal{S}} \sum_{a \in \mathcal{A}} p_{\widetilde{\pi}}(s_t = {s'}) \frac{C(\bm{\tau}, s', a, s, t)}{C(\bm{\tau}, s', \cdot, \cdot, t)} \label{eqn:14}\\
 = & \sum_{s' \in \mathcal{S}} p_{\widetilde{\pi}}(s_t = {s'}) \sum_{a \in \mathcal{A}} \frac{C(\bm{\tau}, s', a, s, t)}{C(\bm{\tau}, s', \cdot, \cdot, t)} \label{eqn:15}\\
 = & \sum_{s' \in \mathcal{S}} p_{\widetilde{\pi}}(s_t = {s'}) \frac{C(\bm{\tau}, s', \cdot, s, t)}{C(\bm{\tau}, s', \cdot, \cdot, t)}. \label{eqn:16}
\end{align}
Note that Equation~(\ref{eqn:13}) is based on the uniformly distributed assumption. Then we prove that $\forall s \in \mathcal{S}$, $\forall t \in \{0, 1, ..., T\}$, $p_{\widetilde{\pi}}(s_t = s) = \frac{C(\bm{\tau}, s, \cdot, \cdot, t)}{|\bm{\tau}|}$. When $t = 0$, we have $\forall s \in \mathcal{S}$, $p_{\widetilde{\pi}}(s_0 = s) = p_0(s) = \frac{C(\bm{\tau}, s, \cdot, \cdot, 0)}{|\bm{\tau}|}$, since $\bm{\tau}$ are uniformly distributed. Given $\forall s \in \mathcal{S}$, $p_{\widetilde{\pi}}(s_k = s) = \frac{C(\bm{\tau}, s, \cdot, \cdot, k)}{|\bm{\tau}|}$ where $k \ge 0$, when $t=k+1$ we have
\begin{align}
& p_{\widetilde{\pi}}(s_{k+1} = s)\label{eqn:17}  \\
 = & \sum_{s' \in \mathcal{S}} p_{\widetilde{\pi}}(s_k = {s'}) \frac{C(\bm{\tau}, s', \cdot, s, k)}{C(\bm{\tau}, s', \cdot, \cdot, k)} \label{eqn:18} \\
 = & \sum_{s' \in \mathcal{S}} \frac{C(\bm{\tau}, s', \cdot, \cdot, k)}{|\bm{\tau}|} \frac{C(\bm{\tau}, s', \cdot, s, k)}{C(\bm{\tau}, s', \cdot, \cdot, k)} \label{eqn:19} \\
 = & \sum_{s' \in \mathcal{S}} \frac{C(\bm{\tau}, s', \cdot, s, k)}{|\bm{\tau}|} \label{eqn:20} \\
 = & \frac{C(\bm{\tau}, \cdot, \cdot, s, k)}{|\bm{\tau}|} \label{eqn:21} \\
 = & \frac{C(\bm{\tau}, \cdot, \cdot, s, k)}{|\bm{\tau}|} \label{eqn:22} \\
 = & \frac{C(\bm{\tau}, s, \cdot, \cdot, k+1)}{|\bm{\tau}|} \label{eqn:23}.
\end{align}
Note that Equation~(\ref{eqn:23}) is based on the fact that the number of instances that transit to state $s$ at timestep $k$ is equivalent to the number of instances that are in state $s$ at timestep $k+1$. By induction, we can conclude that
\begin{equation}
\label{eqn:24}
    \forall s \in \mathcal{S}, \forall t \in \{0, 1, ..., T\}, p_{\widetilde{\pi}}(s_t = s) = \frac{C(\bm{\tau}, s, \cdot, \cdot, t)}{|\bm{\tau}|}.
\end{equation}
Based on Equation~(\ref{eqn:25}), the mean trajectory discounted cumulative reward can be rewritten as follows.
\begin{align}
& \sum_i \frac{R(\bm{\tau}_i)}{|\bm{\tau}|} \label{eqn:25}\\
 = & \frac{\sum_{t=0}^T \sum_{s \in \mathcal{S}} C(\bm{\tau}, s, \cdot, \cdot, t)\gamma^t r(s)}{|\bm{\tau}|} \label{eqn:26}\\
 = & \frac{\sum_{t=0}^T \sum_{s \in \mathcal{S}} C(\bm{\tau}, s, \cdot, \cdot, t)\gamma^t r(s)}{|\bm{\tau}|} \label{eqn:27}\\
 = & \sum_{t=0}^T \sum_{s \in \mathcal{S}} \gamma^t \frac{C(\bm{\tau}, s, \cdot, \cdot, t)}{|\bm{\tau}|} r(s) \label{eqn:28}\\
 = & \sum_{t=0}^T \sum_{s \in \mathcal{S}} \gamma^t p_{\widetilde{\pi}}(s_t = s) r(s) \label{eqn:29}\\
 = & \sum_{s \in \mathcal{S}} \sum_{t=0}^T  \gamma^t p_{\widetilde{\pi}}(s_t = s) r(s) \label{eqn:30}\\
 = & \sum_{s \in \mathcal{S}} \rho_{\widetilde{\pi}}(s) r(s) \label{eqn:31}.
\end{align}
Based on Lemma \ref{lam:1}, we have
\begin{equation}
    \label{eqn:32}
    \sum_i \frac{R(\bm{\tau}_i)}{|\bm{\tau}|} = R(\widetilde{\pi})
\end{equation}
Thus,
\begin{align}
& R(\widetilde{\pi}) - R(\pi) \label{eqn:33}\\
 = & \sum_i \frac{R(\bm{\tau}_i)}{|\bm{\tau}|} - R(\pi) \label{eqn:34} \\
 = & \delta(\bm{\tau}) \ge 0 \label{eqn:35}.
 \end{align}
Thus, by maximizing the trajectories improvement $\delta(\bm{\tau})$ through sampling better trajectories, we guarantee that we could construct an improved hypothetical policy $\widetilde{\pi}$. Imitating this improved hypothetical policy will naturally lead to policy improvement. If we use a neural network as the function approximator, then conducting supervised learning to $\bm{\tau}$ is essentially imitating the improved hypothetical policy since the action distributions of the hypothetical policy are determined by the action frequencies in the data. Therefore, the supervised learning step will lead to policy improvement by imitating a better hypothetical policy. Theorem~\ref{theorem:1} follows.

\subsection{Experimental Details}
In this section, we provide the details of all the experiments appeared in the paper\footnote{The codes will be released upon paper acceptance for reproducing the results.} and additional experimental results.

\subsubsection{Hardwares}
All the experiments are conducted on a server with 24 Intel(R) Xeon(R) Silver 4116 CPU @2.10GHz processors and 64.0 GB memory.

\subsubsection{Simulated Control Tasks}
\begin{figure*}
  \centering
  \begin{subfigure}[b]{0.2\textwidth}
    \centering
    \includegraphics[width=0.9\textwidth]{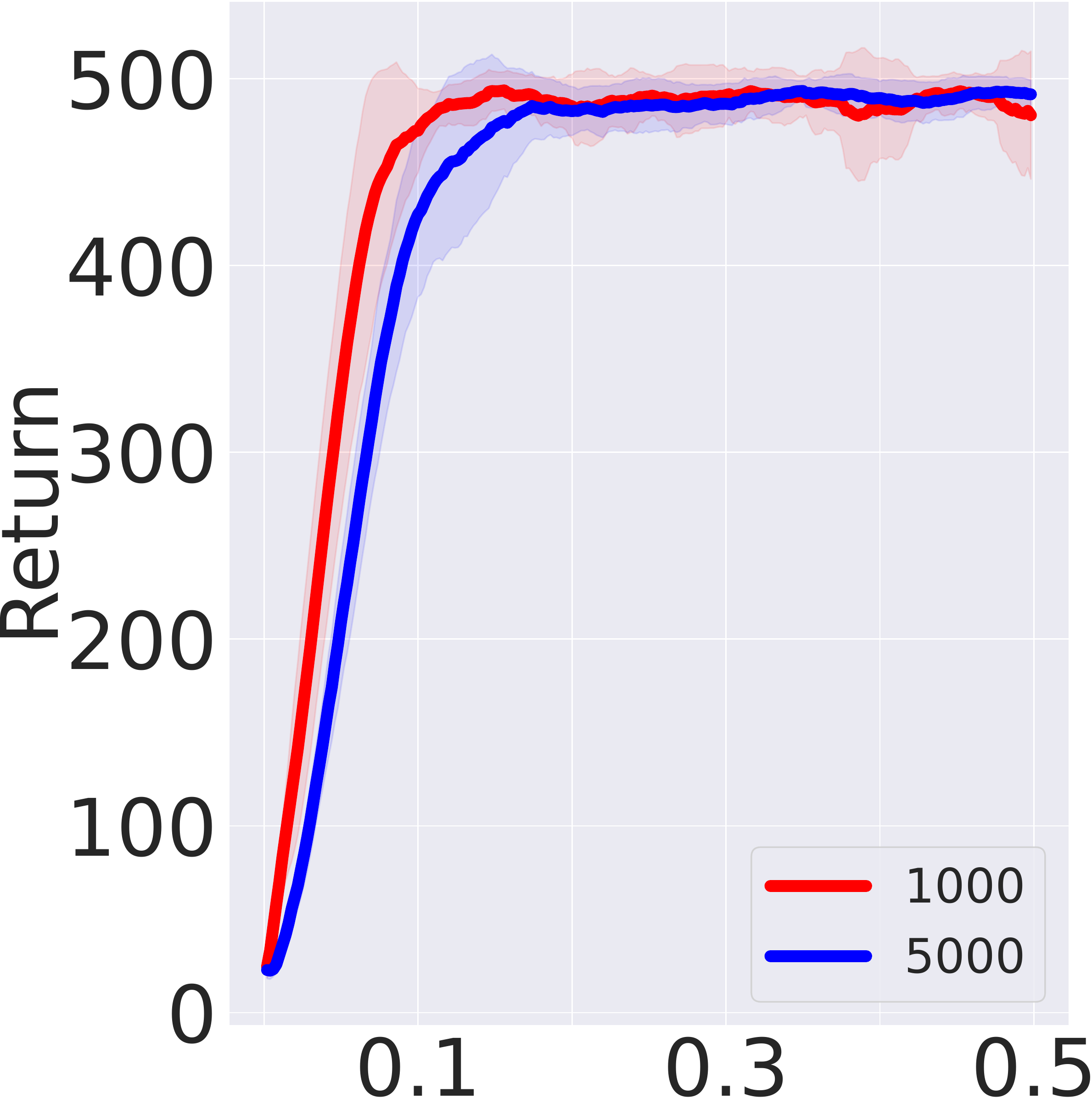}
    \vspace{4pt}
  \end{subfigure}%
  \begin{subfigure}[b]{0.2\textwidth}
    \centering
    \includegraphics[width=0.9\textwidth]{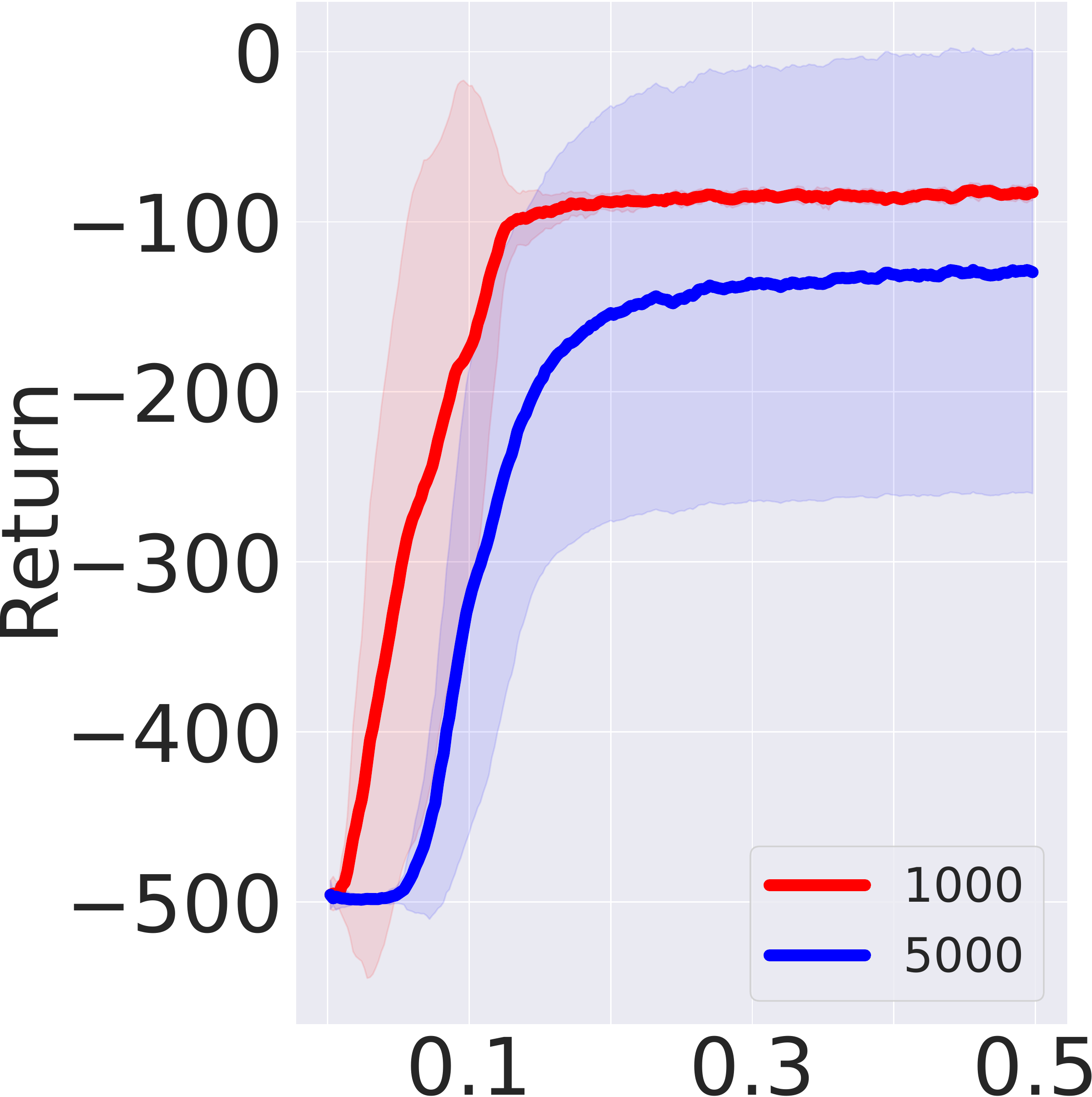}
    \vspace{4pt}
  \end{subfigure}%
  \begin{subfigure}[b]{0.2\textwidth}
    \centering
    \includegraphics[width=0.9\textwidth]{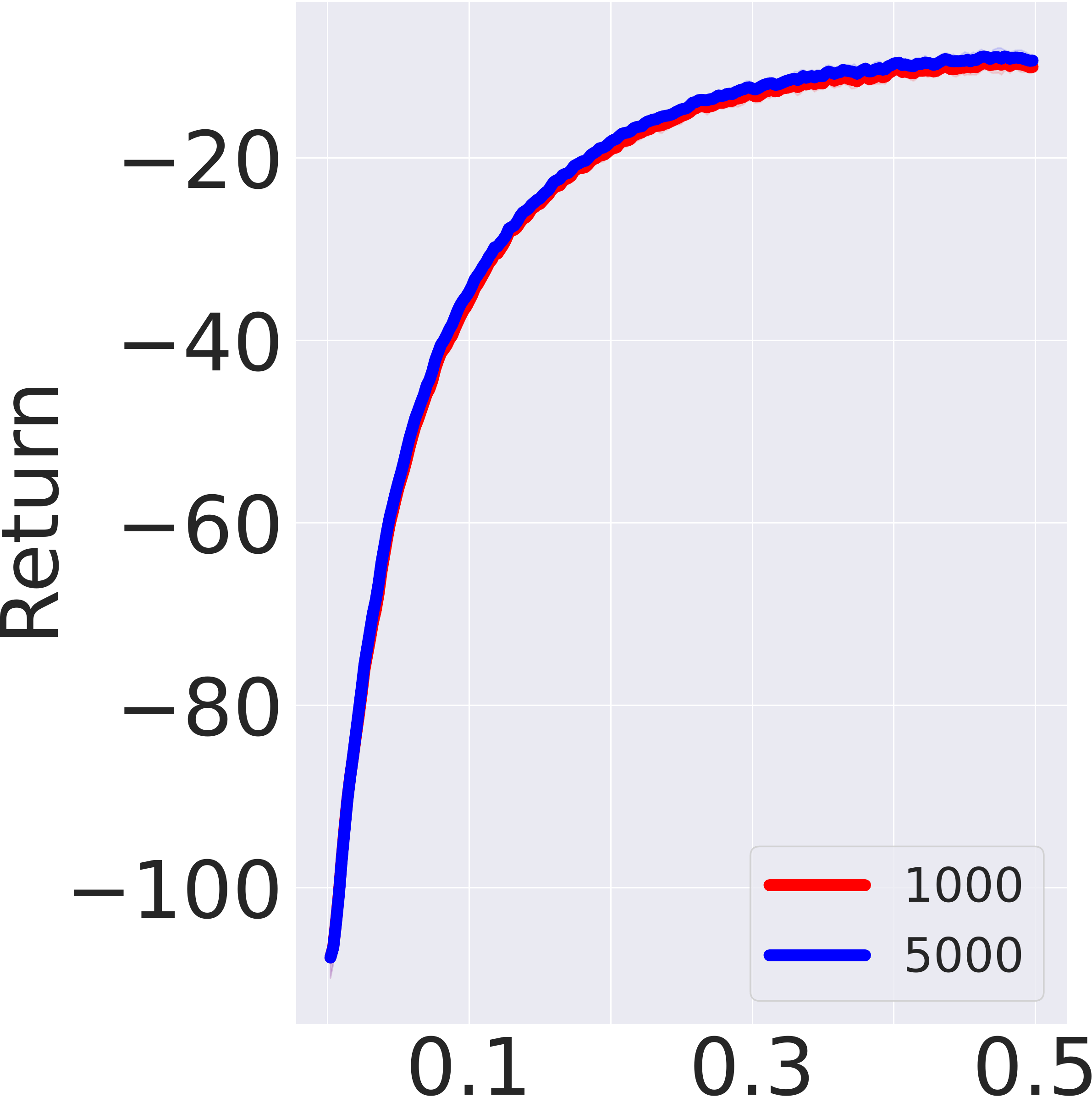}
    \vspace{4pt}
  \end{subfigure}%
  \begin{subfigure}[b]{0.2\textwidth}
    \centering
    \includegraphics[width=0.9\textwidth]{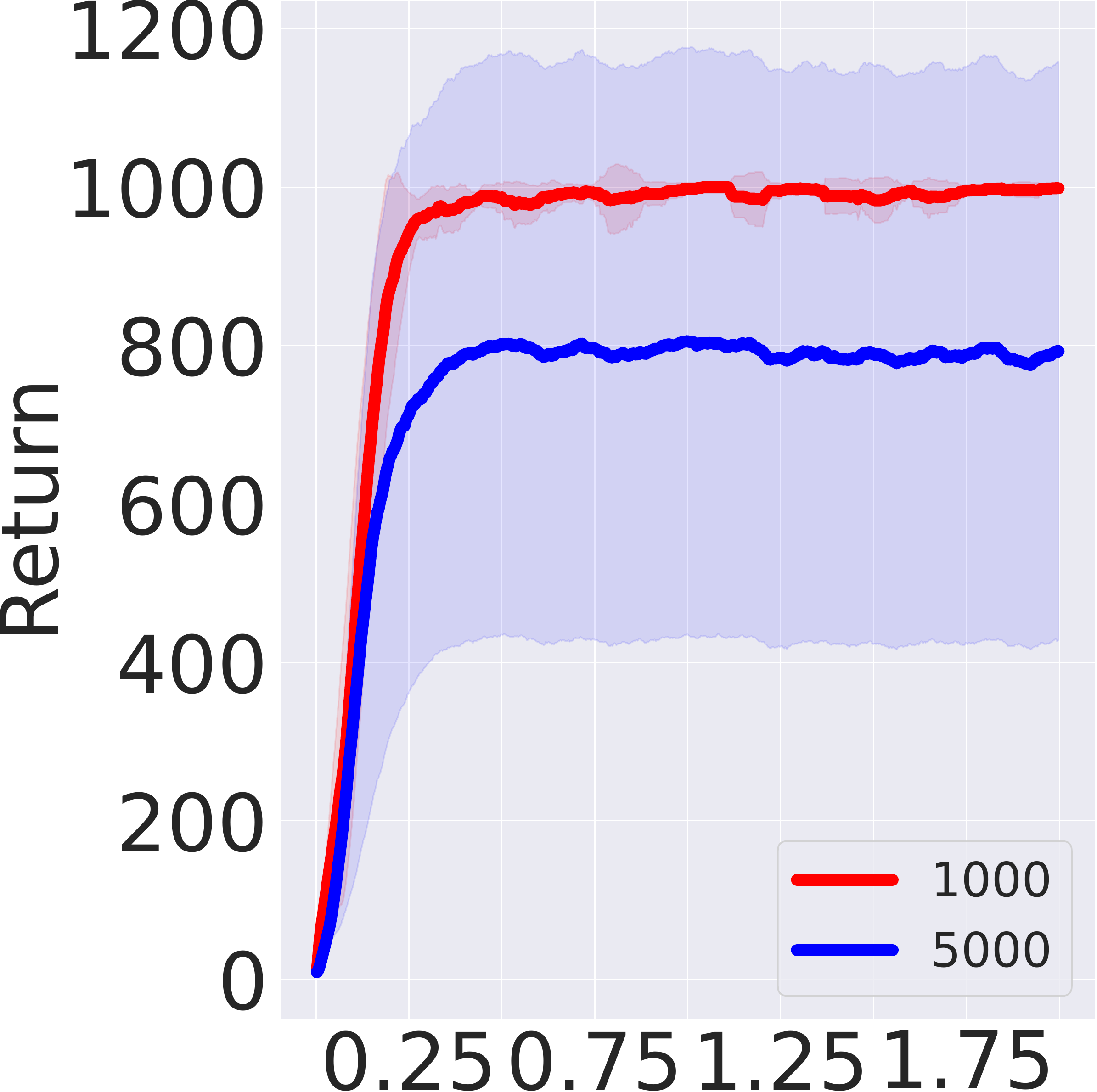}
    \vspace{4pt}
  \end{subfigure}%
  \begin{subfigure}[b]{0.2\textwidth}
    \centering
    \includegraphics[width=0.9\textwidth]{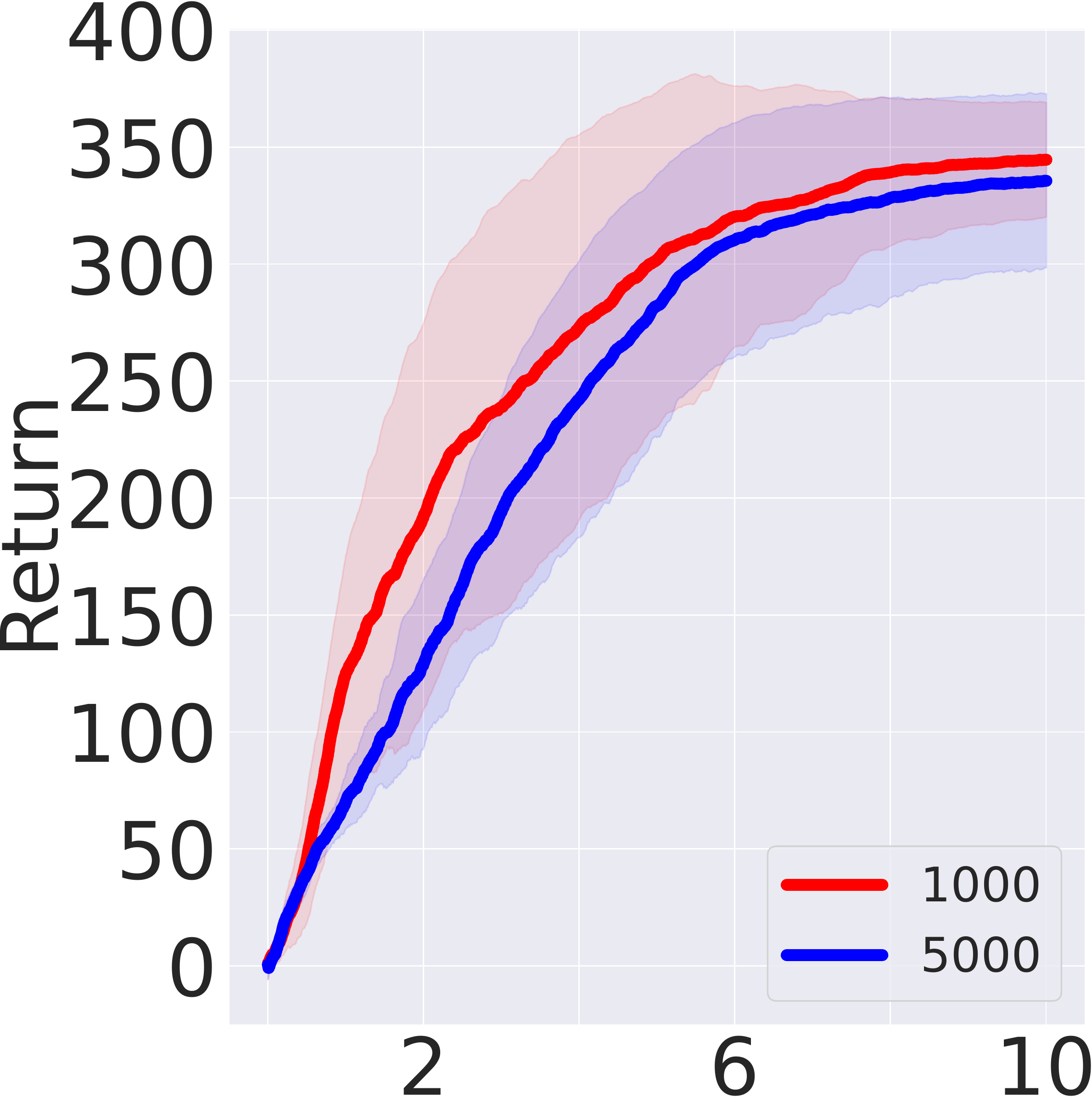}
    \vspace{4pt}
  \end{subfigure}%
  
  \begin{subfigure}[b]{0.2\textwidth}
    \centering
    \includegraphics[width=0.9\textwidth]{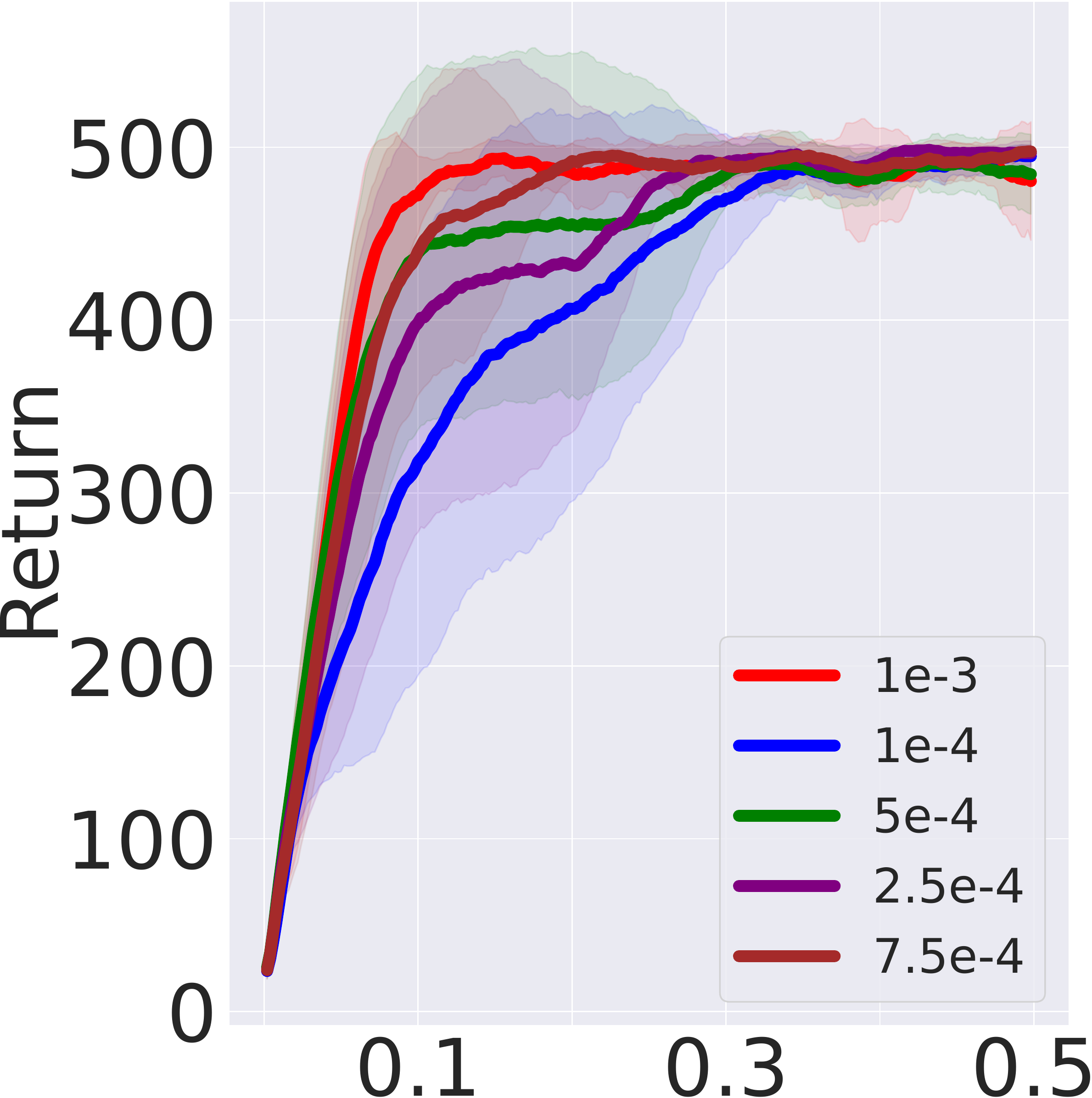}
    \vspace{4pt}
  \end{subfigure}%
  \begin{subfigure}[b]{0.2\textwidth}
    \centering
    \includegraphics[width=0.9\textwidth]{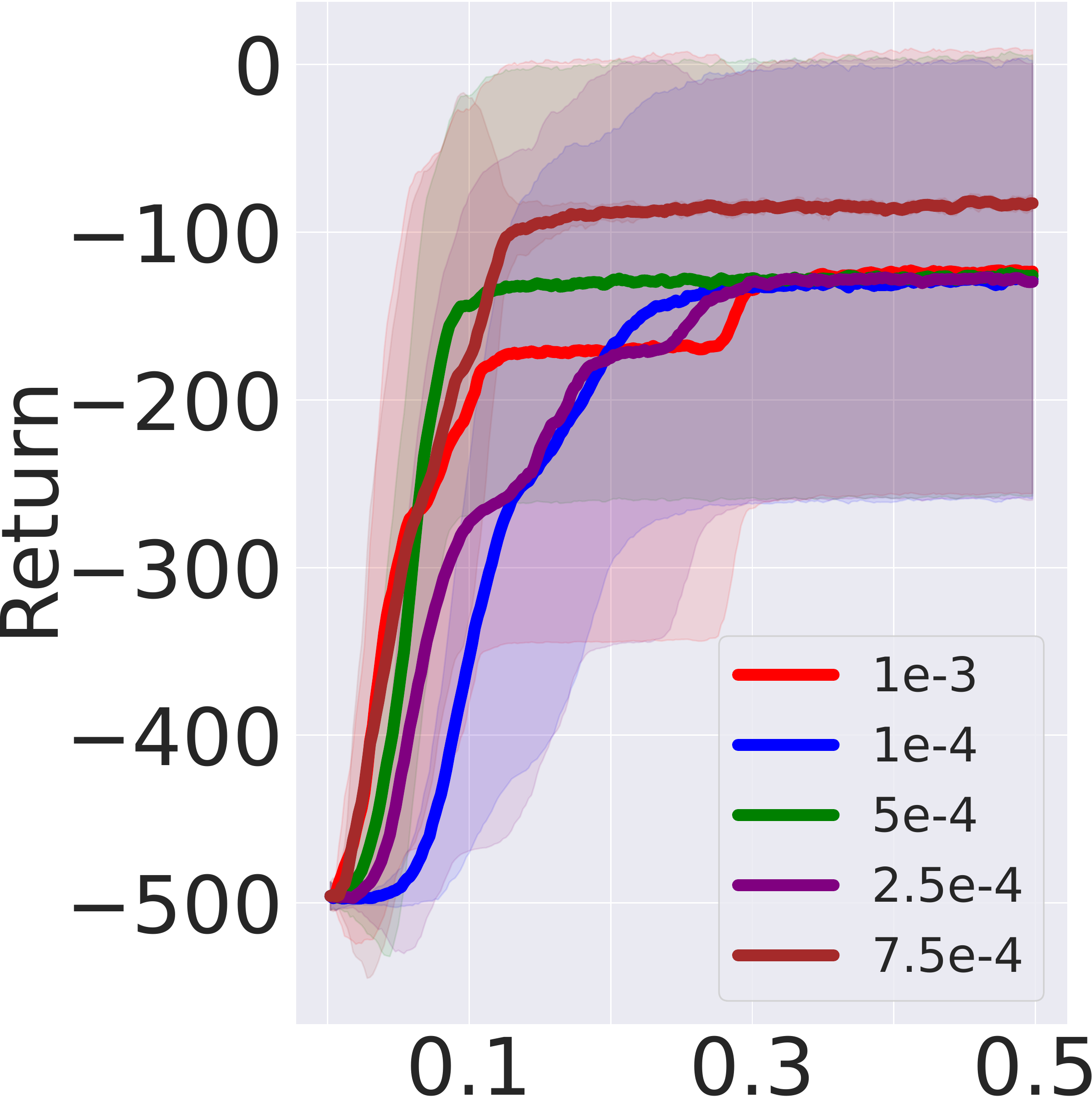}
    \vspace{4pt}
  \end{subfigure}%
  \begin{subfigure}[b]{0.2\textwidth}
    \centering
    \includegraphics[width=0.9\textwidth]{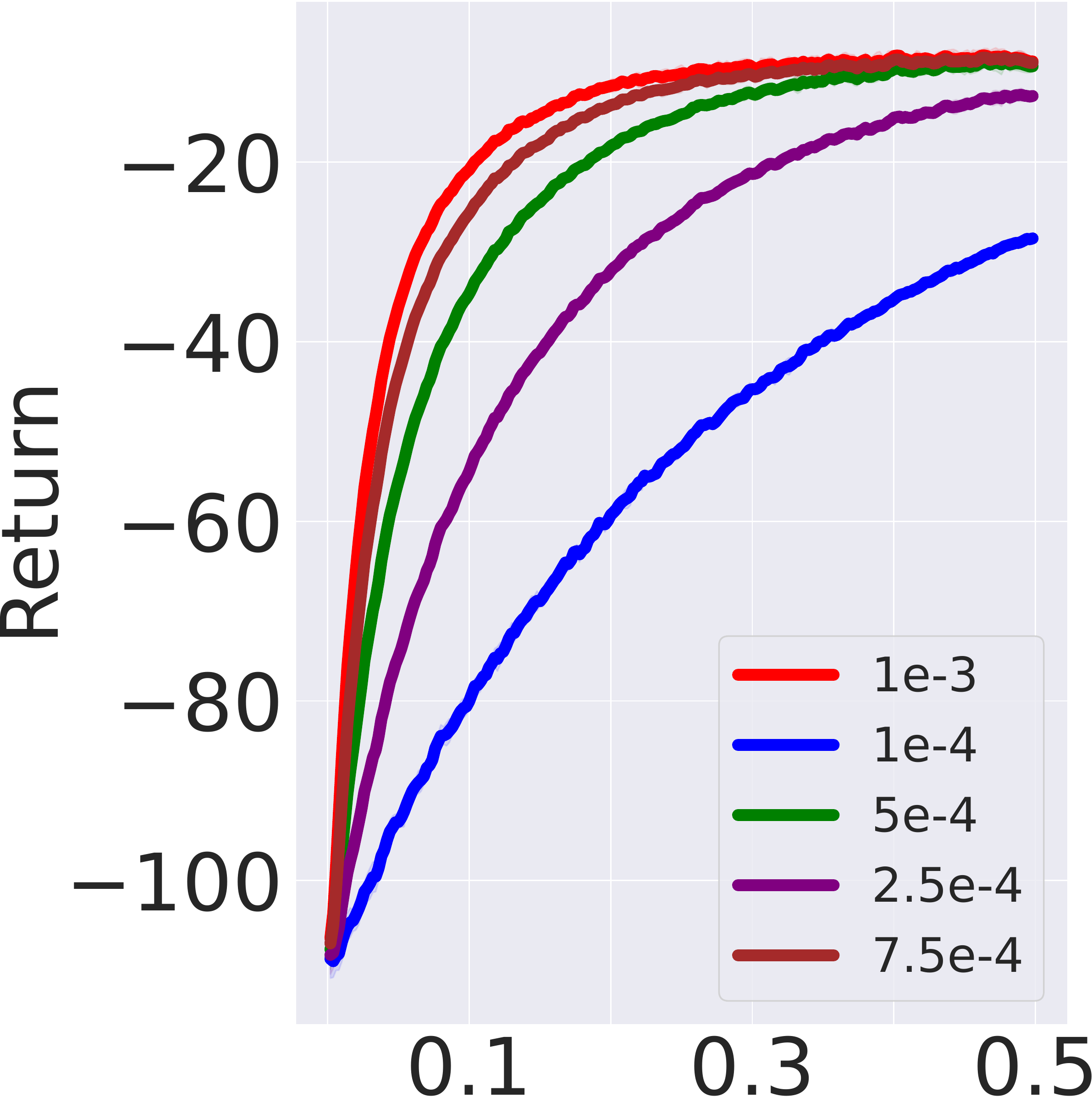}
    \vspace{4pt}
  \end{subfigure}%
  \begin{subfigure}[b]{0.2\textwidth}
    \centering
    \includegraphics[width=0.9\textwidth]{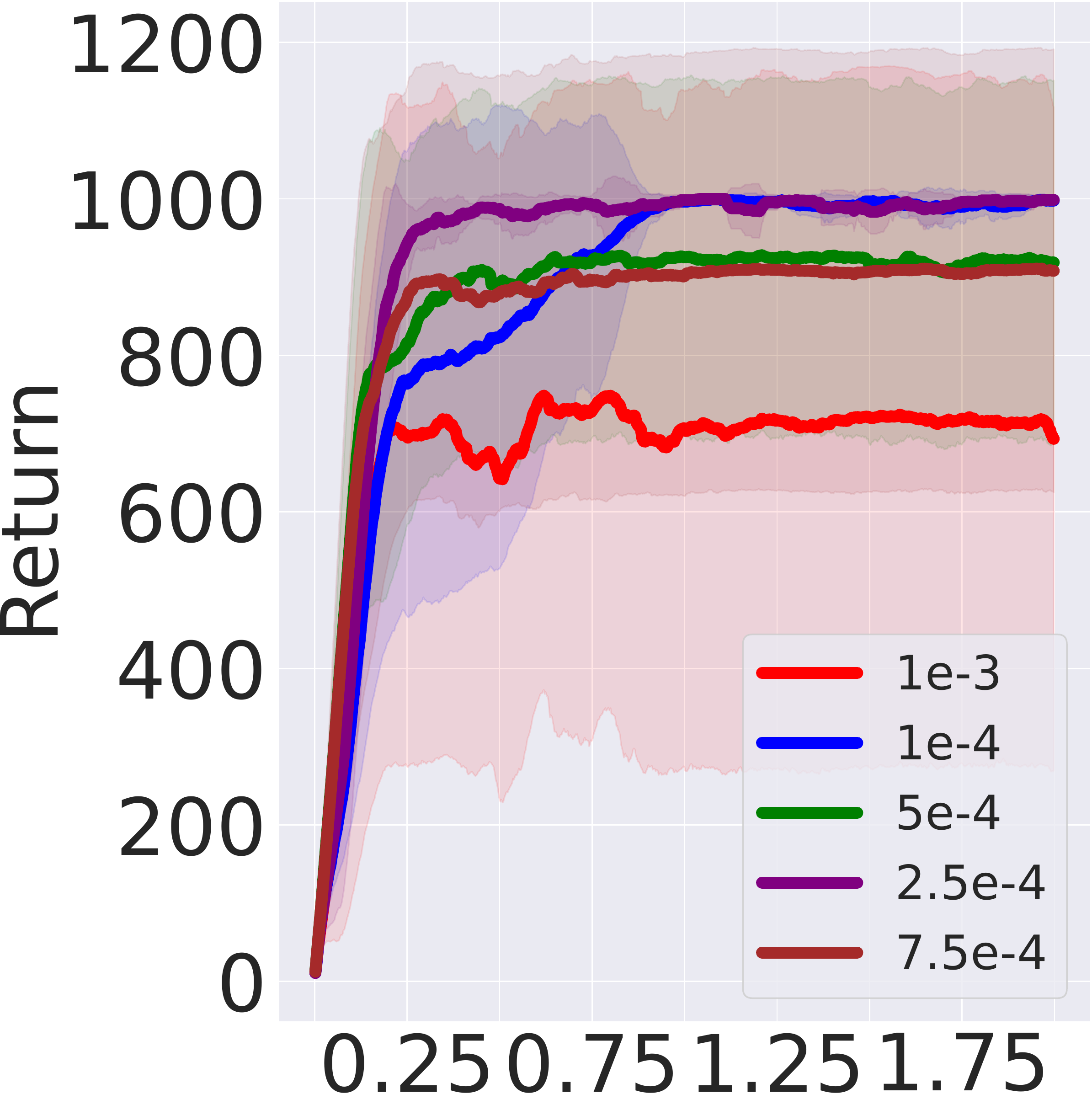}
    \vspace{4pt}
  \end{subfigure}%
  \begin{subfigure}[b]{0.2\textwidth}
    \centering
    \includegraphics[width=0.9\textwidth]{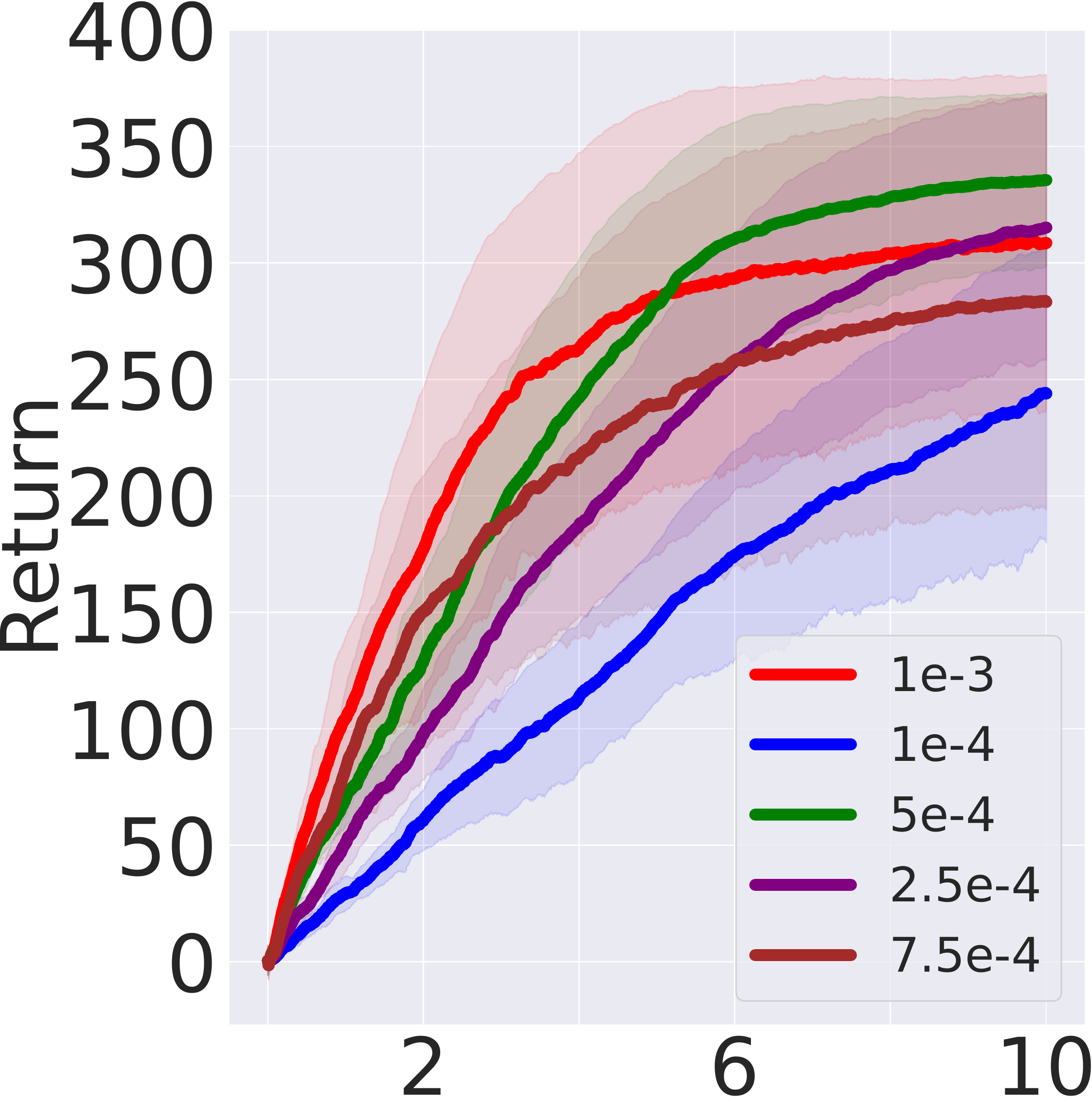}
    \vspace{4pt}
  \end{subfigure}%
  
  \begin{subfigure}[b]{0.2\textwidth}
    \centering
    \includegraphics[width=0.9\textwidth]{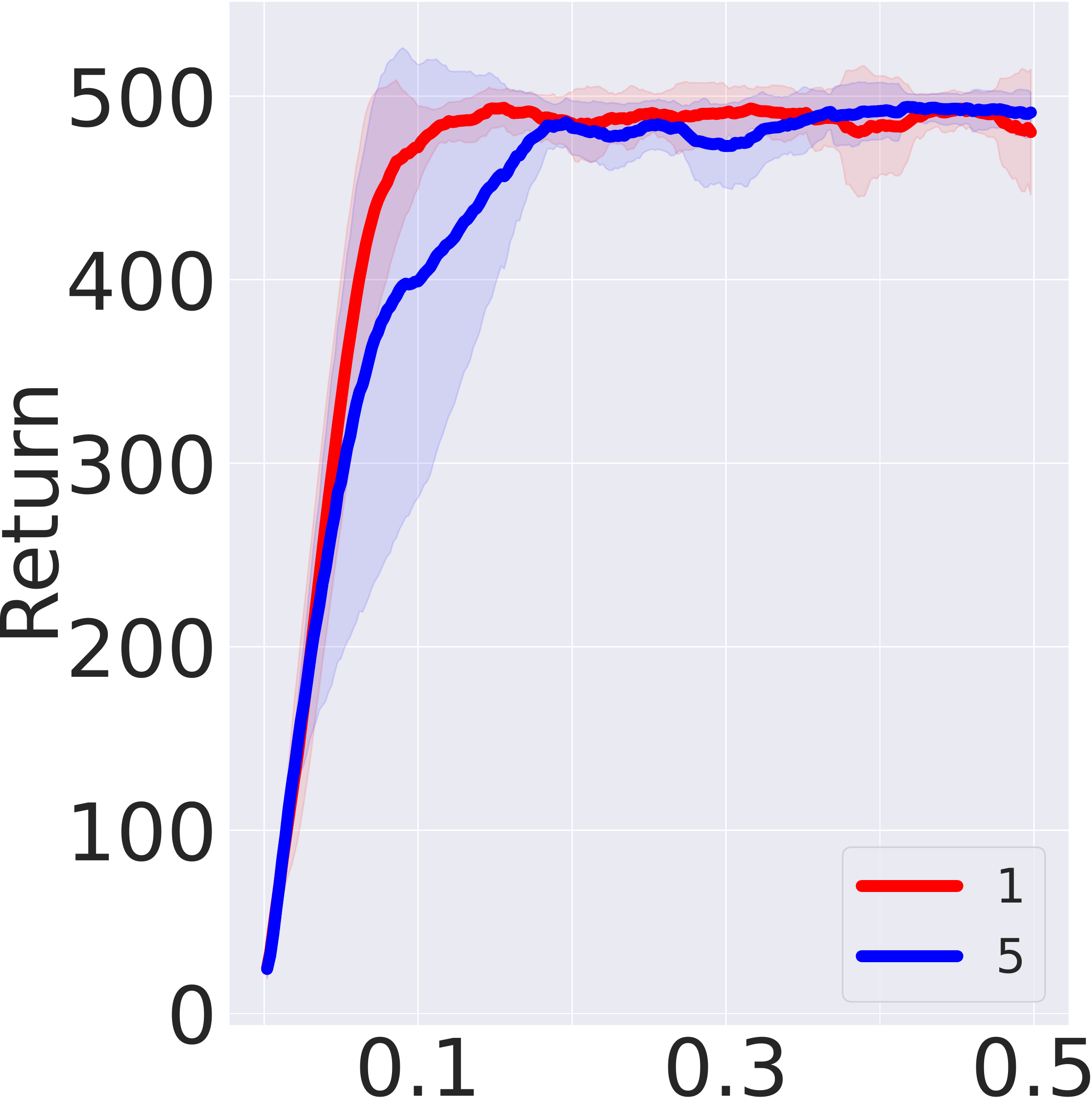}
    \vspace{4pt}
  \end{subfigure}%
  \begin{subfigure}[b]{0.2\textwidth}
    \centering
    \includegraphics[width=0.9\textwidth]{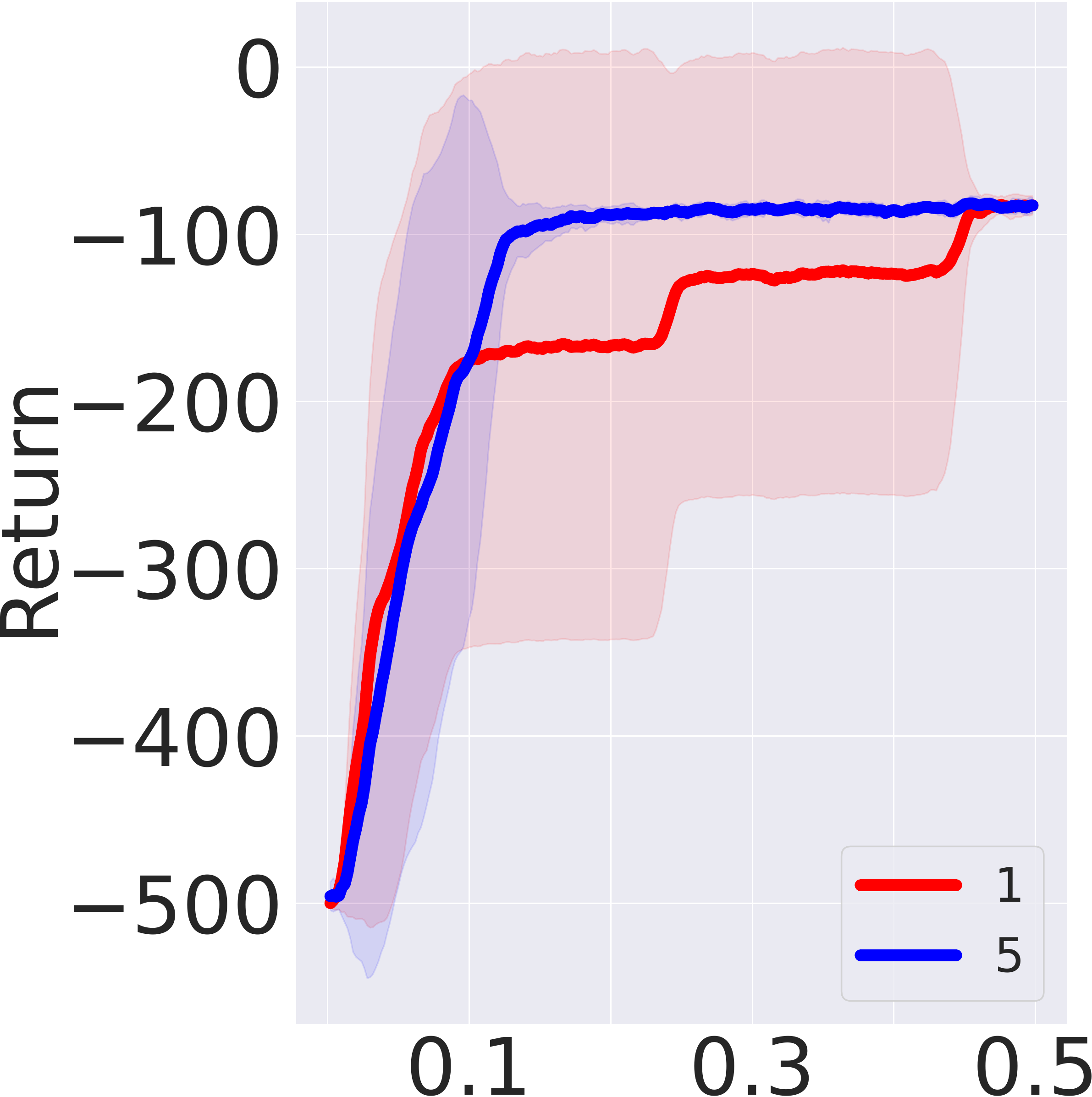}
    \vspace{4pt}
  \end{subfigure}%
  \begin{subfigure}[b]{0.2\textwidth}
    \centering
    \includegraphics[width=0.9\textwidth]{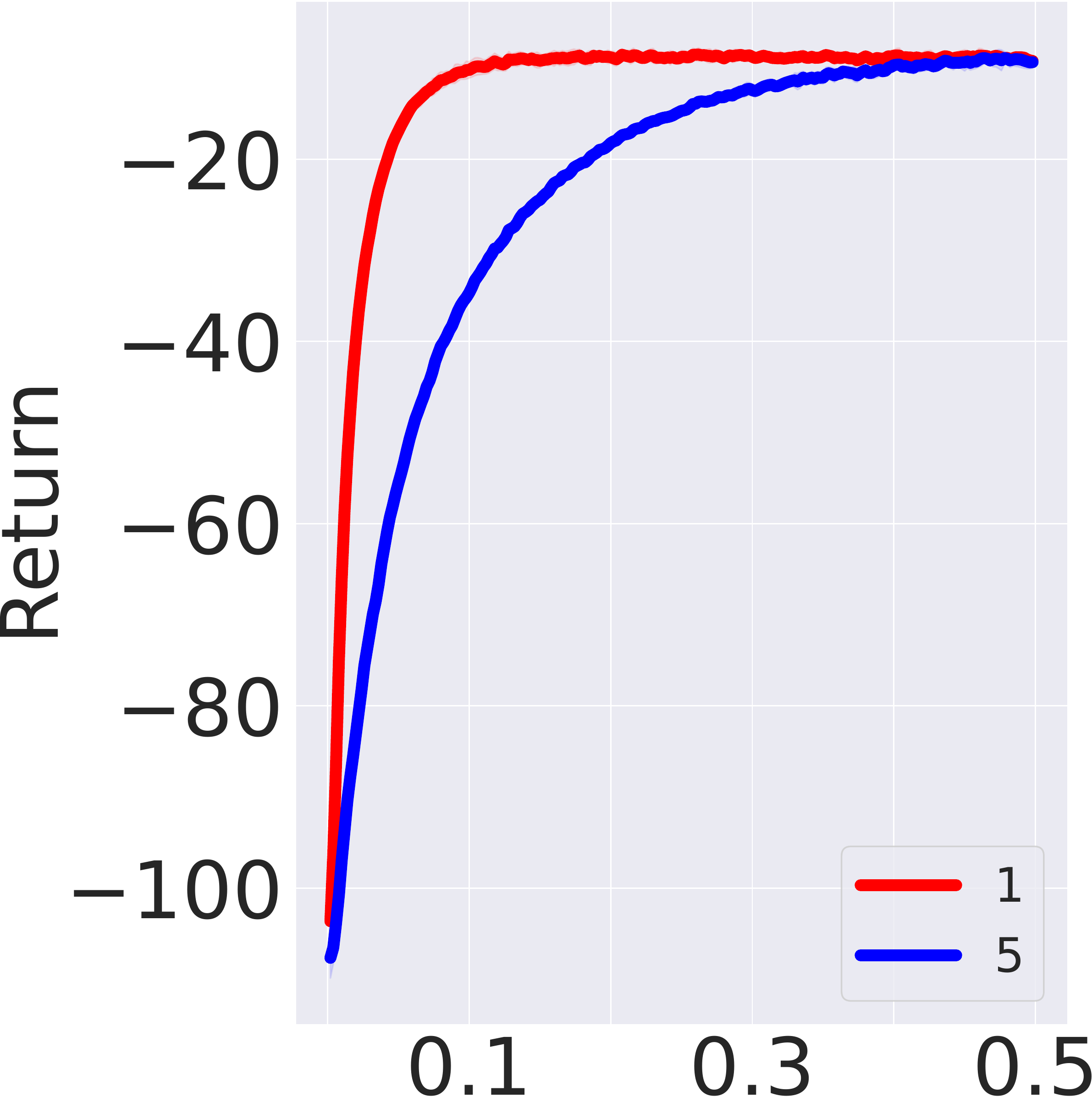}
    \vspace{4pt}
  \end{subfigure}%
  \begin{subfigure}[b]{0.2\textwidth}
    \centering
    \includegraphics[width=0.9\textwidth]{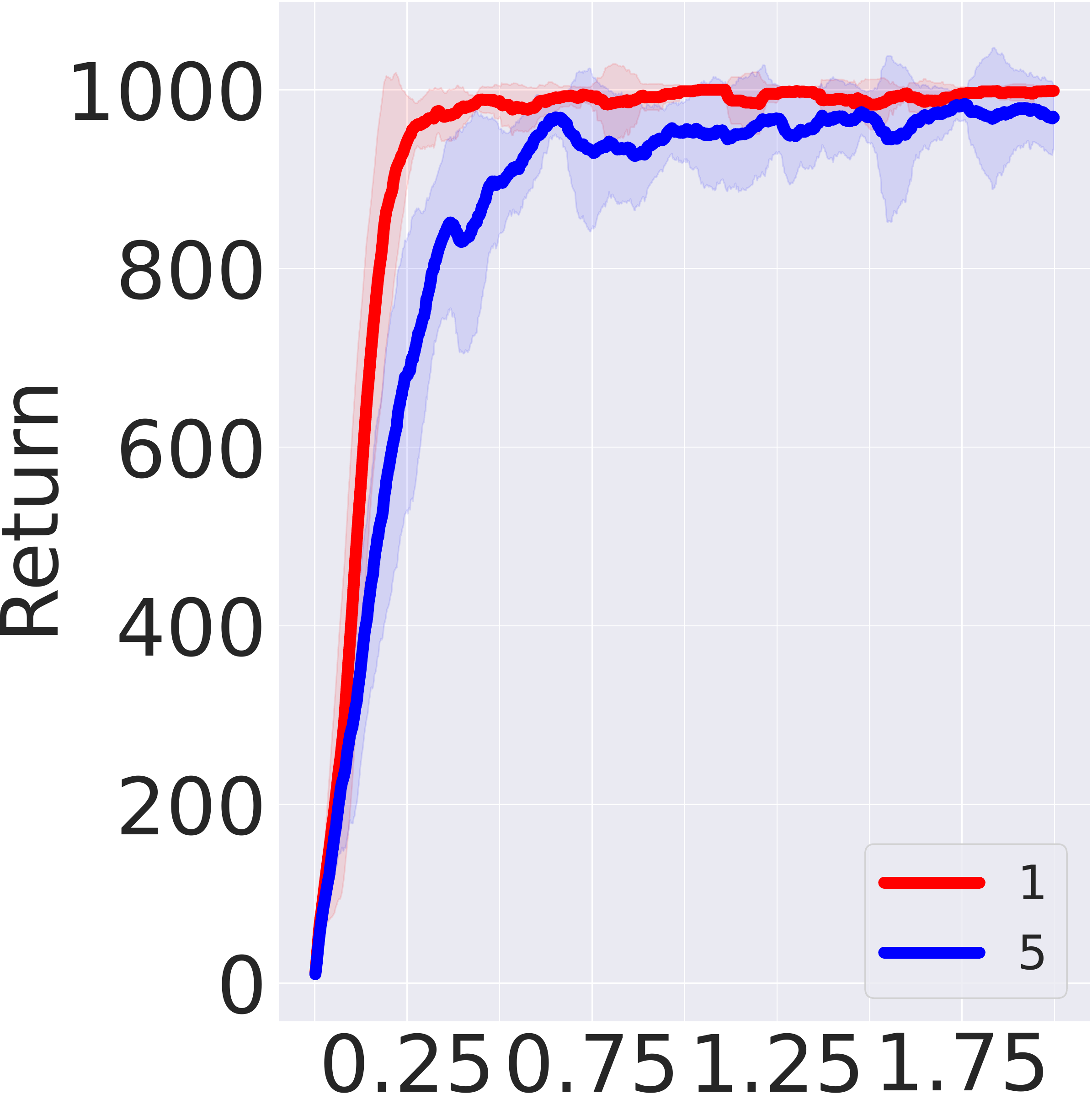}
    \vspace{4pt}
  \end{subfigure}%
  \begin{subfigure}[b]{0.2\textwidth}
    \centering
    \includegraphics[width=0.9\textwidth]{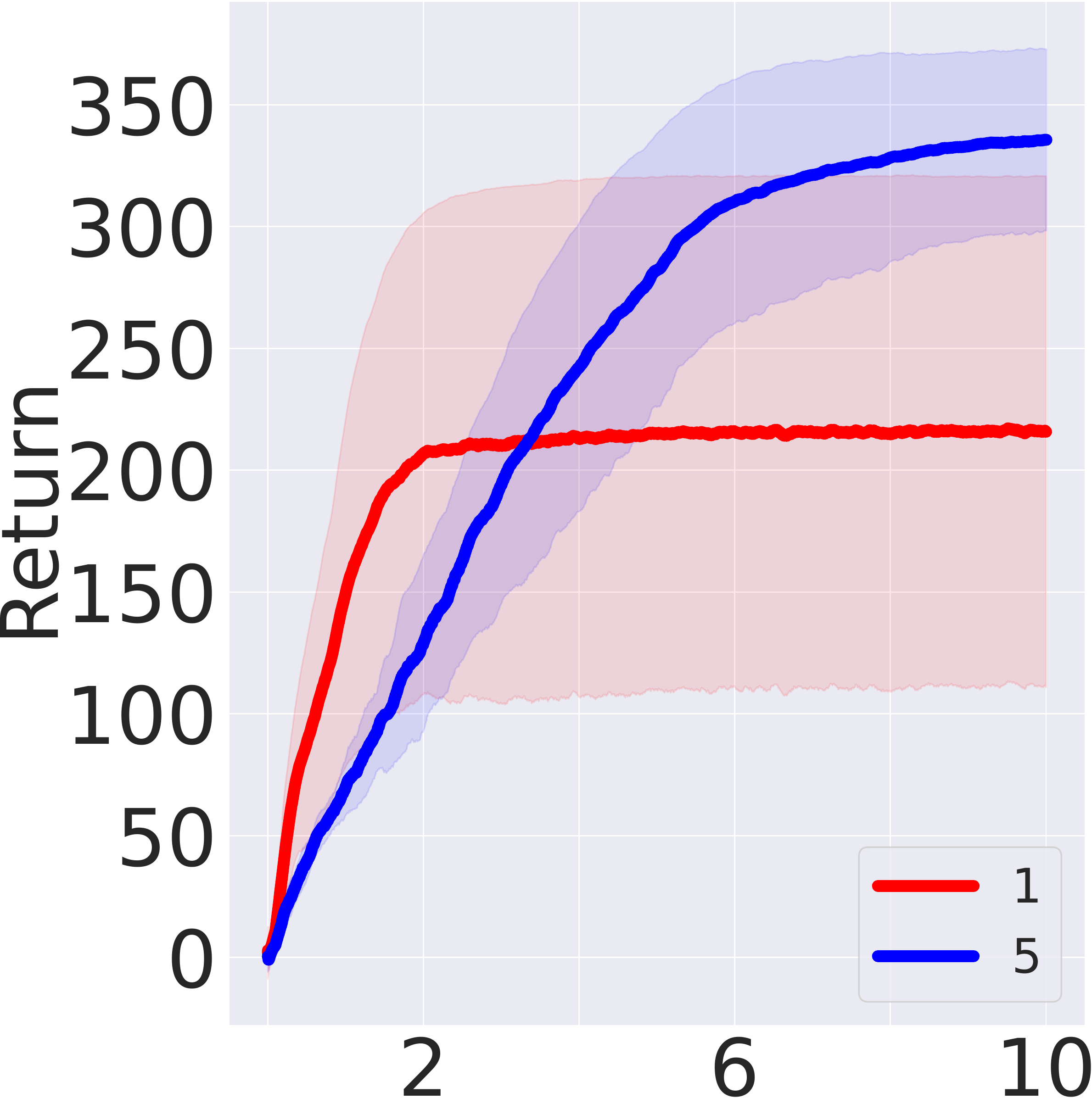}
    \vspace{4pt}
  \end{subfigure}%
  
  \centering
  \begin{subfigure}[b]{0.2\textwidth}
    \centering
    \includegraphics[width=0.9\textwidth]{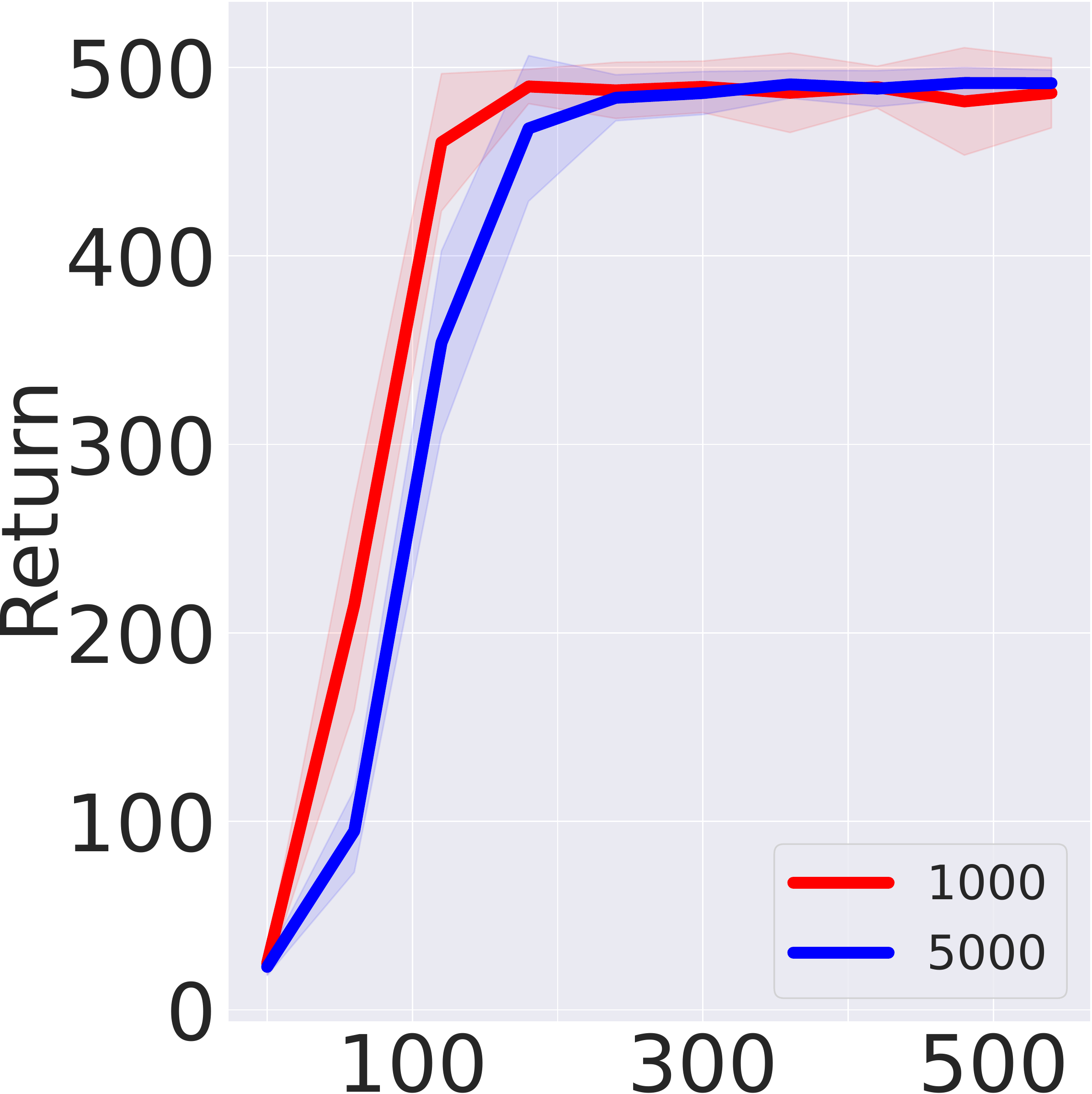}
    \vspace{4pt}
  \end{subfigure}%
  \begin{subfigure}[b]{0.2\textwidth}
    \centering
    \includegraphics[width=0.9\textwidth]{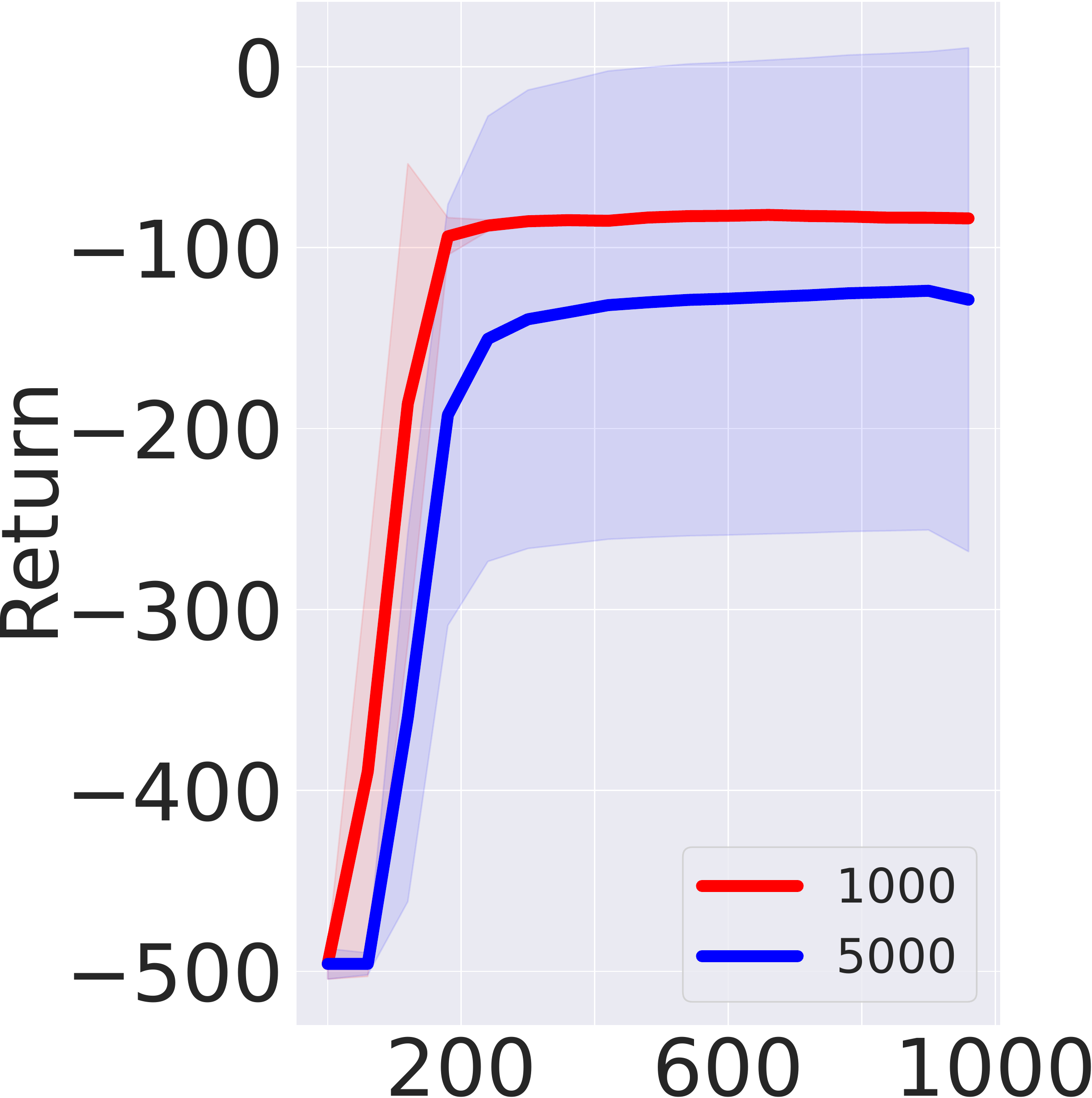}
    \vspace{4pt}
  \end{subfigure}%
  \begin{subfigure}[b]{0.2\textwidth}
    \centering
    \includegraphics[width=0.9\textwidth]{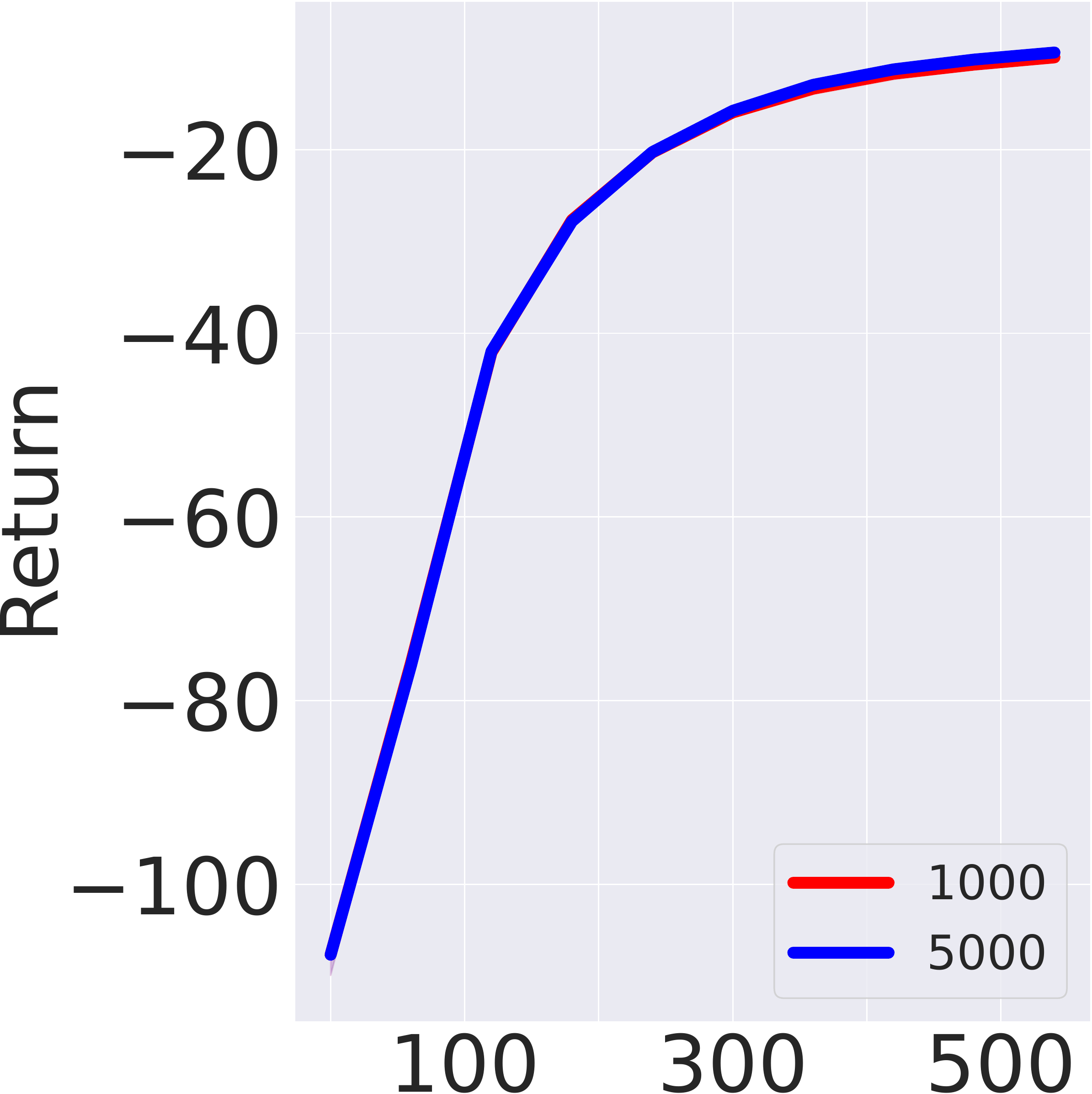}
    \vspace{4pt}
  \end{subfigure}%
  \begin{subfigure}[b]{0.2\textwidth}
    \centering
    \includegraphics[width=0.9\textwidth]{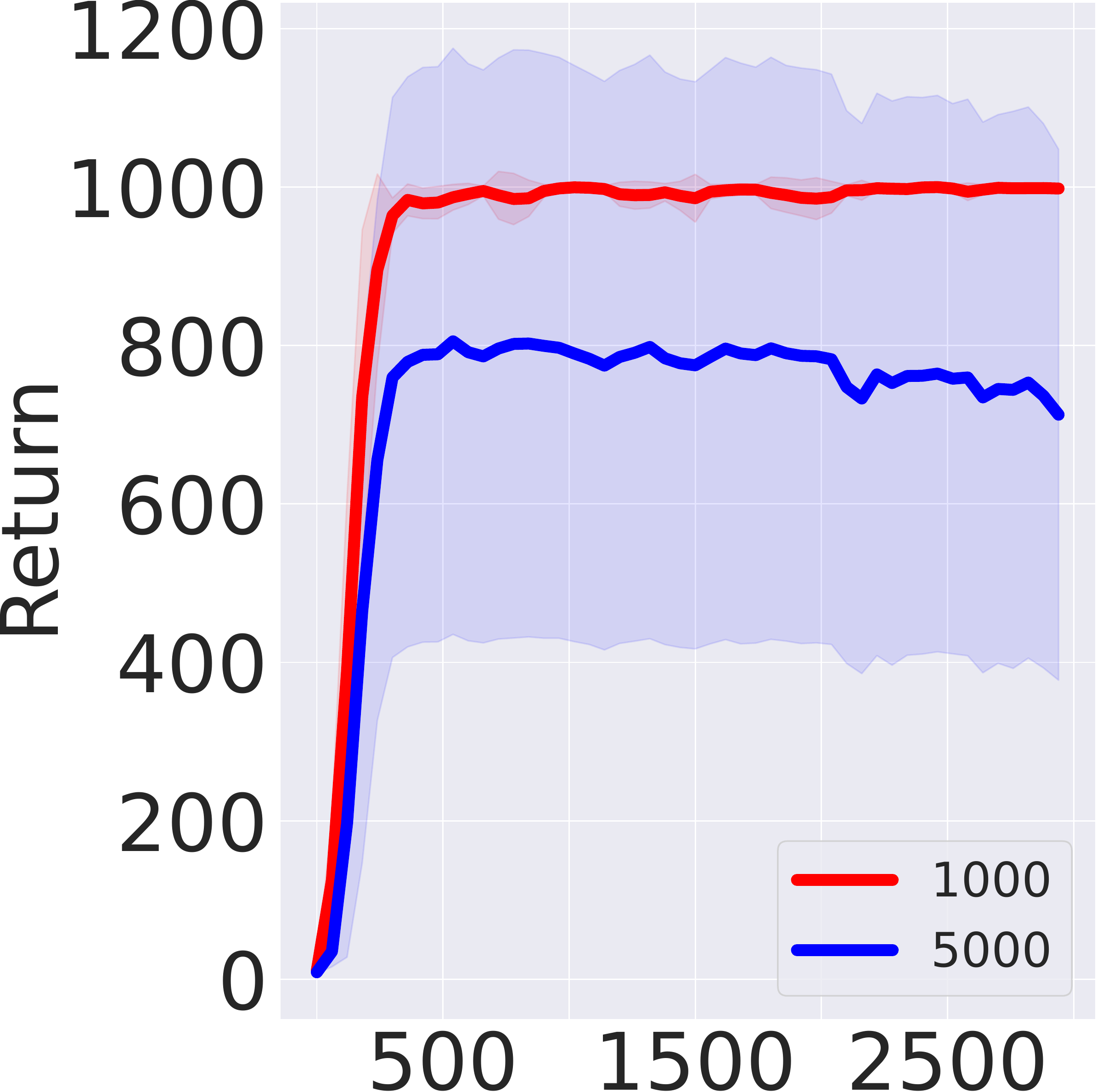}
    \vspace{4pt}
  \end{subfigure}%
  \begin{subfigure}[b]{0.2\textwidth}
    \centering
    \includegraphics[width=0.9\textwidth]{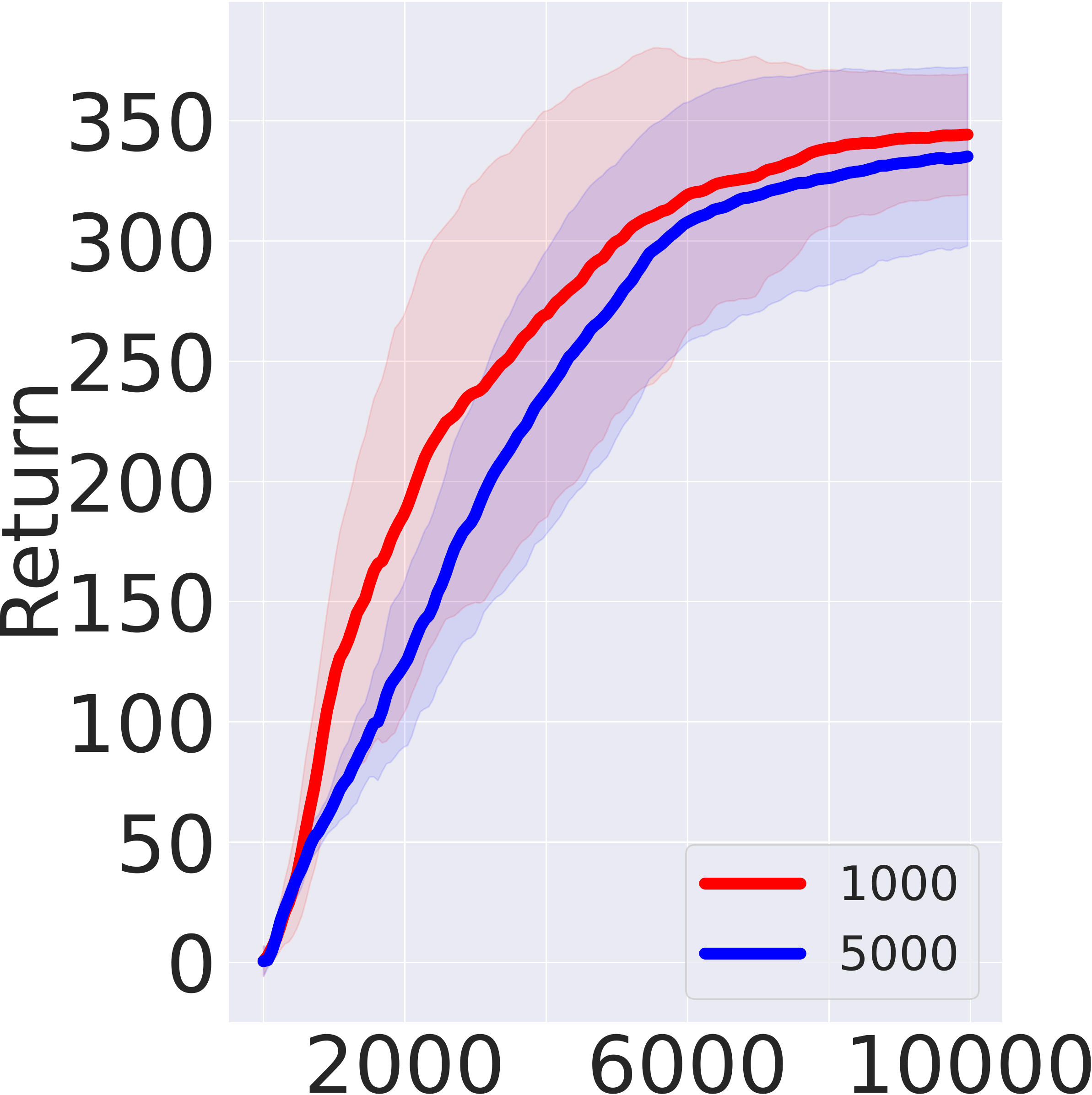}
    \vspace{4pt}
  \end{subfigure}%
  
  \begin{subfigure}[b]{0.2\textwidth}
    \centering
    \includegraphics[width=0.9\textwidth]{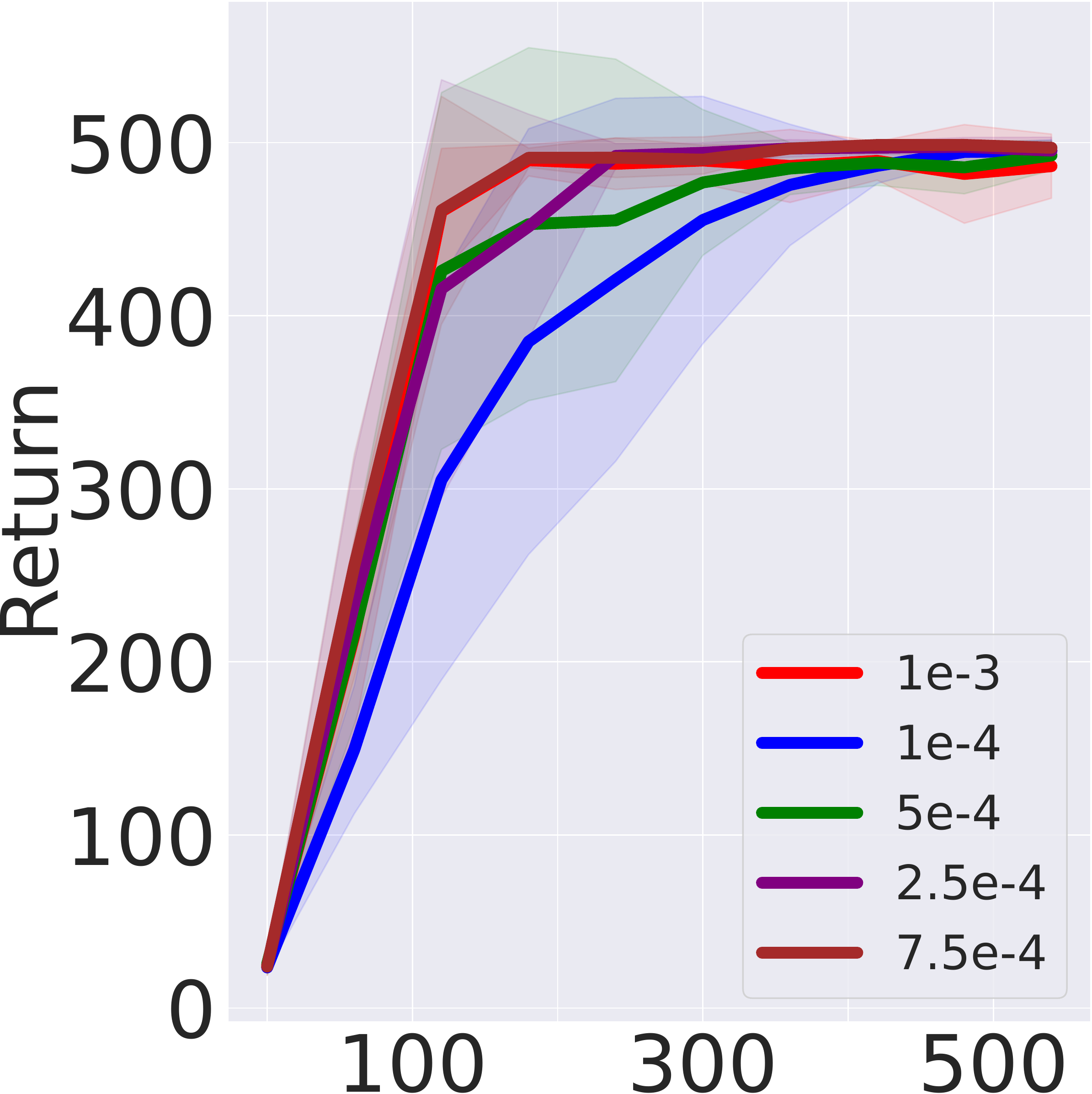}
    \caption{CartPole-v1}
  \end{subfigure}%
  \begin{subfigure}[b]{0.2\textwidth}
    \centering
    \includegraphics[width=0.9\textwidth]{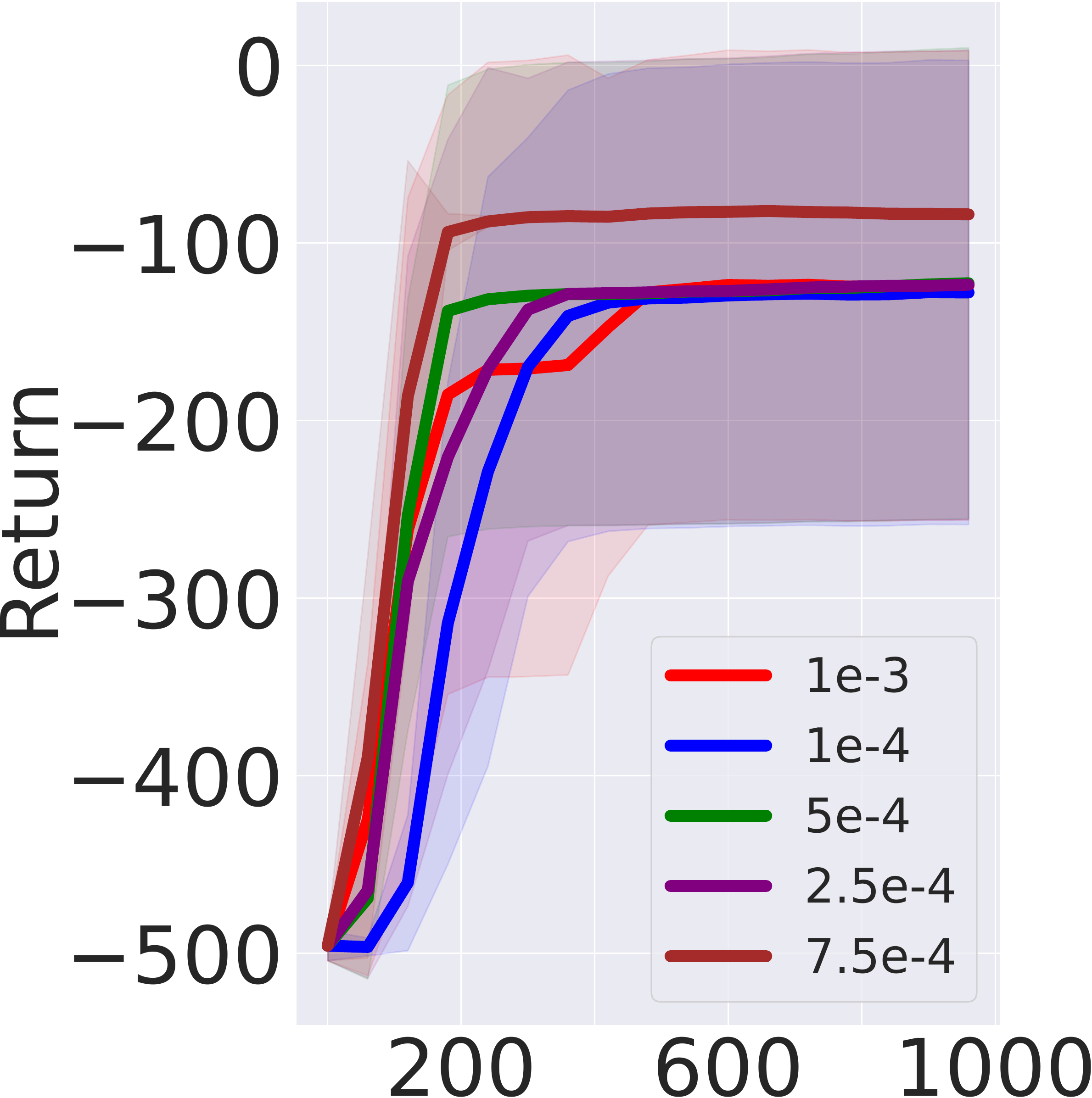}
    \caption{Acrobot-v1}
  \end{subfigure}%
  \begin{subfigure}[b]{0.2\textwidth}
    \centering
    \includegraphics[width=0.9\textwidth]{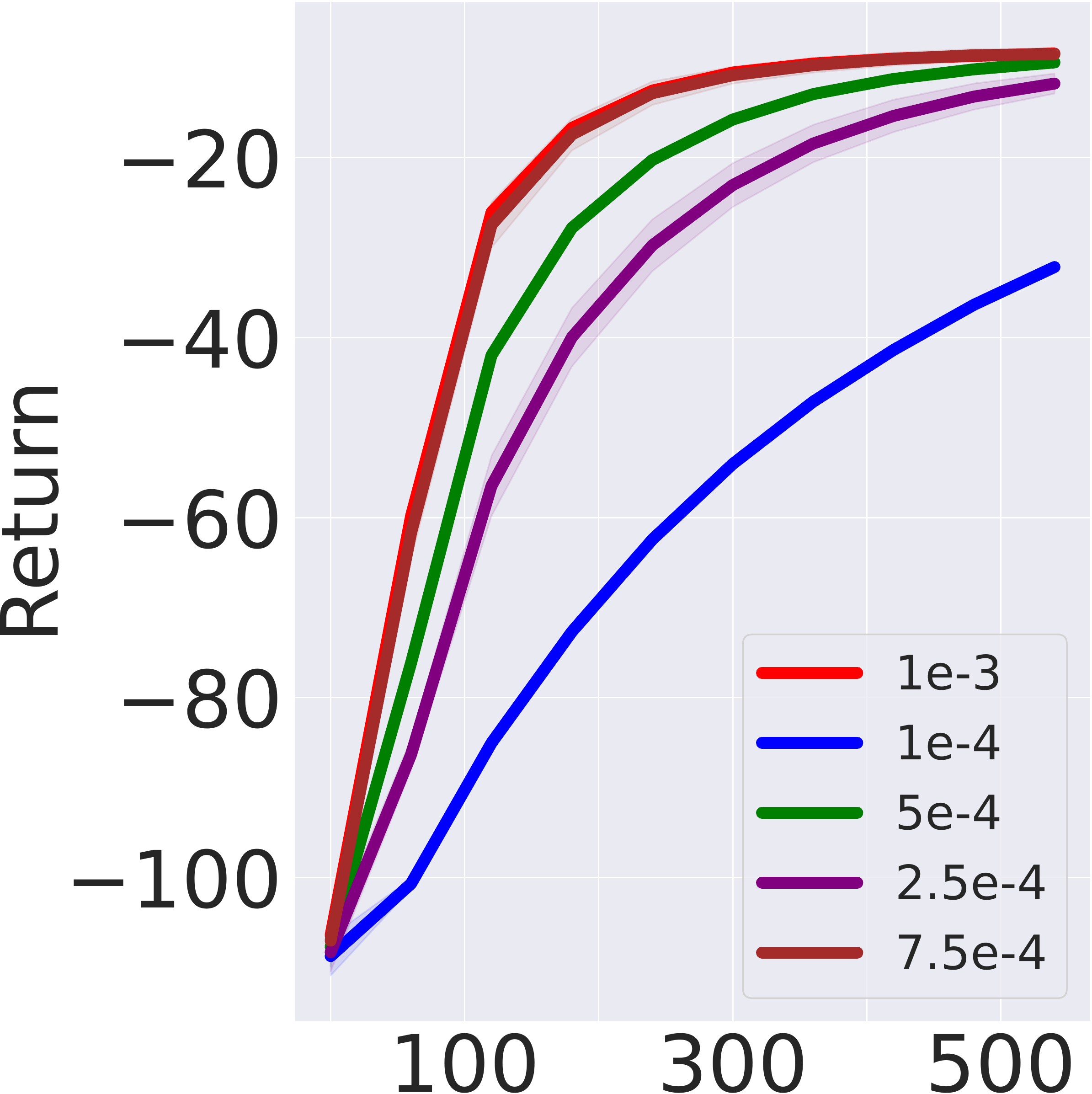}
    \caption{Reacher-v2}
  \end{subfigure}%
  \begin{subfigure}[b]{0.2\textwidth}
    \centering
    \includegraphics[width=0.9\textwidth]{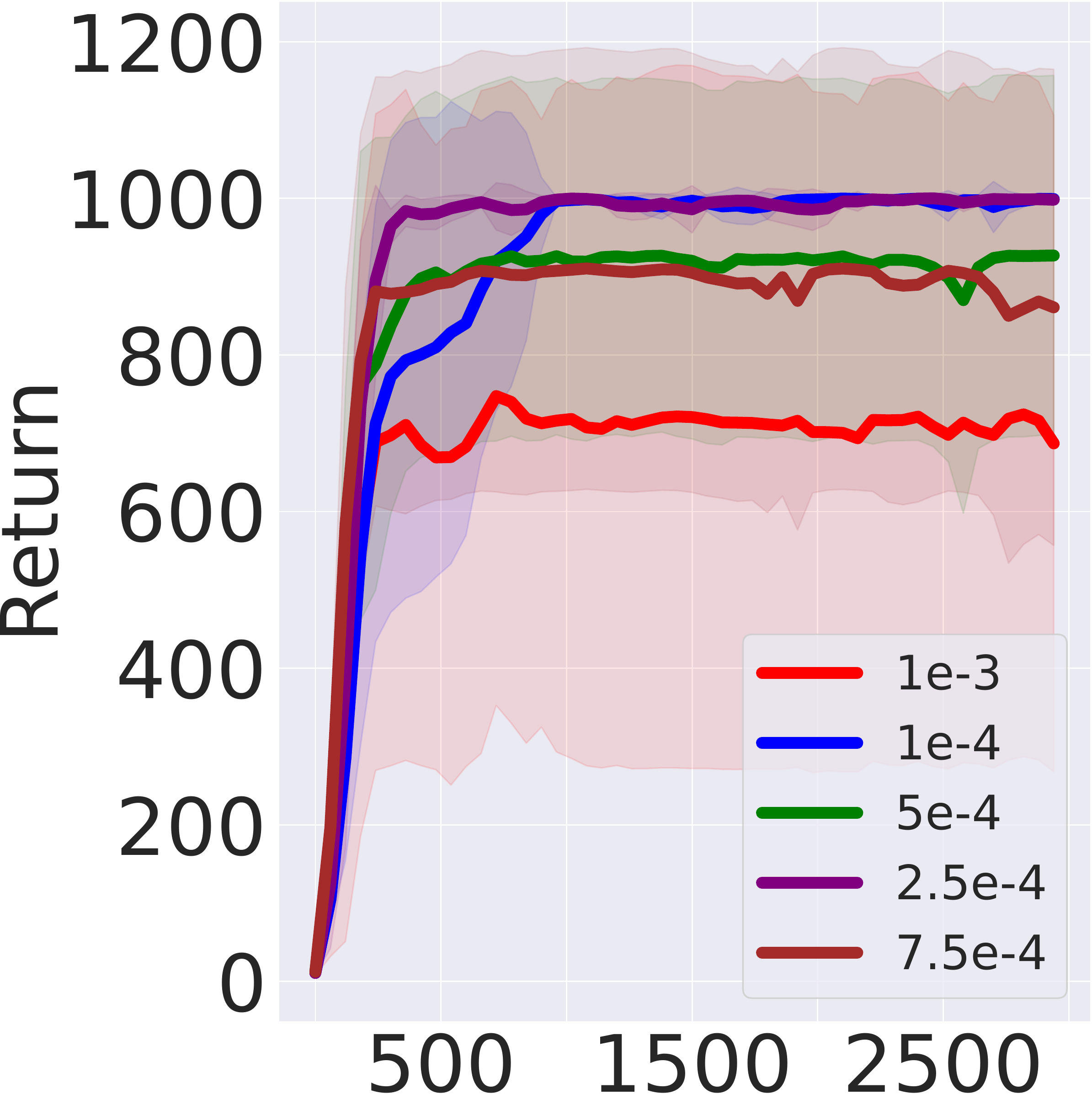}
    \caption{InvertedPendulum-v2}
  \end{subfigure}%
  \begin{subfigure}[b]{0.2\textwidth}
    \centering
    \includegraphics[width=0.9\textwidth]{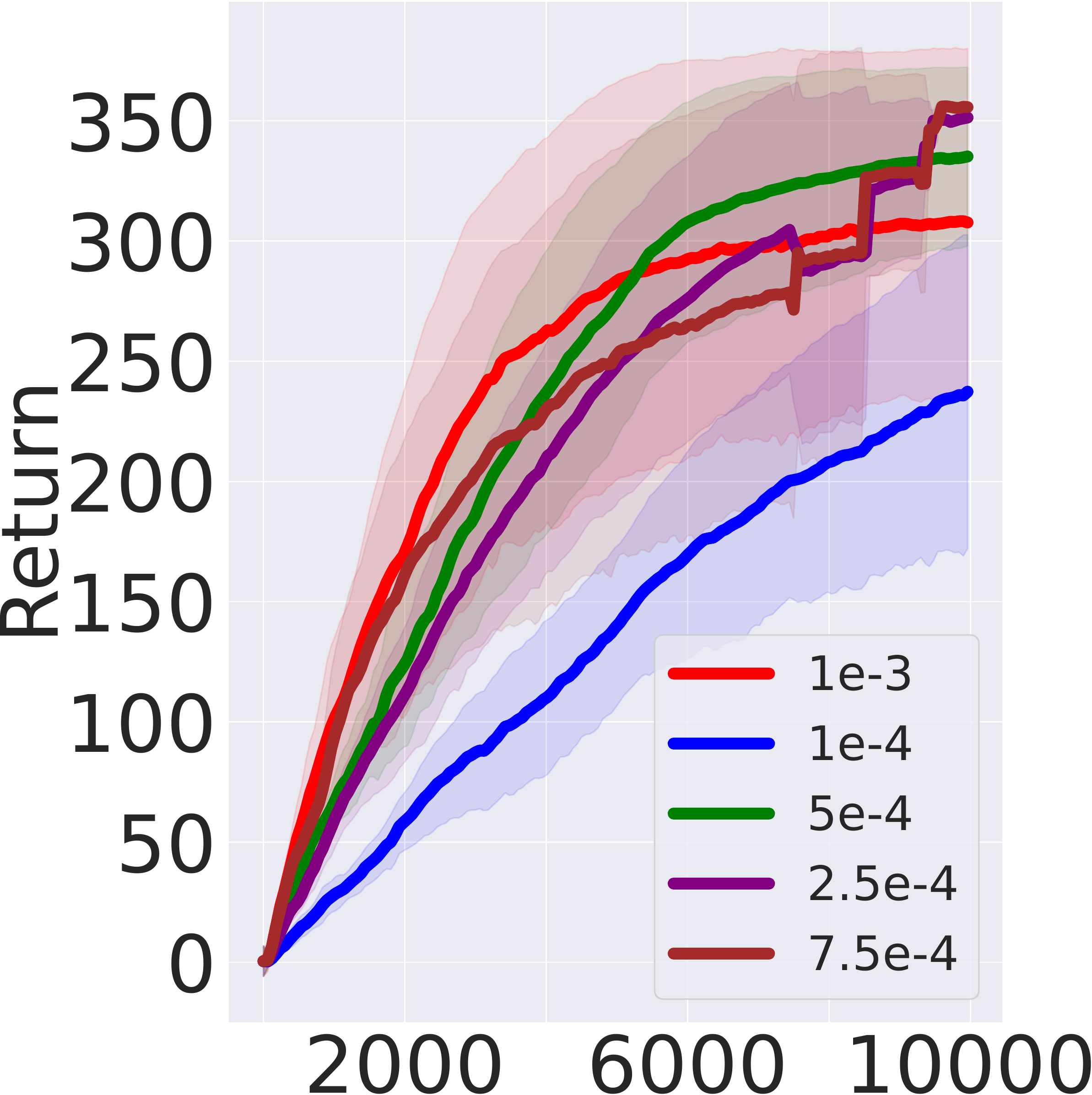}
    \caption{Swimmer-v2}
  \end{subfigure}%
  \caption{Performance of SSRL with different hyperparameters. We vary one of the hyperparameters with other hyperparameters fixed. Top $3$ rows show performance with respect to the timesteps using different buffer size, learning rates and rollout steps, respectively. Bottom $3$ rows show the performance with respect to the running time using different buffer size, learning rates and rollout steps, respectively. All the experiments are run $10$ times with seeds $0$ to $9$. The shaded area represents mean $\pm$ standard deviation. The x-axis in the top $3$ rows denotes millions of timesteps. The x-axis in the bottom $3$ rows denotes the running time in seconds.}
  \label{fig:1}
\end{figure*}

\textbf{Self-Supervised Reinforcement Learning (SSRL)} For both discrete and continuous control tasks, we use the 64-64 MLP implemented in OpenAI baselines\footnote{https://github.com/openai/baselines} as policy networks. The actions are sampled following Gaussian distribution. For discrete control tasks, we conduct supervised learning by minimizing the negative log probability of the action. For continuous domain, we use MSE loss. We implement a small ranking buffer to store the top trajectories. Specifically, when a new episode is added, we rank the transitions in descending order according to the corresponding trajectory rewards. Then we only keep the top $K$ transitions, where $K$ is the buffer size. We use Adam Optimizer with default settings in Tensorflow. We conduct a hyperparameter search on buffer size, learning rate and rollout steps (i.e., the number of episodes collected in each iteration). Table~\ref{tbl:1} lists all the hyperparameters used and searched in both discrete and continuous tasks. To better reproduce our results, we also list the hyperparameters we used for each of the environment in Table~\ref{tbl:2}. To better understand the impact of the hyperparameters, we plot the learning curves with respect to timesteps and running time using different buffer sizes, learning rates and rollout steps. Other hyperparameters are fixed for better visualization. Figure~\ref{fig:1} shows the results.

\textbf{Self-Imitation Learning} We use the code\footnote{\url{https://github.com/junhyukoh/self-imitation-learning}} provided by the authors. We follow the hyperparameters search setting as in the original paper~\cite{oh2018self}. Specifically, the hyperparameters setting is shown in Table~\ref{tbl:3}. For a fair comparison, we use the same random seeds $0$ to $9$.

\textbf{PPO} We use the implementation in OpenAI baselines\footnote{\url{https://github.com/openai/baselines}}. The hyperparameters are searched following~\cite{oh2018self} (Table~\ref{tbl:3}). For a fair comparison, we use the same random seeds $0$ to $9$.

\textbf{Upside-Down Reinforcement Learning~(UDRL)} As far as we know, there is no official implementation. Thus, we use a public implementation in this repository\footnote{\url{https://github.com/haron1100/Upside-Down-Reinforcement-Learning}}. For CartPole, we use the hyperparmeters listed in the repository. These hyperparameters are tuned on CartPole. For Acrobot, we tune the hyperparameters. However, we do not find a working combination. We speculate that it is because it is difficult to choose the command. We do not include the results of UDRL for continuous domains since the repository does not support continuous control. The original paper also only evaluates UDRL on discrete action space~\cite{srivastava2019training}.

\textbf{DDPG \& DQN} We use the implementations in OpenAI baselines. We use the default hyperparameters for DDPG since it is already tuned on MuJoCo. For DQN, we use the default hyperparameters which are tuned on CartPole. For a fair comparison, we use the same random seeds $0$ to $9$.

\subsubsection{A Hundred Seeds Evaluation}
We use the hyperparameters in Table~\ref{tbl:2} and run each experiment $100$ times with seeds $0$ to $99$.

\subsubsection{Playing Games from Raw Image Pixels}
We derive a variant of SSRL for Pong game since we observe that we usually need more data to train the network than the simulated control tasks. Specifically, instead of using a ranking buffer, we use a ring buffer since ranking the experiences in a very large buferr is time consuming. An episode of Pong game can be naturally divided into up to $21$ sub-games, where the agent receives a reward of $1$ if it wins, and $-1$ otherwise. We only feed the data with reward $1$ into the buffer and do not feed the data with reward $-1$. We do not further rank the trajectories in the buffer since the trajectories in the buffer are already good enough. In this way, we essentially regard the winning behaviors as demonstrations. Then we conduct supervised learning to these selected data in the buffer. 

To accelerate the training, we distribute the training process with multiple actors and multiple workers. Each worker will collect trajectories from a separate considered environments and send the data to the buffer. Each worker will sample a batch of data, compute the gradients and send them back to the chief learner asynchronously. In our experiments, we use $4$ actors and $8$ workers.

Our network structure and optimizer follow the setting in A2C~\cite{mnih2016asynchronous}. Specifically, we use the same three Convolutional layers with a fully-connected layer, and RMSprop optimizer. An entropy is also added in the loss function. In each iteration, we collect $1000$ steps of transitions and conduct supervised learning for $250$ steps. All the hyperparmeters are listed in Table~\ref{tbl:4}. The experiment is run $5$ times with random seeds $0$ to $4$.

For A2C, we use the implementation in OpenAI Baselines with default hyperparameters setting. For Rainbow, IQN, C51 and DQN, we use the public results in Dopamine framework\footnote{\url{https://github.com/google/dopamine}}~\cite{castro2018dopamine}.

\begin{table}[t]
\centering
\caption{Hyperparameters of SSRL in Atari Pong.}
\label{tbl:4}
\begin{tabular}{l|l}
\toprule
Hyperparameter & Values \\ \midrule
Buffer size   & $10^6$ \\
Batch size & $256$ \\
Learning rate & $7 \times 10^{-4}$ \\
Entropy coefficient & $0.01$ \\
Rollout steps & $1000$ \\
Training steps & $250$ \\
\bottomrule
\end{tabular}
\end{table}

\begin{table}[t]
\centering
\caption{Hyperparameters of SSRL on  MiniGrid-MultiRoom-N4-S5-v0.}
\label{tbl:5}
\begin{tabular}{l|l}
\toprule
Hyperparameter & Values \\ \midrule
Buffer size   & $1000$ \\
Batch size & $256$ \\
Learning rate & $10^{-3}$ \\
Rollout steps & $1$ \\
Training steps & $5$ \\
\bottomrule
\end{tabular}
\end{table}

\subsubsection{Combining with Exploration Strategies on Hard Exploration Domains}
For SSRL, we use the same code as in simulated control experiments. The hyperparameters are listed in Table~\ref{tbl:5}. For A2C we use the implementation in torch-rl\footnote{\url{https://github.com/lcswillems/rl-starter-files}} which is the recommended implementation that is known to work in the Gym-Minigrid environments\footnote{\url{https://github.com/maximecb/gym-minigrid}}. For more details, please refer to the Github repertory. For the count-based exploration, $\beta$ is set to $0.001$ for SSRL, and tuned from $\{0.1, 0.001, 0.0001, 0.00001\}$ for A2C.

\end{document}